\documentclass[10pt]{article} 
\usepackage[preprint]{tmlr}

\usepackage{graphicx}
\usepackage{amsmath}
\usepackage{amssymb}
\usepackage{booktabs}
\usepackage{multirow}
\usepackage{tikz,pgfplots}
\usepgfplotslibrary{groupplots}
\usepackage[normalem]{ulem}
\usepackage{subcaption}

\newcommand{\vit}[1]{ViT\_B16 #1}

\usepackage[pagebackref,breaklinks,colorlinks]{hyperref}

\usepackage[capitalize]{cleveref}
\crefname{section}{Sec.}{Secs.}
\Crefname{section}{Section}{Sections}
\Crefname{table}{Table}{Tables}
\crefname{table}{Tab.}{Tabs.}

\def\cvprPaperID{11451} 
\def\confName{CVPR}
\def\confYear{2023}

\begin{document}

\title{Data-Efficient Training of CNNs and Transformers with Coresets: A Stability Perspective}
\author{}
\author{Animesh Gupta\textsuperscript{1,2}, Irtiza Hasan\textsuperscript{3}, Dilip K. Prasad\textsuperscript{2} and Deepak K. Gupta\textsuperscript{1,2} \\
\textnormal{\textsuperscript{1}Transmute AI Lab, Indian Institute of Technology, ISM Dhanbad, India} \\ 
\textnormal{\textsuperscript{2}Dept. of Computer Science, UiT The Arctic University of Norway, Tromso, Norway\\
\textnormal{\textsuperscript{3}Immersal, Helsinki, Finland}\\
Email: animeshgupta.thapar@gmail.com, Irtiza.hasan@immersal.com, \{dilip.prasad,deepak.k.gupta\}@uit.no
}}

\maketitle


\begin{abstract}

Coreset selection is among the most effective ways to reduce the training time of CNNs, however, only limited is known on how the resultant models will behave under variations of the coreset size, and choice of datasets and models. Moreover, given the recent paradigm shift towards transformer-based models, it is still an open question how coreset selection would impact their performance. There are several similar intriguing questions that need to be answered for a wide acceptance of coreset selection methods, and this paper attempts to answer some of these. 
    We present a systematic benchmarking setup and perform a rigorous comparison of different coreset selection methods on CNNs and transformers. Our investigation reveals that under certain circumstances, random selection of subsets is more robust and stable when compared with the SOTA selection methods. 
    We demonstrate that the conventional concept of uniform subset sampling across the various classes of the data is not the appropriate choice. Rather samples should be adaptively chosen based on the complexity of the data distribution for each class.
    Transformers are generally pretrained on large datasets, and we show that for certain target datasets, it helps to keep their performance stable at even very small coreset sizes.
    We further show that when no pretraining is done or when the pretrained transformer models are used with non-natural images (\emph{e.g.} medical data), CNNs tend to generalize better than transformers at even very small coreset sizes.
    Lastly, we demonstrate that in the absence of the right pretraining, CNNs are better at learning the semantic coherence between spatially distant objects within an image, and these tend to outperform transformers at almost all choices of the coreset size. \footnote{Our code can be accessed from here \url{https://github.com/transmuteAI/Data-Efficient-Transformers} }
\end{abstract}

\section{Introduction}
\label{sec:intro}
One of the primary reasons for the success of deep learning models is the availability of a large amount of data to train them. However, this success is also accompanied by the requirement of a large amount of computing power, primarily GPU computing, and can lead to significant financial burden as well as energy usage~\citep{strubell2019energy}. For example, training such models on cloud-based GPU servers is often very expensive, and companies would often want to cut down these costs by possibly reducing the training time. Clearly, there is a need to train these deep models in a more efficient manner without compromising on their performance~\citep{asi2019importance}. 

Training large datasets is challenging in terms of required memory, and methods such as stochastic gradient descent or similar are used to gradually optimize the models over small subsets of the training data \citep{effc299a2015}. While such a strategy provides an unbiased estimate of the full-gradient over a series of update steps, it still introduces variance and the convergence is generally very slow for large datasets. The significant increase in the associated training time then becomes a bottleneck for cases where a limited amount of computational time is available. This issue is even more prevalent in adversarial training of deep networks~\citep{dolatabadi2022adversarial}. It is very effective for training robust models against adversarial attacks, however, this comes at the cost of the training process being very slow since it needs to construct adversarial examples for the
entire training data at every iteration.  

Coreset selection is an effective way to reduce the model training time. In simple terms, coreset selection aims at optimally selecting a subset \mbox{$\mathcal{S} \subset \mathcal{D}$} of the full training data $\mathcal{D}$, such that when trained on only $\mathcal{S}$, the model converges approximately to the same solution as $\mathcal{D}$. If this subset $\mathcal{S}$ can be easily found with almost no overhead time, then we see a straightforward speed up of $|\mathcal{D}|/|\mathcal{S}|$ in the overall training time. However, there are several challenges associated with coreset selection, such as identifying the right principle for selecting the training samples, tackling the additional time associated with this process, and appropriately selecting the training hyperparameters based on the selected subset, among others. 
There exist several recent works that attempt to address the issues outlined above. Some examples include CRAIG~\citep{mirzasoleiman2020coresets}, GradMatch~\citep{killamsetty2021grad}, GLISTER~\citep{killamsetty2021glister}, among others. CRAIG and GradMatch are gradient-based methods that identify the optimal coreset by converting the original gradient matching problem to a problem of monotone submodular function optimization, and solve it within a pre-defined error bound. GradMatch additionally uses L2 regularization to reduce the level of dependency on any specific data sample. GLISTER poses a bi-level optimization that solves the selection of subsets as an outer objective and the optimization of model parameters as an inner objective of the optimization problem.
Each of these methods has certain advantages, however, it is unclear at this point as to when to use which method. Or in other words, a rigorous comparative study on model performance vs. training time across different coreset selection methods for recent state-of-the-art (SOTA) models is still missing.

Further, given the recent paradigm shift to transformers, it is still an open question on how coreset selection would impact the performance of transformer networks. It is well known that transformers are data-hungry, and without heavy pretraining, these are known to underperform at downstream tasks. Clearly, with small coresets, this should worsen even more, and this would imply that in the absence of suitable pretraining, coreset selection should not be used at all. However, a clear consensus on this is still missing in the existing literature. It is not yet known which coreset selection method should be preferred for transformers and what should be the coreset sampling frequency. Further, when using pretrained models, it is still unclear whether transformers would be as stable as CNN models when small coresets are chosen. Transformers are known, in general, to work well for tasks related to natural images, and this is especially due to the detailed learning on Imagenet21K \citep{ridnik2021imagenet}, which has similar images. However, there is no clear evidence that the knowledge distilled from Imagenet21K is sufficient enough to keep their performance stable when using small coresets of the training data. Thus, beyond understanding how transformers perform on non-natural images, we also study how the performance is affected when the coreset size is reduced.

The practical applications of deep learning span far beyond image classification on natural images. We consider here two specific cases. First is the application on medical datasets, comprising images different from the standard natural scenes (referred to as non-natural images). For such cases, it is of interest to study whether the conventional pretraining of the models is good enough to keep the performance of transformers stable and at par with a CNN model of equivalent training time. And if this is true, we are further interested in exploring how the performance dips when the coreset size is reduced. Our second application is to seek the goodness of spatial correspondence in non-natural images when working with transformers. Transformers are known to capture the global context very well, however, when seeking strong semantic coherence between spatially distant objects (\emph{e.g.} UltraMNIST classification), it is of interest to see how their pretraining, as well as coreset selection, affect the learning process. Conventionally, coreset selection methods assume an equal number of samples per class for the chosen coreset, however, the complexity of the distribution is different for each class of UltraMNIST, and we are interested to understand how the conventional coreset selection scheme affects the overall performance of CNNs and transformers and what adaptation can be done to improve them.

Finally, we summarize the outcome and novel insights obtained from this study below.
\vspace{-0.5em}
\begin{itemize}
    \itemsep0em
    \item We present a systematic benchmarking setup and perform a rigorous comparison of different coreset selection methods on CNNs and transformers. Our investigation reveals that under certain circumstances, random selection of subsets is more robust and stable when compared with the SOTA selection methods. 
    \item We demonstrate that the conventional concept of uniform subset sampling across the various classes of data is not the appropriate choice. Rather samples should be adaptively sampled based on the complexity of the data distribution for each class.
    \item Transformers are generally pretrained on large datasets, and we show that for certain target datasets, it helps to keep their performance stable at even very small coreset sizes. Thus, these tend to outperform time-equivalent CNN models by large margins.
    \item We further show that when no pretraining is done or when the pretrained transformer models are used with non-natural images (\emph{e.g.} medical data), CNNs tend to generalize better than transformers at even very small coreset sizes.
    \item Lastly, we demonstrate that in the absence of the right pretraining, CNNs are better at learning the semantic coherence between spatially distant objects within an image, and these tend to outperform transformers at almost all choices of the coreset size. 
\end{itemize}

\section{Related Work}
\label{sec:related_work}
Most of the recent deep learning models are computationally expensive and require training on very large datasets. This leads to huge energy-related costs, and there is a strong desire to make the training energy-efficient \citep{strubell2019energy}. This can be achieved in many ways, such as training the model on lower bit representation or using lower resolution images \citep{banner2018neurips, chin2019mlsys}, designing training schemes adapted to the hardware design, \citep{wu2019tnls, hoffer2018neurips}, and using only the important data samples for training \citep{mirzasoleiman2020coresets,killamsetty2021glister, killamsetty2021grad}, among others. This paper focuses on making the training efficient through the selective sampling of the training data, also referred to as coreset selection in the existing literature.

Coreset selection methods have been studied for over two decades \citep{feldman2020introduction}, designed originally for the applications in computational geometry \citep{agarwal2005geometric}, but quickly adopted later by the machine learning community and applied to classical machine learning based problems. As described in \cite{mirzasoleiman2020coresets}, initially, these problems varied from, finding a high likelihood solution for $k$-means and $k$-median clustering \citep{har2004coresets}, SVM \citep{clarkson2010coresets}, graphical model training \citep{molina2018core}, logistic regression \citep{huggins2016coresets},  naive bayes \citep{wei2015submodularity} and nsytrom methods \citep{musco2017recursive}.

With the emergence of data-driven approaches, one of the natural applications of coreset selection is deep learning, as data is a critical component of all deep-learning-based algorithms. However, this application is not straightforward due to the innate nature of coreset algorithms, which makes them extremely task-specific. This intrinsic property of the coreset algorithm poses a challenge in their utility for deep-learning-based methodologies \citep{killamsetty2021grad}. Nonetheless, recent works using coresets in deep learning have shown promise. Starting from one of the pioneering works \citep{mirzasoleiman2020coresets}, where authors proposed to select a subset of training data, that can capture the approximation of the full gradient. 

Some recent works have used coreset algorithms, not only for data sub-selection but also for data cleaning and learning from noisy labels. For instance, Mirzasoleiman et al. \citep{mirzasoleiman2020coresetsb}, proposed a technique based on coresets, which would select clean data points and provide an approximation of low-rank jacobian matrix for a more generic representation learning. Along with noisy data, class imbalance in training data is often the cause of over-fitting and poor model performance. Therefore, coreset also gained attention in this domain, and a few recent works, such as \cite{killamsetty2021glister, killamsetty2021grad}, have successfully deployed coreset algorithms to tackle the class-imbalance problem, which in turn leads to more generalized models. Besides data, other important components of deep-learning-based algorithms are the model size and number of parameters. One of the first works is \cite{baykal2018data}, which proposed to use a coreset, not for data sub-selection but for model parameter reduction. This seminal work laid the foundation for several other works, such as \cite{mussay2021data,mussay2019datab,tukan2022pruning}, using coresets for neural network pruning. However, all of the aforementioned works either limited their study to a specific dataset or studied how coresets can be utilized for neural network pruning, given a specific architecture (CNNs, for instance).


However, in this work, we study coresets from a data perspective. Several recent works have attempted to study coreset selection behaviour using different settings of the data. ~\cite{mirzasoleiman2020coresetsb} used 50\% symmetric noise versions of the CIFAR10 and CIFAR100 and studied the effect of size of coreset on the final accuracy. ~\cite{killamsetty2021glister} showed the effectiveness of their method by using either shallow network architectures (a two-layer fully-connected neural network with 100 hidden nodes) or deep learning models such as Resnet18 on small-scale datasets like MNIST and CIFAR10. These and most other methods are limited to small-scale datasets which share a distribution similar to to ImageNet21k \cite{X}. These work have reported full training time but none of the method studies the coreset selection time of different coreset selection methods. We show in this study that although we generally assume the coreset selection time to be negligible, it is not. Our work presented in this paper covers study on large-scale models like Vision Transformers and is experimented on diverse datasets including those from a different domain such as medical datasets. We also study the influence of the individual coreset selection time taken by each method for different coreset sizes.

While the various coreset selection methods that exist in the literature have been successful in significantly reducing the associated computational time, only limited research has been conducted to perform comparative analysis of these methods. To our knowledge, only ~\cite{killamsetty2021grad} have compared the existing coreset selection methods in terms of performance under a unified setting. However, the scope of this study is still limited in various aspects. First and foremost, the study is restricted to CNN architectures and no experiments with Transformer models were included. With the recent popularity gained by Transformers, it is of interest to study and compare coreset selection methods on them. Further, the previous work has only focused on studying coreset selection methods in the context of  natural images similar to the ImageNet21K dataset, however, we extend our study to also include non-natural images such as medical images. Lastly, coreset selection methods have been tested only for scenarios where the complexity of the distribution across different classes is balanced. In this paper, we also consider a scenario where different classes of a classification problem have different levels of complexity, and study the impact of coreset selection on the model performance for such as dataset. 

Overall, the study of coreset selection methods in the context of data-efficient training has been very limited, and none of the existing works have thoroughly analyzed the influence of the coreset selection time as well. For the first time, we present a thorough quantitative and objective perspective to coreset selection algorithms on different benchmarks. Additionally, we also study how different architectures (CNNs and Transformers) respond to subtle differences in different coreset selection algorithms to provide insights for future research.







\section{Coreset Selection}
\subsection{General description}

\begin{figure}
    \centering
    \includegraphics[trim=0 0 0.1cm 0 , clip=true, scale=0.25]{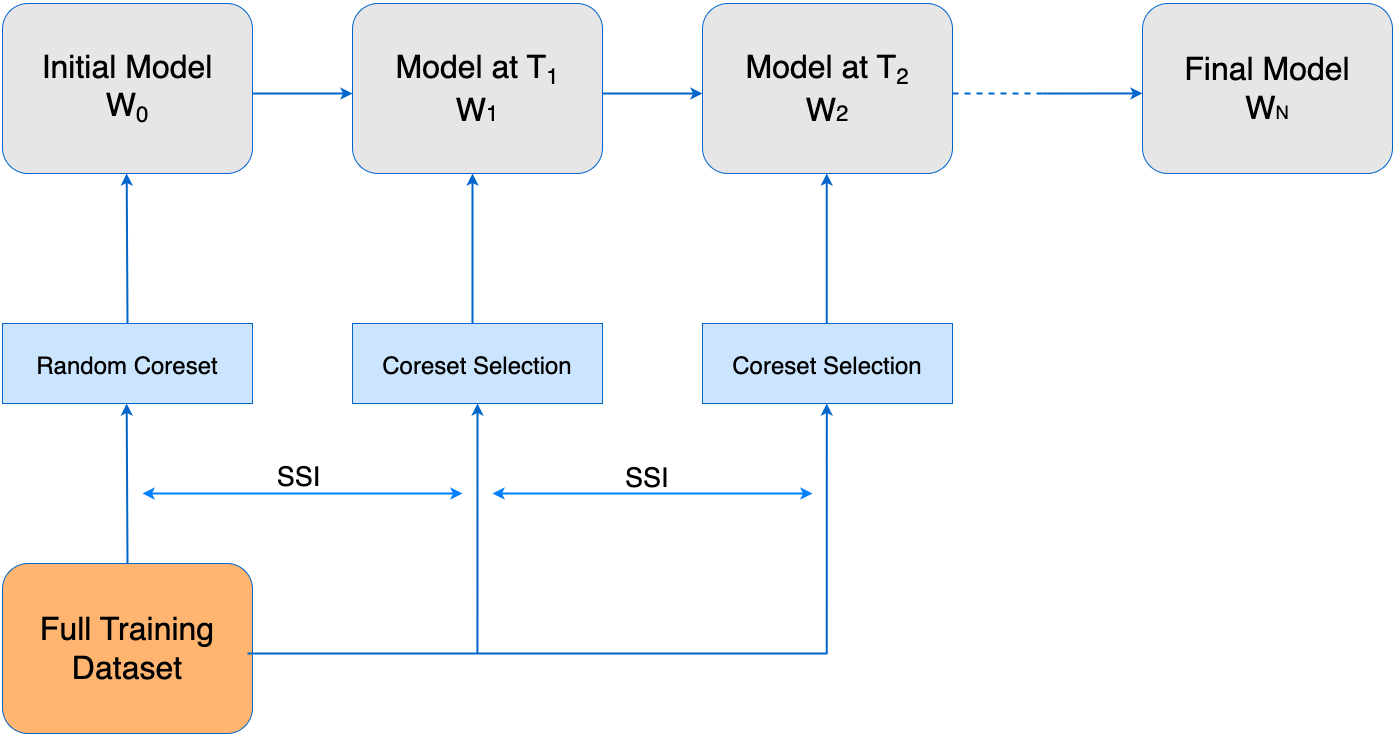}
    \caption{Schematic representation of coreset selection for model training. Here SSI denotes the subset selection interval.}
    \label{fig:schema}
\end{figure}
Coreset selection refers to identifying a subset of the training data that, when used to train the model, can approximately reach the same optimal solution as when trained with the full data. 
Fig. \ref{fig:schema} presents an overview of the process. Generally, a coreset budget is chosen, and an initial random subset is used to choose the first coreset.

This coreset is then used to train the model for a certain number of epochs $T_1$. At this stage, another coreset selection is done, but with more sophisticated approaches, such as GradMatch, and GLISTER, among others. The model is trained again till epoch $T_2$ and another coreset selection happens. This process continues till the model is fully trained. Coreset selection happens after pre-defined intervals (referred to as subset selection internal (SSI) in this paper).

To further explain the concept of coreset selection, we formulate the mathematical optimization problem solved as part of training a neural network model. Let $f: \mathbb{R}^d \rightarrow \mathbb{R}^k$ denote a neural network with weights $\mathbf{w}$. During training, the parameter set $\mathbf{w}$ is optimized with respect to a certain objective function, and the associated optimization problem is to find the optimal weights $\mathbf{w}^*$ as follows.

\begin{equation}
\mathbf{w}^* = \underset{\mathbf{w} \in \mathcal{W}}{\text{argmin}} \enskip \mathcal{L}(f(\mathbf{w; x}), \mathbf{y})
\end{equation}
where $\mathcal{D} = \{\mathbf{x}, \mathbf{y}\}$ denotes the input-output pairs in the training data, and $\mathcal{W}$ denotes the parameter search space. Further, $\mathcal{L}^*$ denotes the optimized objective value above.

The simple idea of coreset selection is to find the data subset $\mathcal{S} \subset \mathcal{D}$ where $\mathcal{S} = \{\mathbf{\tilde{x}}, \mathbf{\tilde{y}}\}$ such that $\tilde{\mathcal{L}}^* \approx \mathcal{L}$, where
\begin{equation}
\tilde{\mathcal{L}}^* = \underset{\mathbf{w} \in \mathcal{W}}{\text{min}} \enskip \mathcal{L}(f(\mathbf{w; \tilde{x}}), \mathbf{\tilde{y}})
\label{eq-coreset}
\end{equation}
 Assuming that the total time spent on coreset selection is very small compared to solving the optimization problem stated in Eq. \ref{eq-coreset}, it is evident that a straightforward speedup of $|\mathcal{D}/\mathcal{S}|$ fold can be obtained. It is important to note that the above outlined assumption is an ideal scenario that still needs to be achieved and a perfect coreset selection method that complies with it is still be to identified. To take this into account, we consistently consider the coreset selection time as well in all the experiments reported in this paper.
 
 However, there are several challenges that make the selection of $\mathcal{S}$ hard.
 The first is to identify the right guiding principle for coreset selection. For example, one could choose more points around the decision boundary or alternatively prefer to go for a more diverse distribution. Depending on the choice, the resultant model is expected to behave differently. Secondly, this paper focuses on minimizing the desired training time, and it is important that the time spent on selecting the coresets is not very significant. Another equally important challenge is appropriately adapting the model's hyperparameters to the obtained coreset. Most SOTA coreset selection methods that exist so far have focused on eliminating the challenges posed above, however, each of these has its own benefits as well as limitations, and we will study this aspect in the paper.

\subsection{Choice of coreset selection methods}
In this paper, we choose 3 recent coreset selection methods, and in addition, we also include the random coreset selection approach. We review in brief these coreset selection algorithms along with their formal definitions \footnote{For more details on the subselection method, please refer to \cite{Killamsetty_CORDS_COResets_and_2022}}.




\noindent\textbf{CRAIG.}
CRAIG~\citep{mirzasoleiman2020coresets} is an incremental gradient (IG) method, in which a weighted coreset is selected based on the estimation of full-gradient by maximizing a sub-modular function. Subsequently, applying IG to the selected coreset is shown to converge to the near-optimal solution at a speed that is inversely proportional to the size of the coreset.  

\begin{equation}
\sum_{{i} \in \mathcal{D}} \underset{{j} \in \mathcal{S},\mathcal{|S|}\leq\mathcal{K}}{\text{min}} \enskip ||\mathbf{\nabla_{\mathbf{w}}}\mathcal{L}^{i}\mathbf{w} - \mathbf{\nabla_{\mathbf{w}}}\mathcal{L}^{j}\mathbf{w}||, 
\label{eq-craig}
\end{equation}

where $\mathcal{K}$ is the budget for the subset. It is worth mentioning that CRAIG can be used to select data subsets at each desired epoch. Due to the nature of the problem (instance of facility location problems) of being a submodular function, it can be solved using greedy selection methods. 

\noindent\textbf{GradMatch. }In brief, the underlying idea of GradMatch~\citep{killamsetty2021grad} is to locate subsets of data that closely mimic the gradients of the training (or validation) set. This localization is done through an orthogonal matching pursuit algorithm. More formally, the GradMatch algorithm can be defined as

\begin{equation}
\underset{\boldsymbol\phi,\mathcal{S}:\mathcal{|S|}\leq\mathcal{K}}{\text{min}} ||\sum_{{i} \in \mathcal{S}} \enskip {\phi}_{i} \mathbf{\nabla_{\mathbf{w}}}\mathcal{L}^{i}\mathbf{(w)} - \mathbf{\nabla_{\mathbf{w}}}\mathcal{L}\mathbf{(w)} ||,
\label{eq-gradmatch}
\end{equation}

where $\boldsymbol\phi$ is a weight vector containing the weight for each data instance and $\mathcal{K}$ is the budget for the subset.  As discussed above, the optimization problem is solved using the Orthogonal Matching Pursuit (OMP) algorithm.


\noindent\textbf{GLISTER. }
This approach \citep{killamsetty2021glister} tackles the data sub-selection task by solving a bi-level optimization problem where the goal is to maximize the log-likelihood on a held-out validation set. More formally, it can be stated as

\begin{equation}
\underset{\mathcal{S} \subseteq \mathcal{D},\mathcal{|S|}\leq\mathcal{K}}{\text{argmin}}\mathcal{L}_{v}  (\underset{\mathbf{w} \in \mathcal{W}}{\text{argmin}} - \mathbf{\eta}\mathbf{\nabla_{\theta}} \mathcal{L}(\mathbf{\mathbf{w}}, \mathcal{S})\mathbf{(\nu)}).
\label{eq-glister}
\end{equation}

Here, $\nu$ refers to the validation set and $\mathcal{L}_{v}$ accordingly denotes the validation loss.  Lastly, $\eta$ is the step size. Since GLISTER is a bi-level optimization problem, it solves the first problem by utilizing online one-step meta approximation by taking a single gradient step. The second approximation is made by using  Taylor-Series approximation (through the greedy search algorithm).


\noindent\textbf{Random selection. }It involves selecting a random coreset from the full training data, without any heuristics and learning. In this work, random selection serves as a baseline.

\subsection{Benchmarking strategy}
For a fair comparison of different coreset selection methods across the different CNN and transformer methods, we define here a benchmarking setup. Note that this setup is designed with respect to making the training process more efficient. In this regard, we also introduce some parameters that will be used consistently across all experiments of this paper to compare the results. 

\textbf{Definitions. }There are several parameters that are important for the setup, and we define them here. We also provide the motivation behind the relevance of each parameter in the context of comparing various coreset methods for CNNs and transformers across the different coreset sizes.

\textit{Effective data per epoch (EDPE). } It refers to the fraction of the trainset used as a coreset for training the model per epoch. EDPE gives an approximate estimate of the reduction in the training time of the model and excludes the time spent on selecting the coresets.

\textit{Subset selection interval (SSI). }It refers to the number of epochs that need to be run between two consecutive coreset selections. Thus, for a setup with a fixed number of epochs, a higher value of SSI would imply a lesser number of coreset selection cycles. In general, the influence of SSI on the overall performance of the trained model is not known well, and we experiment with different values of SSI to find the right number of coreset selections that maximize the performance of the model.

\textit{Total selection time. }It refers to the amount of time spent on selecting the optimal subset of the training data using the chosen coreset selection method. Although it is assumed that the cost associated with coreset selection is negligible, we have seen that this time plays a crucial role in evaluating the performance of any given coreset selection method. Details on the total selection time for all the experiments are reported in the supplementary section of this paper.

\textit{Total time. }It corresponds to the total time spent on training the model as well as coreset selection. In terms of computing the efficiency of a training process, we consider total time as the right metric and report it frequently in our results presented later in this paper.

\textit{Epochs. }Similar to any training procedure in deep learning, epoch refers to the number of passes through the entire dataset (coreset here). It can also be interpreted as the iterative effort invested to identify the best model and has a significant impact on the training cost. Note that best models for different methods might be obtained at a different epoch value. However, when calculating the training cost, we consider the total number of epochs that are planned beforehand. The reason is that from the implementation perspective, the best model cannot be known till the training process has been completed.

\textbf{Setup. }Among the defined control parameters, we fix the total number of epochs to be constant across all the experiments of this paper, and it is denoted as $N$. Further, we experiment with different values of SSI to identify the right balance between the subset sampling frequency and the performance of the trained model. No constraint is imposed on the other three parameters, and these are measured for different experiments. Among the total time and the total selection time, total time is a more practical choice for study. However, since the coreset selection can significantly vary across CPUs depending on the availability of threads, we also study training time for each model.

\section{Experiments}
\subsection{Implementation details}

\textbf{Models. }In this work, we experimented with four baselines, two CNN-based and two transformer-based backbones. We adopted a ResNet-50 model, along with an application-oriented, embedded vision-based model MobileNetV3-Large (referred to as MobileNetV3). For ResNet, we used the recent ResNetV2-50x1 model (referring further as ResNet50), where x1 denotes the factor by which every hidden layer is widened. Saved checkpoint models for ResNet50 and MobilenetV3 were obtained from \cite{kolesnikov2020big} and \cite{rw2019timm}, respectively. For transformers, we used two SOTA models, namely ViT (B16) and Swin-Transformer, the base variant. These choices were dictated by the fact that these models have achieved SOTA results on several visual perception tasks (classification, object detection, etc.). We used the standard ImageNet21k weights publicly available for initializing our models.



\textbf{Datasets. }For benchmarking the coreset selection methods, we experiment on four popular datasets.

\textit{CIFAR10. } CIFAR10 \citep{krizhevsky2009learning} dataset contains 50000 images with 10 classes. We used $224 \times 224$ as the input to the models in our experiments. We chose this dataset due to its small size and to see the performance of Transformers when trained with small coreset sizes.

\textit{TinyImageNet. } TinyImageNet \citep{wu2017tiny} is a subset of ImageNet21k containing a total of 100k images with 200 classes. This dataset has been created as a smaller, but very good representative of the original Imagenet \cite{russakovsky2015imagenet}. Thus, we use it to show a comparison study of coresets with different backbones close to the ImageNet. 

\textit{APTOS-2019.} APTOS 2019 \citep{aptos2019-blindness-detection}, is medical dataset related to blindness detection. The dataset contains 5 classes, with classification labels ranging from 0 to 4, showing the severity level. In total, APTOS \citep{aptos2019-blindness-detection}, contains 3662 fundus images. This dataset is a good example of non-natural images, and we expect that strong priors learnt on ImageNet21k might not adapt well to this domain.

\textit{UltraMNIST. }This dataset \citep{ultamnist2022} is designed primarily to assess the capability of models to capture semantic coherency in large images. We use $512 \times 512$ variants of the original dataset. Each sample of UltraMNIST comprises 3-5 MNIST digits of varying scales, and the sum of these digits can vary from 0 to 27, also defining the class labels 0-27. Note that each label in the dataset can be formed with different combinations of the digits, with some classes having very less combinations, while some can be formed by a large number of combinations. Thus, the complexity of the distribution of every class is different, and this makes UltraMNIST an interesting dataset for coreset selection study. For more details related to the dataset, see \cite{ultamnist2022}.


\textbf{Configuration of hyperparameters. } We used the hyperparameters taken from the extensive study by Steiner et al.~\citep{steiner2021train}: SGD momentum optimizer, cosine decay schedule, no weight decay, gradient clipping at global norm 1 and 224×224 image resolution for CIFAR10, Tiny-Imagenet, APTOS-2019, and 512x512 for UltraMNIST.
For the sake of fair comparison, all other hyperparameters are similar across all the backbones and datasets used in this study.



\subsection{Results}
\begin{figure}
\centering
\begin{subfigure}{1\linewidth}  
\centering
\begin{tikzpicture}[scale=0.68]
   \begin{axis}[ xlabel={Training Time [Mins]},
    ylabel={Accuracy [\%]},
   xmin=0, xmax=280,
    ymin=15, ymax=85,
    xtick={3, 40, 80, 120, 160, 200, 240, 280},
    ytick={15, 25, 35, 45, 55, 65, 75, 85},
    legend pos=south east,
    ymajorgrids=true,
    grid style=dashed,
    line width=1pt,
    ]
 \addplot[
    color=blue,
    mark=star,
    ]
    coordinates {
    (3.53,60.83)
    (31.78, 72.21)
    (95.35,78.94)
    (155.38,80.87)
    (250.73,79.79)};
    \addlegendentry{CRAIG-ResNet50}

\addplot[
    color=red,
    mark=star,
    ]
    coordinates {
    (3.53, 72.49)
    (31.78, 80.24)
    (95.35, 81.09)
    (155.38, 80.41)
    (250.73, 79.40)};
    \addlegendentry{GradMatch-ResNet50}

\addplot[
    color=violet,
    mark=star,
    ]
    coordinates {
    (3.53, 63.99)
    (31.78, 75.08)
    (95.35, 76.73)
    (155.38, 78.11)
    (250.73, 79.09)};
    \addlegendentry{GLISTER-ResNet50}

    \addplot[
        color=black,
        mark=star,
        ]
        coordinates {
        (3.53, 52.21)
        (31.78, 72.07)
        (95.35, 75.59)
        (155.38, 78.17)
        (250.73, 78.81)
        };
        \addlegendentry{Random-ResNet50}

    \addplot[
    color=blue,
    mark=star,
    dashed
    ]
    coordinates {
    (3.36,17.15)
    (30.24, 59.15)
    (89.05,72.10)
    (147.87,75.35)
    (236.92,75.76)};
    \addlegendentry{CRAIG-MobileNetV3}

\addplot[
    color=red,
    mark=star,
    dashed
    ]
    coordinates {
    (3.36, 29.35)
    (30.24, 72.42)
    (89.05, 75.47)
    (147.87, 75.75)
    (236.92, 75.39)};
    \addlegendentry{GradMatch-MobileNetV3}

\addplot[
    color=violet,
    mark=star,
    dashed
    ]
    coordinates {
    (3.36, 20.68)
    (30.24, 68.37)
    (89.05, 75.15)
    (147.87, 75.45)
    (236.92, 75.87)};
    \addlegendentry{GLISTER-MobileNetV3}
    \addplot[
        color=black,
        mark=star,
        dashed
        ]
        coordinates {
        (3.36, 22.72)
        (30.24, 62.58)
        (89.05, 70.60)
        (147.87, 72.34)
        (236.92, 74.44)
        };
        \addlegendentry{Random-MobileNetV3}
    \end{axis}
    \end{tikzpicture}%
    ~%
    \begin{tikzpicture}[ scale=0.68]
    \begin{axis}[ xlabel={Total time [Mins]},
    ylabel={Accuracy [\%]},
    xmin=0, xmax=420,
    ymin=15, ymax=85,
    xtick={3, 60, 120, 180, 240, 300, 360, 420},
    ytick={15, 25, 35, 45, 55, 65, 75, 85},
    ymajorgrids=true,
    grid style=dashed,
    line width=1pt,    
    ]
\addplot[
    color=blue,
    mark=star,
    ]
    coordinates {
    (140.30,60.83)
    (168.94, 72.21)
    (231.58,78.94)
    (294.23,80.87)
    (389.09,79.79)};

\addplot[
    color=red,
    mark=star,
    ]
    coordinates {
    (81.42, 72.49)
    (109.02, 80.24)
    (169.40, 81.09)
    (229.79, 80.41)
    (321.22, 79.40)};

\addplot[
    color=violet,
    mark=star,
    ]
    coordinates {
    (66.01, 63.99)
    (94.27, 75.08)
    (154.30, 76.73)
    (217.87, 78.11)
    (309.69, 79.09)};

    \addplot[
        color=black,
        mark=star,
        ]
        coordinates {
        (3.53, 52.21)
        (31.78, 72.07)
        (95.35, 75.59)
        (155.38, 78.17)
        (250.73, 78.81)
        };

    \addplot[
    color=blue,
    mark=star,
    dashed
    ]
    coordinates {
    (165.27,17.15)
    (192.15, 59.15)
    (250.97,72.10)
    (309.78,75.35)
    (398.84,75.76)};

\addplot[
    color=red,
    mark=star,
    dashed
    ]
    coordinates {
    (80.91, 29.35)
    (107.79, 72.42)
    (166.60, 75.47)
    (225.42, 75.75)
    (314.47, 75.39)};

\addplot[
    color=violet,
    mark=star,
    dashed
    ]
    coordinates {
    (42.30, 20.68)
    (69.18, 68.37)
    (127.99, 75.15)
    (186.81, 75.45)
    (275.87, 75.87)};

\addplot[
    color=black,
    mark=star,
    dashed
    ]
    coordinates {
    (3.36, 22.72)
    (30.24, 62.58)
    (89.05, 70.60)
    (147.87, 72.34)
    (236.92, 74.44)
    };
    \end{axis}
    \end{tikzpicture}%
\vspace{-0.2em}
\caption{CNN (Solid - ResNet50; Dashed - MobileNetV3)}
\label{fig;sub-comp-corecnn}
\end{subfigure}

\begin{subfigure}{1\linewidth}   
\centering
\begin{tikzpicture}[scale=0.68]
   \begin{axis}[ xlabel={Training Time [Mins]},
    ylabel={Accuracy [\%]},
   xmin=0, xmax=950,
    ymin=35, ymax=95,
    xtick={5, 150, 350, 550, 750, 950},
    ytick={35, 45, 55, 65, 75, 85, 95},
    legend pos=south east,
    ymajorgrids=true,
    grid style=dashed,
    line width=1pt,
    ]
 \addplot[
    color=blue,
    mark=star,
    ]
    coordinates {
    (5.53,48.30)
    (49.77, 84.97)
    (146.56,88.48)
    (243.35,89.97)
    (389.91,89.46)};
    \addlegendentry{CRAIG-ViT\_B16}

\addplot[
    color=red,
    mark=star,
    ]
    coordinates {
    (5.53,59.51)
    (49.77, 89.07)
    (146.56,89.89)
    (243.35,89.82)
    (389.91,89.50)};
    \addlegendentry{GradMatch-ViT\_B16}

\addplot[
    color=violet,
    mark=star,
    ]
    coordinates {
    (5.53,45.59)
    (49.77, 88.89)
    (146.56,89.82)
    (243.35,89.69)
    (389.91,89.60)};
    \addlegendentry{GLISTER-ViT\_B16}

    \addplot[
        color=black,
        mark=star,
        ]
        coordinates {
        (5.53,40.24)
        (49.77, 86.69)
        (146.56,88.44)
        (243.35,88.67)
        (389.91,89.33)};
        \addlegendentry{Random-ViT\_B16}

    \addplot[
    color=blue,
    mark=star,
    dashed
    ]
    coordinates {
    (13.11, 76.14)
    (118.04, 86.60)
    (347.57, 91.28)
    (577.09, 91.51)
    (924.67, 90.89)
    };
    \addlegendentry{CRAIG-Swin}

\addplot[
    color=red,
    mark=star,
    dashed
    ]
    coordinates {
    (13.11, 85.37)
    (118.04, 91.48)
    (347.57, 91.90)
    (577.09, 91.44)
    };
    \addlegendentry{GradMatch-Swin}

\addplot[
    color=violet,
    mark=star,
    dashed
    ]
    coordinates {
    (13.11, 86.41)
    (118.04, 91.63)
    (347.57, 91.87)
    (577.09, 91.47)
    (924.67, 90.97)
    };
    \addlegendentry{GLISTER-Swin}

    \addplot[
        color=black,
        mark=star,
        dashed
        ]
        coordinates {
        (13.11, 64.97)
        (118.04, 87.77)
        (347.57, 90.02)
        (577.09, 90.00)
        };
        \addlegendentry{Random-Swin}
    \end{axis}
    \end{tikzpicture}%
    ~%
    \begin{tikzpicture}[ scale=0.68]
    \begin{axis}[ xlabel={Total time [Mins]},
    ylabel={Accuracy [\%]},
    xmin=0, xmax=1310,
    ymin=30, ymax=100,
    xtick={5, 110, 310, 510, 710, 910, 1110, 1310},
    ytick={30, 40, 50, 60, 70, 80, 90, 100},
    ymajorgrids=true,
    grid style=dashed,
    line width=1pt,    
    ]

    \addplot[
    color=blue,
    mark=star,
    ]
    coordinates {
    (191.59,48.30)
    (235.84, 84.97)
    (332.62,88.48)
    (429.41,89.97)
    (575.97,89.46)};

    \addplot[
        color=red,
        mark=star,
        ]
        coordinates {
        (133.82,59.51)
        (178.07, 89.07)
        (274.86,89.89)
        (371.64,89.82)
        (518.21,89.50)};

    \addplot[
        color=violet,
        mark=star,
        ]
        coordinates {
        (87.67,45.59)
        (131.92, 88.89)
        (228.71,89.82)
        (325.49,89.69)
        (472.06,89.60)};

    \addplot[
        color=black,
        mark=star,
        ]
        coordinates {
        (5.53,40.24)
        (49.77, 86.69)
        (146.56,88.44)
        (243.35,88.67)
        (389.91,89.33)};

    \addplot[
        color=blue,
        mark=star,
        dashed
        ]
        coordinates {
        (239.26, 76.14)
        (343.56, 86.60)
        (573.72, 91.28)
        (803.24, 91.51)
        (1150.82, 90.89)
        };

    \addplot[
        color=red,
        mark=star,
        dashed
        ]
        coordinates {
        (138.36, 85.37)
        (243.29, 91.48)
        (472.82, 91.90)
        (702.34, 91.44)
        };

    \addplot[
        color=violet,
        mark=star,
        dashed
        ]
        coordinates {
        (113.39, 86.41)
        (218.32, 91.63)
        (447.89, 91.87)
        (677.37, 91.47)
        (1024.94, 90.97)
        };

    \addplot[
        color=black,
        mark=star,
        dashed
        ]
        coordinates {
        (13.11, 64.97)
        (118.04, 87.77)
        (347.57, 90.02)
        (577.09, 90.00)
        };

    \end{axis}
    \end{tikzpicture}%
\vspace{-0.2em}
\caption{Transformer (Solid - ViT\_B16; Dashed - Swin)}
\label{fig:sub-comp-coretran}
\end{subfigure}
\vspace{-2em}
\caption{Performance scores for various coreset selection methods for different choices of training time as well as total time. Models are pretrained on ImageNet21k and further trained for TinyImageNet classification.}
\label{fig:sub-comp-core}
\end{figure}
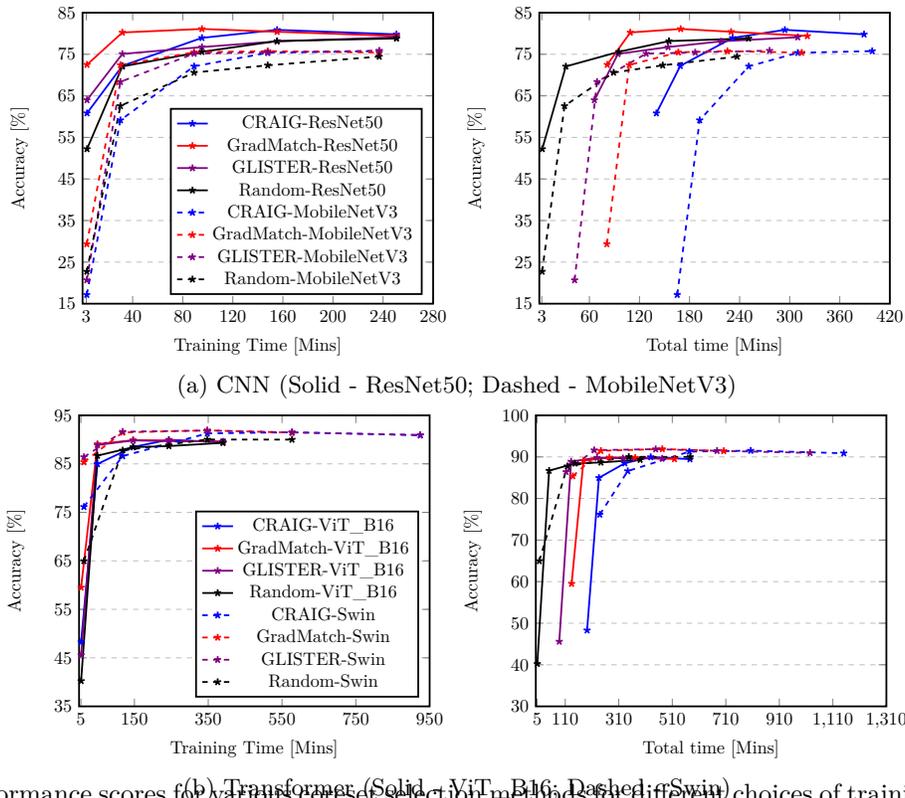


We discuss below the results related to the various benchmarking experiments conducted in this paper.

\textbf{Comparison of coreset selection methods. }To the best of our knowledge, a clear comparison of the recent coreset selection methods in terms of training time does not exist, especially for transformers. We begin with benchmarking our selected models, using 4 coreset selection methods as described above. Fig. \ref{fig:sub-comp-core}, illustrates the performance comparison in terms of training time along with the total time (training + coreset selection). From a practical perspective, the total time needs to be reduced. However, it can be argued that coreset selection can be scaled across multiple CPU threads, thereby making the coreset selection time very small. 


For ResNet and MobileNet, except for random selection, GradMatch consistently outperforms other methods. Whereas, for longer training time, CRAIG is only marginally better. For very small coresets, GLISTER is outperformed by even the random selection method. As for the total time, except for GradMatch, all methods get outperformed by random selection (especially, small time budgets), which is counterintuitive and implies that the coreset selection methods are not stable for CNNs at low training time budgets. 

Similar observations are made for transformer models (all training time). For small time budgets, we see that for ViT\_B16 as well as the Swin transformer, random coreset selection performs better than other methods, clearly indicating that a generalizable solution to make models time-efficient through coreset selection is still missing.


 

\begin{figure}
        \centering
\begin{tikzpicture}[scale=0.68]
   \begin{axis}[ xlabel={EDPE [\%]},
    ylabel={Accuracy [\%]},
    xmin=0, xmax=100,
    ymin=25, ymax=95,
   xtick={1, 10, 20, 30, 40, 50, 60, 70, 80, 90, 100},
    ytick={0, 25, 35, 45, 55, 65, 75, 85, 95},
    legend pos=south east,
    ymajorgrids=true,
    grid style=dashed,
    line width=1pt,
    ]
\addplot[
    color=blue,
    mark=star,
    dashed
    ]
    coordinates {
    (1,33.04)
    (10,50.57)
    (20,53.70)
    (30,57.79)
    (40,60.10)
    (50,62.02)
    (60,63.12)
    (70,64.88)
    (80,66.44)
    (90,67.03)
    };
    \addlegendentry{ViT\_B16-Craig}

\addplot[
    color=blue,
    mark=star,
    dotted
    ]
    coordinates {
    (1,30.00)
    (10,49.04)
    (50,63.53)
    (80,67.22)
    };
    \addlegendentry{ViT\_B16-GradMatch}

    \addplot[
    color=red,
    mark=star,
    dashed
    ]
    coordinates {
    (1,28.98)
    (10,61.27)
    (20,71.22)
    (30,77.76)
    (40,81.99)
    (50,84.60)
    (60,86.79)
    (70,87.97)
    (80,89.33)
    (90,89.64)
    };
    \addlegendentry{ResNet50-Craig}

    \addplot[
    color=red,
    mark=star,
    dotted
    ]
    coordinates {
    (1,27.15)
    (10,60.61)
    (50,85.09)
    (80,89.55)
    };
    \addlegendentry{ResNet50-GradMatch}
    \end{axis}
    \end{tikzpicture}%
    ~%
    \begin{tikzpicture}[ scale=0.68]
    \begin{axis}[ xlabel={Time [Mins]},
    ylabel={Accuracy [\%]},
    xmin=0, xmax=245,
    ymin=25, ymax=95,
    xtick={5, 35, 65, 95, 125, 155, 185, 215, 245},
    ytick={0, 25, 35, 45, 55, 65, 75, 85, 95},
    legend pos=south east,
    ymajorgrids=true,
    grid style=dashed,
    line width=1pt,    
    ]
\addplot[
    color=blue,
    mark=star,
    dashed
    ]
    coordinates {
    (8.02,33.04)
    (29.24,50.57)
    (52.25,53.70)
    (78.51,57.79)
    (100.47,60.10)
    (122.60,62.02)
    (147.65,63.12)
    (170.78,64.88)
    (193.10,66.44)
    (217.38,67.03)
    };
    \addlegendentry{ViT\_B16-Craig}

\addplot[
    color=blue,
    mark=star,
    dotted
    ]
    coordinates {
    (7.25,30.00)
    (28.16,49.04)
    (116.13,63.53)
    (183.22,67.22)
    };
    \addlegendentry{ViT\_B16-GradMatch}

\addplot[
    color=red,
    mark=star,
    dashed
    ]
    coordinates {
    (4.13,28.98)
    (17.29,61.27)
    (31.11,71.22)
    (46.30,77.76)
    (61.54,81.99)
    (75.40,84.60)
    (91.03,86.79)
    (107.29,87.97)
    (120.97,89.33)
    (134.64,89.64)
    };
    \addlegendentry{ResNet50-Craig}

    \addplot[
    color=red,
    mark=star,
    dotted
    ]
    coordinates {
    (3.62,27.15)
    (18.57,60.61)
    (73.27,85.09)
    (115.79,89.55)
    };
    \addlegendentry{ResNet50-GradMatch}
    \end{axis}
    \end{tikzpicture}%
\vspace{-1em}
\caption{Performance comparison between Randomly Initialised \vit and ResNet50 for time, EDPE and performance on CIFAR10, SSI=50.}
\label{fig:rand-train}
\end{figure}
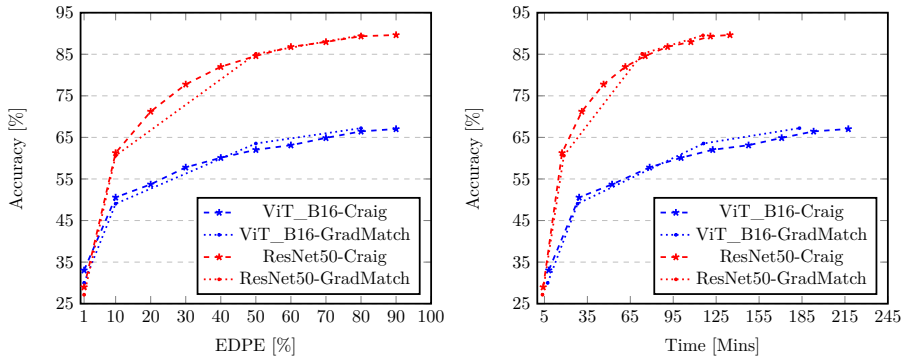

\textbf{Data subselection under no pretraining. } 
For this study, we conducted experiments on the CIFAR10 dataset, using ResNet50 and VIT-B16 for CRAIG and GradMatch methods. Related results are shown in Fig. \ref{fig:rand-train}. It can be seen that CNN architecture outperforms transformers significantly without pretraining. We argue that it is because transformers are data-hungry and thus are affected severely in the absence of pretraining. More importantly, this also illustrates that without appropriate pretraining, for transformers, coreset selection might become irrelevant. 





\begin{figure}
        \centering
\begin{tikzpicture}[scale=0.68]
   \begin{axis}[ xlabel={EDPE [\%]},
    ylabel={Accuracy [\%]},
    xmin=0, xmax=100,
    ymin=90, ymax=100,
   xtick={1, 10, 20, 30, 40, 50, 60, 70, 80, 90, 100},
    ytick={90, 92, 94, 96, 98, 100},
    legend pos=south east,
    ymajorgrids=true,
    grid style=dashed,
    line width=1pt,
    ]
 \addplot[
    color=blue,
    mark=star,
    dashed
    ]
    coordinates {
    (1,97.49)
    (10,98.25)
    (20,98.64)
    (30,98.69)
    (40,98.87)
    (50,98.92)
    (60,98.99)
    (70,98.94)
    (80,98.86)
    (90,98.89)
    };
    \addlegendentry{ViT\_B16-10}

\addplot[
    color=blue,
   mark=star,
    dotted
    ]
    coordinates {
    (1,97.59)
    (10,98.28)
    (20,98.58)
    (30,98.67)
    (40,98.73)
    (50,98.97)
    (60,98.94)
    (70,98.82)
    (80,98.98)
    (90,98.94)
    };
    \addlegendentry{ViT\_B16-20}

\addplot[
    color=blue,
    mark=star,
    ]
    coordinates {
    (1,97.39)
    (10,98.25)
    (20,98.63)
    (30,98.59)
    (40,98.96)
    (50,98.76)
    (60,98.98)
    (70,98.82)
    (80,98.94)
    (90,98.95)
    };
    \addlegendentry{ViT\_B16-50}

\addplot[
    color=red,
    mark=star,
    dashed
    ]
    coordinates {
    (1,92.31)
    (10,96.67)
    (20,97.28)
    (30,97.49)
    (40,97.69)
    (50,97.82)
    (60,97.92)
    (70,97.74)
    (80,97.90)
    (90,97.67)
    };
    \addlegendentry{ResNet50-10}

\addplot[
    color=red,
    mark=star,
    dotted
    ]
    coordinates {
    (1,93.00)
    (10,96.70)
    (20,97.16)
    (30,97.52)
    (40,97.55)
    (50,97.59)
    (60,97.80)
    (70,97.66)
    (80,97.67)
    (90,97.76)
    };
    \addlegendentry{ResNet50-20}

\addplot[
    color=red,
    mark=star,
    ]
    coordinates {
    (1,93.14)
    (10,96.38)
    (20,97.07)
    (30,97.25)
    (40,97.59)
    (50,97.55)
    (60,97.56)
    (70,97.69)
    (80,97.56)
    (90,97.39)
    };
    \addlegendentry{ResNet50-50}
    \end{axis}
    \end{tikzpicture}%
    ~%
    \begin{tikzpicture}[ scale=0.68]
    \begin{axis}[ xlabel={Time [Mins]},
    ylabel={Accuracy [\%]},
    xmin=0, xmax=245,
    ymin=90, ymax=100,
    xtick={5, 35, 65, 95, 125, 155, 185, 215, 245},
    ytick={90, 92, 94, 96, 98, 100},
    legend pos=south east,
    ymajorgrids=true,
    grid style=dashed,
    line width=1pt,    
    ]
\addplot[
    color=blue,
    mark=star,
    dashed
    ]
    coordinates {
    (30.38,97.49)
    (54.29,98.25)
    (75.67,98.64)
    (99.90,98.69)
    (125.13,98.87)
    (148.45,98.92)
    (171.45,98.99)
    (196.18,98.94)
    (217.94,98.86)
    (240.75,98.89)
    };
    \addlegendentry{ViT\_B16-10}

\addplot[
    color=blue,
   mark=star,
    dotted
    ]
    coordinates {
    (16.49,97.59)
    (39.01,98.28)
    (60.52,98.58)
    (87.05,98.67)
    (109.69,98.73)
    (132.09,98.97)
    (156.75,98.94)
    (181.35,98.82)
    (201.95,98.98)
    (227.10,98.94)
    };
    \addlegendentry{ViT\_B16-20}

\addplot[
    color=blue,
    mark=star,
    ]
    coordinates {
    (8.02,97.39)
    (29.24,98.25)
    (52.25,98.63)
    (78.51,98.59)
    (100.47,98.96)
    (122.60,98.76)
    (147.65,98.98)
    (170.78,98.82)
    (193.10,98.94)
    (217.38,98.95)
    };
    \addlegendentry{ViT\_B16-50}

\addplot[
    color=red,
    mark=star,
    dashed
    ]
    coordinates {
    (14.30,92.31)
    (27.07,96.67)
    (42.39,97.28)
    (55.72,97.49)
    (72.94,97.69)
    (84.95,97.82)
    (102.69,97.92)
    (118.97,97.74)
    (130.57,97.90)
    (152.17,97.67)
    };
    \addlegendentry{ResNet50-10}

\addplot[
    color=red,
    mark=star,
    dotted
    ]
    coordinates {
    (7.87,93.00)
    (20.67,96.70)
    (34.51,97.16)
    (49.19,97.52)
    (64.51,97.55)
    (77.55,97.59)
    (96.62,97.80)
    (113.80,97.66)
    (118.59,97.67)
    (144.45,97.76)
    };
    \addlegendentry{ResNet50-20}

\addplot[
    color=red,
    mark=star,
    ]
    coordinates {
    (4.13,93.14)
    (17.29,96.38)
    (31.11,97.07)
    (46.30,97.25)
    (61.54,97.59)
    (75.40,97.55)
    (91.03,97.56)
    (107.29,97.69)
    (120.97,97.56)
    (134.64,97.39)
    };
    \addlegendentry{ResNet50-50}
    \end{axis}
    \end{tikzpicture}%
\vspace{-1em}
\caption{Performance scores for ResNet50 and ViT\_B16 for various EDPE values and total time budgets for time, EDPE and performance on CIFAR10 (Method=CRAIG).}
\label{fig:pretrain-cf10}
\end{figure}
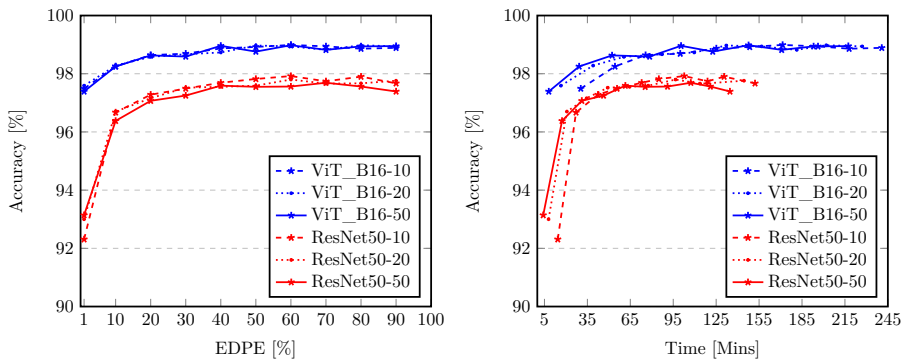

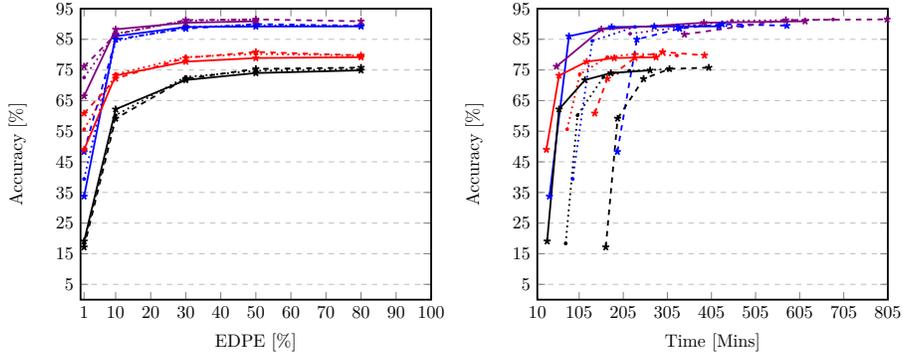
\begin{figure}
        \centering
\begin{tikzpicture}[scale=0.68]
   \begin{axis}[ xlabel={EDPE [\%]},
    ylabel={Accuracy [\%]},
    xmin=0, xmax=100,
    ymin=0, ymax=95,
   xtick={1, 10, 20, 30, 40, 50, 60, 70, 80, 90, 100},
    ytick={5, 15, 25, 35, 45, 55, 65, 75, 85, 95},
    legend pos=south east,
    ymajorgrids=true,
    grid style=dashed,
    line width=1pt,
    legend style={font=\tiny}
    ]
 \addplot[
    color=blue,
    mark=star,
    dashed,
    ]
    coordinates {
    (1,48.30)
    (10, 84.97)
    (30,88.48)
    (50,89.97)
    (80,89.46)
    };
    \label{vit10}

\addplot[
    color=blue,
    mark=star,
    dotted,
    ]
    coordinates {
    (1,39.39)
    (10, 84.46)
    (30,89.05)
    (50,89.59)
    (80,89.28)
    };
    \label{vit20}

\addplot[
    color=blue,
    mark=star,
    ]
    coordinates {
    (1,33.73)
    (10, 86.00)
    (30,89.01)
    (50,89.18)
    (80,89.23)
    };
    \label{vit50}

\addplot[
    color=red,
    mark=star,
    dashed,
    ]
    coordinates {
    (1,60.83)
    (10, 72.21)
    (30,78.94)
    (50,80.87)
    (80,79.76)
    };
    \label{r10}

\addplot[
    color=red,
    mark=star,
    dotted,
    ]
    coordinates {
    (1,55.63)
    (10, 73.59)
    (30,79.19)
    (50,80.23)
    (80,79.70)
    };
    \label{r20}

\addplot[
    color=red,
    mark=star,
    ]
    coordinates {
    (1,49.00)
    (10, 73.18)
    (30,77.71)
    (50,78.89)
    (80,79.16)
    };
    \label{r50}

\addplot[
    color=black,
    mark=star,
    dashed,
    ]
    coordinates {
    (1,17.15)
    (10, 59.15)
    (30,72.10)
    (50,75.35)
    (80,75.76)
    };
    \label{mob10}

\addplot[
    color=black,
    mark=star,
    dotted,
    ]
    coordinates {
    (1,18.36)
    (10, 60.22)
    (30,72.71)
    (50,74.73)
    (80,75.24)
    };
    \label{mob20}

\addplot[
    color=black,
    mark=star,
    ]
    coordinates {
    (1,19.08)
    (10, 62.17)
    (30,71.72)
    (50,73.99)
    (80,74.88)
    };
    \label{mob50}

    \addplot[
    color=violet,
    mark=star,
    dashed,
    ]
    coordinates {
    (1,76.14)
    (10, 86.60)
    (30,91.28)
    (50,91.51)
    (80,90.89)
    };
    \label{s10}

\addplot[
    color=violet,
    mark=star,
    dotted,
    ]
    coordinates {
    (1,72.53)
    (10, 86.83)
    (30,91.05)
    (50,91.49)
    };
    \label{s20}

\addplot[
    color=violet,
    mark=star,
    ]
    coordinates {
    (1,66.49)
    (10, 88.26)
    (30,90.42)
    (50,90.90)
    };
    \label{s50}
    \end{axis}
    \end{tikzpicture}%
    ~%
    \begin{tikzpicture}[ scale=0.68]
    \begin{axis}[ xlabel={Time [Mins]},
    ylabel={Accuracy [\%]},
    xmin=10, xmax=805,
    ymin=0, ymax=95,
    xtick={10, 105, 205, 305, 405, 505, 605, 705, 805},
    ytick={5, 15, 25, 35, 45, 55, 65, 75, 85, 95},
    ymajorgrids=true,
    grid style=dashed,
    line width=1pt,
    ]
\addplot[
    color=blue,
    mark=star,
    dashed,
    ]
    coordinates {
    (191.59,48.30)
    (235.84, 84.97)
    (332.62,88.48)
    (429.41,89.97)
    (575.97,89.46)
    };

\addplot[
    color=blue,
    mark=star,
    dotted,
    ]
    coordinates {
    (90.30,39.39)
    (134.54, 84.46)
    (231.33,89.05)
    (328.12,89.59)
    (474.68,89.28)
    };

\addplot[
    color=blue,
    mark=star,
    ]
    coordinates {
    (37.50,33.73)
    (81.75, 86.00)
    (178.53,89.01)
    (275.32,89.18)
    (421.88,89.23)
    };

\addplot[
    color=red,
    mark=star,
    dashed,
    ]
    coordinates {
    (140.30,60.83)
    (168.94, 72.21)
    (231.58,78.94)
    (294.23,80.87)
    (389.09,79.76)
    };

\addplot[
    color=red,
    mark=star,
    dotted,
    ]
    coordinates {
    (77.69,55.63)
    (106.33, 73.59)
    (168.98,79.19)
    (231.62,80.23)
    (326.49,79.70)
    };

\addplot[
    color=red,
    mark=star,
    ]
    coordinates {
    (30.60,49.00)
    (59.24, 73.18)
    (121.89,77.71)
    (184.53,78.89)
    (279.40,79.16)
    };

\addplot[
    color=black,
    mark=star,
    dashed,
    ]
    coordinates {
    (165.27,17.15)
    (192.15, 59.15)
    (250.97,72.10)
    (309.78,75.35)
    (398.84,75.76)
    };

\addplot[
    color=black,
    mark=star,
    dotted,
    ]
    coordinates {
    (74.34,18.36)
    (101.23, 60.22)
    (160.04,72.71)
    (218.85,74.73)
    (307.91,75.24)
    };

\addplot[
    color=black,
    mark=star,
    ]
    coordinates {
    (32.29,19.08)
    (59.18, 62.17)
    (117.99,71.72)
    (176.80,73.99)
    (265.86,74.88)
    };

    \addplot[
    color=violet,
    mark=star,
    dashed,
    ]
    coordinates {
    (343.56,86.60)
    (573.72, 91.28)
    (803.24,91.51)
    };

\addplot[
    color=violet,
    mark=star,
    dotted,
    ]
    coordinates {
    (220.54,86.83)
    (451.32, 91.05)
    (680.85,91.49)
    };

\addplot[
    color=violet,
    mark=star,
    ]
    coordinates {
    (53.38, 76.14)
    (155.19,88.26)
    (387.84, 90.42)
    (617.37,90.90)
    };
    \end{axis}
    \end{tikzpicture}%
\vspace{-1em}
\caption{Performance scores on TinyImagenet at SSI values of 10, 20 and 50 for pretrained CNN and transformer models (Method=CRAIG). (ViT\_B16-10 (\ref{vit10}), ViT\_B16-20 (\ref{vit20}), ViT\_B16-50 (\ref{vit50}), ResNet50-10 (\ref{r10}), ResNet50-20 (\ref{r20}), ResNet50-50 (\ref{r50}), MobileNetV3-10 (\ref{mob10}), MobileNetV3-20 (\ref{mob20}), MobileNetV3-50 (\ref{mob50}), Swin-10 (\ref{s10}), Swin-20 (\ref{s20}), and Swin-50 (\ref{s50}))}
\label{fig:pretrain-tim}
\end{figure}

\textbf{Data subselection under pretraining. } In this section, we investigate how pretraining (ImageNet21k) impacts model performance using coreset. Fig. \ref{fig:pretrain-cf10} shows the performance scores, and we use here CRAIG as the coreset selection method. A stable performance by both models can be seen, interestingly even when EDPE is as low as 10\% both models perform similarly. Furthermore, ViT\_B16 is consistently better than the ResNet50 model. We also see that the performance trend is consistent for $SSI$ values of 10, 20 and 50. For very low EDPE values, a significant drop in the performance of the CNN models can be seen (Fig. \ref{fig:pretrain-cf10}). Overall, the stable performance for both the architectures at even very small coreset sizes as well as time budgets reveals that the pretraining on Imagenet21K serves as a very strong prior and not many samples of CIFAR10 are needed to further amplify the discriminative power of these models. 


Results on TinyImageNet classification are shown in \mbox{Fig. \ref{fig:pretrain-tim}}. We use all 4 architectures (CNN and transformer) and present scores for $SSI$ values of 10, 20 and 50. It can be seen that $SSI$, in general, does not impact model performance. Among the used models, the Swin transformer seems the most robust even at an EDPE value of 1\%, whereas others deteriorate significantly. However, Swin Transformer is computationally expensive. Moreover, ViT\_16 outperforms the CNN models at all coreset sizes. When compared with ResNet50, we see that at extremely low subsets, ResNet50 performs better than ViT\_B16, thus being more stable than transformers at this low extent of subset selection. Finally, we observe, when pretrained, transformer models are stable in performance across different coreset sizes.



\begin{figure}
        \centering
\begin{tikzpicture}[scale=0.68]
   \begin{axis}[ xlabel={EDPE [\%]},
    ylabel={Quadratic $\kappa$},
    xmin=0, xmax=100,
    ymin=0.65, ymax=0.95,
   xtick={1, 10, 20, 30, 40, 50, 60, 70, 80, 90, 100},
    ytick={0.65, 0.7, 0.75, 0.8, 0.85, 0.9, 0.95},
    legend pos=north east,
    ymajorgrids=true,
    grid style=dashed,
    line width=1pt,
    ]
\addplot[
    color=blue,
    mark=star,
    dashed
    ]
    coordinates {
    (1,0.74)  
    (10,0.83)  
    (30,0.87)
    (50,0.87)
    (80,0.88) 
    };
    \label{ViTcraig}

\addplot[
    color=red,
    mark=star,
    dashed
    ]
    coordinates {
    (1,0.68)
    (10,0.84)
    (30,0.87) 
    (50,0.87)
    (80,0.89) 
    };
    \label{ViTgrad}

    \addplot[
    color=violet,
    mark=star,
    dashed
    ]
    coordinates {
    (1,0.70) 
    (10,0.82)
    (30,0.84)
    (50,0.88) 
    (80,0.89)
    };
    \label{ViTglis}

    \addplot[
    color=black,
    mark=star,
    dashed
    ]
    coordinates {
    (1,0.73) 
    (10,0.85)
    (50,0.88) 
    (80,0.86) 
    };
    \label{ViTrand}

    \addplot[
    color=blue,
    mark=star,
    ]
    coordinates {
    (1,0.79)
    (10,0.83) 
    (30,0.86)
    (50,0.88) 
    (80,0.90) 
    };
    \label{rescraig}

    \addplot[
    color=red,
    mark=star,
    ]
    coordinates {
    (1,0.76) 
    (8.92,0.85) 
    (30, 0.87)
    (50,0.89) 
    (80,0.90) 
    };
    \label{resgrad}

    \addplot[
    color=violet,
    mark=star,
    ]
    coordinates {
    (1,0.72) 
    (10,0.82)
    (30,0.85)
    (50,0.88) 
    (80,0.89) 
    
    };
    \label{resglis}

    \addplot[
    color=black,
    mark=star,
    ]
    coordinates {
    (1,0.80) 
    (10,0.85) 
    (30,0.86)
    (50,0.88) 
    (80,0.89) 
    };
    \label{resrand}
    \end{axis}
    \end{tikzpicture}%
    ~%
    \begin{tikzpicture}[ scale=0.68]
    \begin{axis}[ xlabel={Time [Mins]},
    ylabel={Quadratic $\kappa$},
    xmin=0, xmax=12,
    ymin=0.65, ymax=0.95,
    xtick={0, 3, 6, 9, 12},
    ytick={0.65, 0.7, 0.75, 0.8, 0.85, 0.9, 0.95},
    legend pos=south east,
    ymajorgrids=true,
    grid style=dashed,
    line width=1pt,    
    ]
    
   \addplot[
    color=blue,
    mark=star,
    dashed
    ]
    coordinates {
    (1.75,0.74)  
    (3.16,0.83)  
    (4.94,0.87)
    (7.70,0.87)
    (10.99,0.88)
    };

\addplot[
    color=red,
    mark=star,
    dashed
    ]
    coordinates {
    (1.61,0.68)
    (3.30,0.84)
    (5.07,0.87) 
    (7.84,0.88) 
    (11.13,0.89) 
    };

    \addplot[
    color=violet,
    mark=star,
    dashed
    ]
    coordinates {
    (2.14,0.70) 
    (3.55,0.82)
    (5.33,0.84)
    (8.09,0.88) 
    (11.38,0.89) 
    };

    \addplot[
    color=black,
    mark=star,
    dashed
    ]
    coordinates {
    (0.17,0.73) 
    (2.07,0.85)
    (3.29,0.86) 
    (6.05,0.88) 
    (9.34,0.86)
    };

    \addplot[
    color=blue,
    mark=star,
    ]
    coordinates {
    (1.65,0.79)
    (2.56,0.83) 
    (3.71,0.86)
    (5.50,0.88) 
    (7.63,0.90) 
    };

    \addplot[
    color=red,
    mark=star,
    ]
    coordinates {
    (1.80,0.76) 
    (2.71,0.85) 
    (3.86, 0.87) 
    (5.65,0.89) 
    (7.78,0.90) 
    };
    \label{resgrad}

    \addplot[
    color=violet,
    mark=star,
    ]
    coordinates {
    (3.08,0.72) 
    (4.00,0.82)
    (5.15,0.85) 
    (6.94,0.88) 
    (9.07,0.89) 
    
    };

    \addplot[
    color=black,
    mark=star,
    ]
    coordinates {
    (0.11,0.80) 
    (1.34,0.85) 
    (2.13,0.86) 
    (3.92,0.88)
    (6.05,0.89) 
    };
    \end{axis}
    \end{tikzpicture}%
\vspace{-1em}
\caption{Performance on APTOS-2019 classification task as a function of EDPE and total time for pretrained ResNet50 and ViT\_B16 for the 4 coreset selection methods (SSI=20). (ViT\_B16-Craig (\ref{ViTcraig}), ViT\_B16-GradMatch (\ref{ViTgrad}), ViT\_B16-GLISTER (\ref{ViTglis}), ViT\_B16-Random (\ref{ViTrand}), ResNet50-Craig (\ref{rescraig}), ResNet50-GradMatch (\ref{resgrad}), ResNet50-GLISTER (\ref{resglis}), ResNet50-Random (\ref{resrand}))}
\label{fig:ap-coresets}
\end{figure}
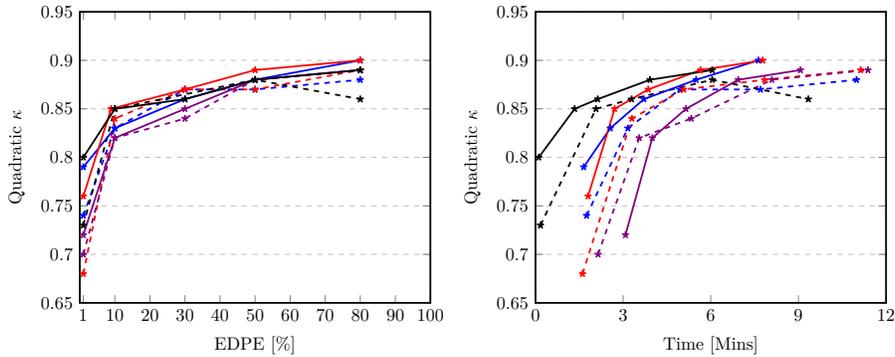

\textbf{Performance on the medical dataset. } We extend our experiments to non-natural images. Fig. \ref{fig:ap-coresets}, illustrates results on the APTOS-2019 blindness detection dataset for ResNet50 and ViT\_B16. It can be seen that the ResNet50 performs better than ViT\_B16 at all coreset sizes, thereby confirming that the heavy pretraining in this case is not helping the transformers. We see that for CNN, random coreset selection is better or almost at par with GradMatch which itself performs better than the other coreset methods. For low coreset size (EDPE), random selection seems to be the most stable solution. Interestingly, when we look at the computational time, including the coreset selection time, random selection outperforms the other methods. For larger time investments, GradMatch is better than the random choice, however, for lower time budgets, random selection is the preferred choice.



\begin{figure}
        \centering
\begin{tikzpicture}[scale=0.68]
   \begin{axis}[ xlabel={EDPE [\%]},
    ylabel={Accuracy [\%]},
    xmin=0, xmax=100,
    ymin=3, ymax=65,
   xtick={1, 10, 20, 30, 40, 50, 60, 70, 80, 90, 100},
    ytick={0, 3, 15, 25, 35, 45, 55, 65},
    legend pos=north east,
    ymajorgrids=true,
    grid style=dashed,
    line width=1pt,
    ]
\addplot[
    color=blue,
    mark=star,
    dashed
    ]
    coordinates {
    (1,4.42)
    (10,8.75)
    (30,15.21)
    (50,22.96)
    (80,34.50)
    (100,40.35)
    };
    \label{ViTcraig}

\addplot[
    color=red,
    mark=star,
    dashed
    ]
    coordinates {
    (1,4.14)
    (10,7.53)
    (30,14.78)
    (50,22.82)
    (80,33.21)
    (100,40.35)
    };
    \label{ViTgrad}

    \addplot[
    color=violet,
    mark=star,
    dashed
    ]
    coordinates {
    (1,3.96)
    (10,7.46)
    (30,15.64)
    (50,24.28)
    (80,35.42)
    (100,40.35)
    };
    \label{ViTglis}

    \addplot[
    color=black,
    mark=star,
    dashed
    ]
    coordinates {
    (1,3.96)
    (10,7.50)
    (30,13.03)
    (50,18.92)
    (80,33.89)
    (100,40.35)
    };
    \label{ViTrand}

    \addplot[
    color=blue,
    mark=star,
    ]
    coordinates {
    (1,5.50)
    (10,13.00)
    (30,21.39)
    (50,34.25)
    (80,51.78)
    (100,61.28)
    };
    \label{rescraig}

    \addplot[
    color=red,
    mark=star,
    ]
    coordinates {
    (1,6.67)
    (10,13.28)
    (30, 22.46)
    (50,38.07)
    (80,51.46)
    (100,61.28)
    };
    \label{resgrad}

    \addplot[
    color=violet,
    mark=star,
    ]
    coordinates {
    (1,7.82)
    (10,13.64)
    (30,22.78)
    (50,34.85)
    (80,53.00) 
    (100,61.28)
    
    };
    \label{resglis}

    \addplot[
    color=black,
    mark=star,
    ]
    coordinates {
    (1,6.89)
    (10,12.89)
    (30,19.17)
    (50,28.75)
    (80,48.53) 
    (100,61.28)
    };
    \label{resrand}
    \end{axis}
    \end{tikzpicture}%
    ~%
    \begin{tikzpicture}[ scale=0.68]
    \begin{axis}[ xlabel={Time [Mins]},
    ylabel={Accuracy [\%]},
    xmin=0, xmax=285,
    ymin=3, ymax=65,
    xtick={5, 35, 65, 95, 125, 155, 185, 215, 245, 285},
    ytick={3, 15, 25, 35, 45, 55, 65},
    legend pos=south east,
    ymajorgrids=true,
    grid style=dashed,
    line width=1pt,    
    ]
    
    \addplot[
    color=blue,
    mark=star,
    dashed
    ]
    coordinates {
    (91.98,4.42)
    (102.86,8.75)
    (151.86,15.21)
    (200.85,22.96)
    (271.63,34.50)
    (239.53,40.35)
    };

\addplot[
    color=red,
    mark=star,
    dashed
    ]
    coordinates {
    (53.60,4.14)
    (64.49,7.53)
    (113.49,14.78)
    (162.48,22.82)
    (233.25,33.21)
    (239.53,40.35)
    };

    \addplot[
    color=violet,
    mark=star,
    dashed
    ]
    coordinates {
    (88.53,3.96)
    (99.42,7.46)
    (148.41,15.64)
    (197.41,24.28)
    (268.18,35.42)
    (239.53,40.35)
    };

    \addplot[
    color=black,
    mark=star,
    dashed
    ]
    coordinates {
    (10.88,3.96)
    (21.77,7.50)
    (70.77,13.03)
    (119.76,18.92)
    (190.53,33.89)
    (239.53,40.35)
    };

    \addplot[
    color=blue,
    mark=star,
    ]
    coordinates {
    (28.81,5.50)
    (34.03,13.00)
    (57.55,21.39)
    (81.07,34.25)
    (115.05,51.78)
    (114.99,61.28)
    };

    \addplot[
    color=red,
    mark=star,
    ]
    coordinates {
    (22.96,6.67)
    (28.19,13.28)
    (51.71, 22.46)
    (75.23,38.07)
    (109.21,51.46)
    (114.99,61.28)
    };

    \addplot[
    color=violet,
    mark=star,
    ]
    coordinates {
    (24.96,7.82)
    (30.19,13.64)
    (53.71,22.78)
    (77.23,34.85)
    (111.20,53.00) 
    (114.99,61.28)
    
    };

    \addplot[
    color=black,
    mark=star,
    ]
    coordinates {
    (5.22,6.89)
    (10.45,12.89)
    (33.97,19.17)
    (57.49,28.75)
    (91.47,48.53) 
    (114.99,61.28)
    };
    \end{axis}
    \end{tikzpicture}%
\vspace{-1em}
\caption{Performance comparison between ImageNet-21k weight initialized \vit and ResNet50 for time, EDPE and performance on UltraMNIST with different coreset methods, SSI=20 such that ViT\_B16-Craig (\ref{ViTcraig}), ViT\_B16-GradMatch (\ref{ViTgrad}), ViT\_B16-GLISTER (\ref{ViTglis}), ViT\_B16-Random (\ref{ViTrand}), ResNet50-Craig (\ref{rescraig}), ResNet50-GradMatch (\ref{resgrad}), ResNet50-GLISTER (\ref{resglis}), ResNet50-Random (\ref{resrand})}
\label{fig:um-coresets}
\end{figure}
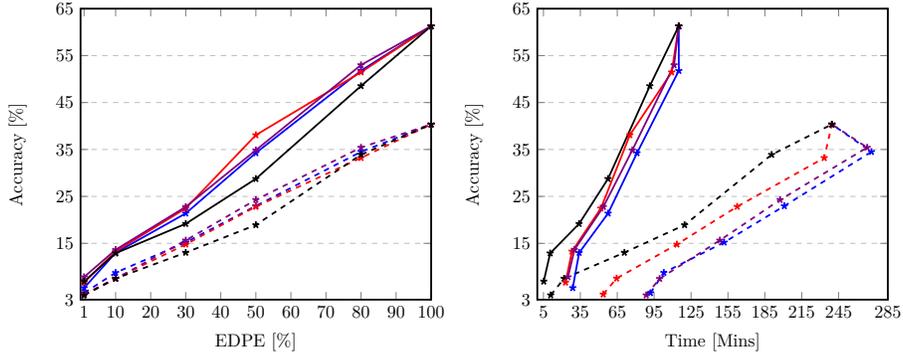

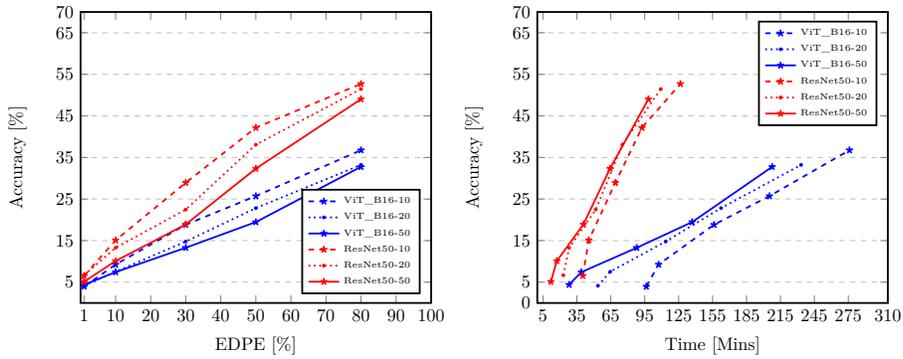
\begin{figure}
        \centering
\begin{tikzpicture}[scale=0.68]
   \begin{axis}[ xlabel={EDPE [\%]},
    ylabel={Accuracy [\%]},
    xmin=0, xmax=100,
    ymin=0, ymax=70,
   xtick={1, 10, 20, 30, 40, 50, 60, 70, 80, 90, 100},
    ytick={5, 15, 25, 35, 45, 55, 65, 70},
    legend pos=south east,
    ymajorgrids=true,
    grid style=dashed,
    line width=1pt,
    legend style={font=\tiny}
    ]
 \addplot[
    color=blue,
    mark=star,
    dashed
    ]
    coordinates {
    (1,3.96)
    (10,9.25)
    (30,18.82)
    (50,25.71)
    (80,36.75)
    };
    \addlegendentry{ViT\_B16-10}

\addplot[
    color=blue,
   mark=star,
    dotted
    ]
    coordinates {
    (1,4.14)
    (10,7.53)
    (30,14.78)
    (50,22.82)
    (80,33.21)
    };
    \addlegendentry{ViT\_B16-20}

\addplot[
    color=blue,
    mark=star,
    ]
    coordinates {
    (1,4.39)
    (10,7.42)
    (30,13.28)
    (50,19.46)
    (80,32.75)
    };
    \addlegendentry{ViT\_B16-50}

\addplot[
    color=red,
    mark=star,
    dashed
    ]
    coordinates {
    (1,6.53)
    (10,15.10)
    (30,28.92)
    (50,42.17)
    (80,52.71)
    };
    \addlegendentry{ResNet50-10}

\addplot[
    color=red,
    mark=star,
    dotted
    ]
    coordinates {
    (1,6.67)
    (10,13.28)
    (30,22.46)
    (50,38.07)
    (80,51.46)
    };
    \addlegendentry{ResNet50-20}

\addplot[
    color=red,
    mark=star,
    ]
    coordinates {
    (1,5.10)
    (10,10.10)
    (30,18.89)
    (50,32.32)
    (80,49.03)
    };
    \addlegendentry{ResNet50-50}
    \end{axis}
    \end{tikzpicture}%
    ~%
    \begin{tikzpicture}[ scale=0.68]
    \begin{axis}[ xlabel={Time [Mins]},
    ylabel={Accuracy [\%]},
    xmin=0, xmax=310,
    ymin=0, ymax=70,
    xtick={5, 35, 65, 95, 125, 155, 185, 215, 245, 275, 310},
    ytick={0, 5, 15, 25, 35, 45, 55, 65, 70},
    legend pos=north east,
    ymajorgrids=true,
    grid style=dashed,
    line width=1pt,
    legend style={font=\tiny}
    ]
\addplot[
    color=blue,
    mark=star,
    dashed
    ]
    coordinates {
    (96.41,3.96)
    (107.30,9.25)
    (156.29,18.82)
    (205.29,25.71)
    (276.06,36.75)
    
    };
    \addlegendentry{ViT\_B16-10}

\addplot[
    color=blue,
   mark=star,
    dotted
    ]
    coordinates {
    (53.60,4.14)
    (64.49,7.53)
    (113.49,14.78)
    (162.48,22.82)
    (233.25,33.21)
    };
    \addlegendentry{ViT\_B16-20}

\addplot[
    color=blue,
    mark=star,
    ]
    coordinates {
    (28.05,4.39)
    (38.94,7.42)
    (87.94,13.28)
    (136.93,19.46)
    (207.70,32.75)
    };
    \addlegendentry{ViT\_B16-50}

\addplot[
    color=red,
    mark=star,
    dashed
    ]
    coordinates {
    (40.29,6.53)
    (45.51,15.10)
    (69.03,28.92)
    (92.55,42.17)
    (126.53,52.71)
    };
    \addlegendentry{ResNet50-10}

\addplot[
    color=red,
    mark=star,
    dotted
    ]
    coordinates {
    (22.96,6.67)
    (28.19,13.28)
    (51.71,22.46)
    (75.23,38.07)
    (109.21,51.46)
    };
    \addlegendentry{ResNet50-20}

\addplot[
    color=red,
    mark=star,
    ]
    coordinates {
    (12.02,5.10)
    (17.24,10.10)
    (40.76,18.89)
    (64.29,32.32)
    (98.26,49.03)
    };
    \addlegendentry{ResNet50-50}
    \end{axis}
    \end{tikzpicture}%
\vspace{-1em}
\caption{Performance comparison between pretrained ViT\_B16 and ResNet50 for time, EDPE on UltraMNIST dataset (Method=GradMatch).}
\label{fig:um-ssi}
\end{figure}

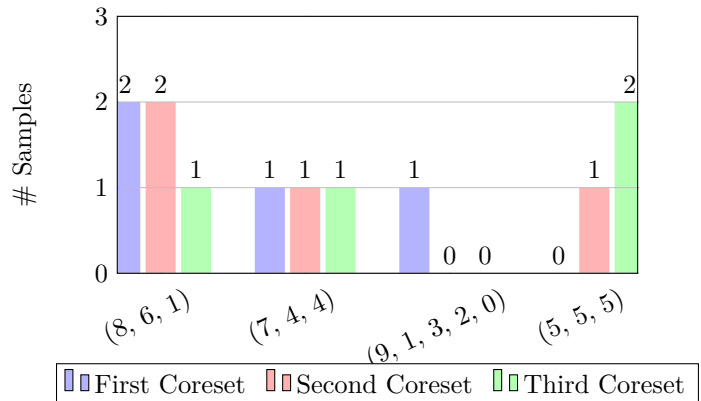
\begin{figure}
    \centering
    \begin{tikzpicture}[scale=1.0]
  \begin{axis}[
        ybar, axis on top,
        height=5cm, width=8.5cm,
        bar width=0.4cm,
        ymajorgrids, tick align=inside,
        ymin=0, ymax=3,
        tickwidth=0pt,
        legend style={
            at={(0.5,-0.35)},
            anchor=north,
            legend columns=-1,
            /tikz/every even column/.append style={column sep=0.2cm}
        },
        ylabel={\# Samples},
        symbolic x coords={
           {(8, 6, 1)},
           {(7, 4, 4)},
           {(9, 1, 3, 2, 0)},
          {(5, 5, 5)}},
       xtick=data,
       xticklabel style={xshift=-5pt, rotate=25},
       nodes near coords={
        \pgfmathprintnumber[precision=0]{\pgfplotspointmeta}
       } 
    ]
    \addplot [draw=none, fill=blue!30] coordinates {
      ({(8, 6, 1)}, 2) 
      ({(7, 4, 4)},1) 
      ({(9, 1, 3, 2, 0)},1)
      ({(5, 5, 5)},0) 
      };
   \addplot [draw=none,fill=red!30] coordinates {
      ({(8, 6, 1)},2) 
      ({(7, 4, 4)},1) 
      ({(9, 1, 3, 2, 0)},0)
      ({(5, 5, 5)},1) 
      };
   \addplot [draw=none, fill=green!30] coordinates {
      ({(8, 6, 1)},1) 
      ({(7, 4, 4)},1) 
      ({(9, 1, 3, 2, 0)},0)
      ({(5, 5, 5)},2) 
      };

    \legend{First Coreset,Second Coreset,Third Coreset}
  \end{axis}
  \end{tikzpicture}
  \vspace{-0.8em}
    \caption{Number of samples present in different combinations of class 15 of UltraMNIST dataset in each coreset after coreset selection is done using GradMatch method.}
    \label{fig:bar-plot}
\end{figure}

\begin{figure*}[t]
    \centering
    \includegraphics[scale=0.11, trim=0 4.5cm 0 4.5cm, clip=true]{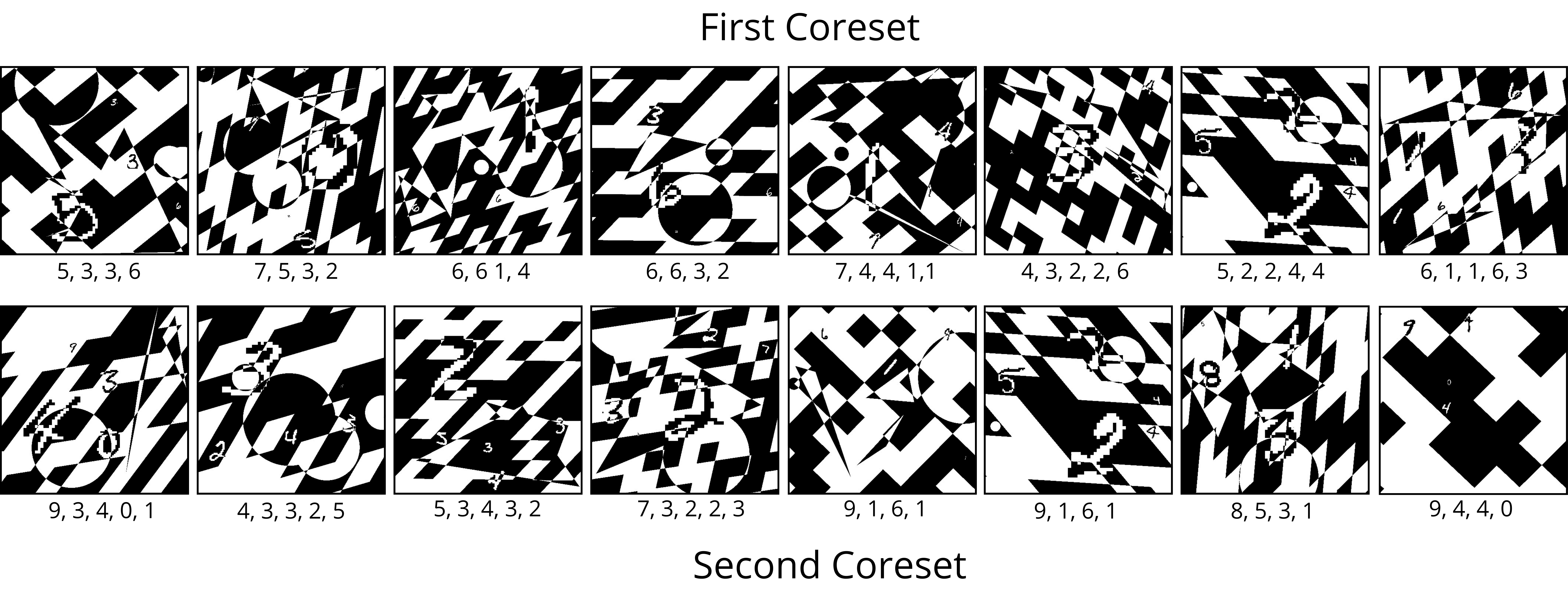}
    \vspace{-1em}
    \caption{UltraMNIST samples for class label 17 from the two different coresets (top and bottom) for EDPE=1\% obtained with GradMatch. Note that there is no overlap between the two coresets in terms of the combination of the digits in each image.}
    \label{fig:um-samples}
\end{figure*}

\textbf{Learning semantic coherency in non-natural images. }

We analyze here how well transformers can learn the semantic coherency of the dataset, given that the data comprises non-natural images (very different from those used in pretraining).

First, we study how the performance of the models is affected using different coreset selection methods. Fig. \ref{fig:um-coresets} presents the results for ViT\_B16 and ResNet50. We see that independent of the coreset method, the CNN model is always superior to the transformer model across all coreset sizes. This observation is consistent across the various values of SSI (Fig. \ref{fig:um-ssi}). UltraMNIST is a hard dataset to learn, and it is possible that with no similar pretraining, it is hard for transformers to learn well the desired semantic context.

In terms of the coreset size (EDPE), GradMatch outperforms the other methods while the random selection of coresets seems to perform the worst. However, in terms of the total time spent, random subset selection outperforms the other methods. This clearly indicates that although GradMatch might be a better selector of useful samples, the time spent scrutinizing the samples is not worth spending when compared to a random selection approach. For the SSI values shown in Fig. \ref{fig:um-ssi}, we see that when evaluating performance with respect to EDPE, more number of coreset selection intervals are favoured, however, when evaluating in terms of the time, lesser selection cycles are preferred. This is because the coreset selection time is significant for most methods, however, EDPE does not take that into account.

We also look at why the performance of CNNs and transformer models is extremely low when a coreset size of as low as 1\% is chosen. Our investigation reveals that this bottleneck is imposed by the fact that we conventionally choose an equal number of samples per class of the data. However, the complexity of the distribution of each class in UltraMNIST is very different. For example, class label 0 can only be formed by the combination of 3-5 occurrences of 0. However, 15 can be formed by many combinations of the digits. When selecting a coreset for this class, we tend to miss some of the combinations, thereby limiting the discriminative power of the trained model for such instances. This is also evident from the distributions shown in Fig. \ref{fig:bar-plot}. There are certain combinations of 15 that miss from one coreset to another. Clearly, the model will catastrophically forget them as soon as they are completely out of the coresets. Interestingly, this is not the worst. If the distributions of the classes are not similar, uniform sampling could lead to a situation, where the distribution of the hard class could change completely between two coresets. Fig. \ref{fig:um-samples} shows the training samples of class 17 from two different coresets of EDPE = 1\%. As can be seen, the two sets are completely different which implies that it would be hard for the model to generalize for this class for very small coreset sizes. Overall, the takeaway is that the complexity of the distribution per class should be taken into consideration when constructing the coresets.

\subsection{Discussion}
In this section, we summarize the insights from experiments and present answers to some intriguing questions.

\textit{Should random coreset selection be a preferred choice? }Our results reveal that when comparing performance in terms of time budget, the random selection of the coresets is a more time-efficient choice than the other coreset selection methods. Clearly, one of the factors in favour of random selection is that the coreset selection time is almost zero. From a practical point of view, the impact of the small gains observed by other coreset selection methods would be diminished if the associated subset selection time is significant. Thus, unless the coreset method is very efficient or parallelizable to the extent that the selection time can be made very small, random coreset selection should be a preferred choice for training models in a data-efficient manner.

\textit{Choosing the right subset selection interval (SSI). }There is no right SSI value, but one could optimize this choice based on the available resources. It is evident that when measured purely in terms of EDPE, coreset selection methods perform better than a random choice, and it can be expected that the more the selection frequency, the better the model performance. However, so far, we have seen that the computational time associated with coreset selection is not negligible, and as long as the overall time is below the desired threshold, the choice of SSI is good.

\textit{Should coresets be chosen uniformly across the classes?}
We argue that when constructing the coresets, methods should also focus on constructing the approximate distribution of every class. This would require sampling more samples from the class that is more complex while sparse sampling from the easier one. This way, the small coreset budget could be used in a more appropriate. While this could raise the challenges of class imbalance among others, we still believe that coresets should not be chosen uniformly across the classes, but in a more adaptive manner.

\section{Conclusion}
 We have presented a systematic benchmarking scheme that facilitates a fair comparison of different coreset selection methods for data-efficient training. A thorough comparison of the various coreset methods on model performance at different coreset sizes is discussed. 
We demonstrated that the conventional concept of uniform subset sampling across the various classes of data is not the appropriate choice. Rather samples should be adaptively chosen based on the complexity of the data distribution for each class. We further performed a comparative study between CNNs and transformers and demonstrated coreset selection scenarios where transformers outperform CNNs and vice versa.
Overall, we hope that the insights presented in this paper provide a better understanding of the usage of various coreset selection methods with CNNs and transformers, thereby helping to contribute better towards developing energy-efficient systems for model training.



\section*{Acknowledgements}
Authors would like to thank Texmin Foundation for the supported provided through the research grant under Project PSF-IH-1Y-022.

\clearpage
\appendix
\section{Appendix}
\subsection{Introduction}
We have also provided quantitative results, further illustrating the observations and analysis presented in the main manuscript. 

Moreover, to ease reproducibility and promote further research, we have also shared our code base, which can be accessed using the following link \url{https://github.com/transmuteAI/Data-Efficient-Transformers}. 



\subsection{Results}

\subsubsection{CRAIG Method Results on CIFAR10}
We initially started our experiments on the CIFAR10 dataset. Tables [\ref{tab:vit-c10-craig-pre}, \ref{tab:vit-c10-craig-rand}, \ref{tab:r50-c10-craig-pre}, \ref{tab:r50-c10-craig-rand}] gives a comparative study for these models. We used Fig.\ref{fig:rand-train} and Fig. \ref{fig:pretrain-cf10} to demonstrate that CNN underperforms across all coreset sizes than transformers under the pretraining setting. We can also see that at tiny coreset sizes, transformers are more stable than CNNs for the CIFAR10 dataset.   


\begin{table*}[]
\centering
\resizebox{0.9\textwidth}{!}{%
\begin{tabular}{@{}cc | ccc | ccc | ccc@{}}
\toprule
\textbf{EDPE (\%)} & \textbf{Epoch} & \multicolumn{3}{c|}{\textbf{Total Time (Mins)}} & \multicolumn{3}{c|}{\textbf{Total Selection Time (Mins)}} & \multicolumn{3}{c}{\textbf{Accuracy}} \\ \midrule

\multirow{3}{*}{100} 
& 105 &  & 147.46 &  &  & - &  &  & 99.01 &  \\
& 89 &  & 125.88 &  &  & - &  &  & 98.61 &  \\
 & 79 &  & 112.06 &  &  & - &  &  & 98.86 &  \\
 \midrule

 \multicolumn{1}{c}{} & \multicolumn{1}{c|}{} & \textit{SSI=10} & \textit{SSI=20} & \textit{SSI=50} & \textit{SSI=10} & \textit{SSI=20} & \textit{SSI=50} & \textit{SSI=10} & \textit{SSI=20} & \textit{SSI=50}\\ \midrule

\multirow{3}{*}{90} 
& 105 & 240.75 & 227.10 & 217.38 & 31.20 & 15.41 & 6.24 & 98.89 & 98.94 & 98.95 \\
 & 89 & 203.89 & 193.08 & 183.40 & 24.96 & 12.33 & 3.12 & 98.86 & 98.89 & 98.87 \\
 & 79 & 180.12 & 169.14 & 162.60 & 21.84 & 9.25 & 3.12 & 98.73 & 98.78 & 98.78 \\
\bottomrule
 
 \end{tabular}%
}
\caption{Performance scores for ViT-B\_16 with ImageNet 21K weights on CIFAR10 and its ....EDPE = Effective data per epoch, method=CRAIG}
\label{tab:vit-c10-craig-pre}
\end{table*}


\begin{table*}[]
\centering
\resizebox{0.9\textwidth}{!}{%
\begin{tabular}{@{}cc | ccc | ccc | ccc@{}}
\toprule
\textbf{EDPE (\%)} & \textbf{Epoch} & \multicolumn{3}{c|}{\textbf{Total Time (Mins)}} & \multicolumn{3}{c|}{\textbf{Total Selection Time (Mins)}} & \multicolumn{3}{c}{\textbf{Accuracy}} \\ \midrule

\multirow{3}{*}{100} 
& 105 &  & 147.46 &  &  & - &  &  & 66.84 &  \\
& 89 &  & 125.88 &  &  & - &  &  & 62.21 &  \\
 & 79 &  & 112.06 &  &  & - &  &  & 64.15 &  \\
 \midrule
 \multicolumn{1}{c}{} & \multicolumn{1}{c|}{} & \textit{SSI=10} & \textit{SSI=20} & \textit{SSI=50} & \textit{SSI=10} & \textit{SSI=20} & \textit{SSI=50} & \textit{SSI=10} & \textit{SSI=20} & \textit{SSI=50}\\ \midrule

\multirow{3}{*}{90} 
& 105 & 240.75 & 227.10 & 217.38 & 31.20 & 15.41 & 6.24 & 66.66 & 66.76 & 67.03 \\
 & 89 & 203.89 & 193.08 & 183.40 & 24.96 & 12.33 & 3.12 & 66.42 & 66.46 & 66.85 \\
 & 79 & 180.12 & 169.14 & 162.60 & 21.84 & 9.25 & 3.12 & 65.47 & 66.16 & 66.29 \\
\bottomrule
 
 \end{tabular}%
}
\caption{Performance scores for ViT-B\_16 with Random Initialization on CIFAR10 and its ....EDPE = Effective data per epoch, method=CRAIG}
\label{tab:vit-c10-craig-rand}
\end{table*}


\begin{table*}[]
\centering
\resizebox{0.9\textwidth}{!}{%
\begin{tabular}{@{}cc | ccc | ccc | ccc@{}}
\toprule
\textbf{EDPE (\%)} & \textbf{Epoch} & \multicolumn{3}{c|}{\textbf{Total Time (Mins)}} & \multicolumn{3}{c|}{\textbf{Total Selection Time (Mins)}} & \multicolumn{3}{c}{\textbf{Accuracy}} \\ \midrule
\

\multirow{3}{*}{100} 
& 105 &  & 65.45 &  &  & - &  &  & 97.96 &  \\
 & 89 &  & 55.87 &  &  & - &  &  & 97.10 &  \\
 & 79 &  & 49.74 &  &  & - &  &  & 97.09 &  \\ \midrule

 \multicolumn{1}{c}{} & \multicolumn{1}{c|}{} & \textit{SSI=10} & \textit{SSI=20} & \textit{SSI=50} & \textit{SSI=10} & \textit{SSI=20} & \textit{SSI=50} & \textit{SSI=10} & \textit{SSI=20} & \textit{SSI=50}\\ \midrule

\multirow{3}{*}{90} 
& 105 & 152.17 & 144.45 & 134.64 & 16.73 & 11.36 & 5.05 & 97.67 & 97.76 & 97.39 \\
 & 89 & 129.03 & 122.73 & 113.18 & 13.39 & 9.09 & 2.52 & 97.67 & 97.70 & 97.33 \\
 & 79 & 114.02 & 107.35 & 100.41 & 11.71 & 6.81 & 2.52 & 97.52 & 97.70 & 97.37 \\
\bottomrule
 
 \end{tabular}%
}
\caption{Performance scores for ResNet50V2 with ImageNet21k weights Initialization on CIFAR10 and its ....EDPE = Effective data per epoch, method=CRAIG}
\label{tab:r50-c10-craig-pre}
\end{table*}

\begin{table*}[]
\centering
\resizebox{0.9\textwidth}{!}{%
\begin{tabular}{@{}cc | ccc | ccc | ccc@{}}
\toprule
\textbf{EDPE (\%)} & \textbf{Epoch} & \multicolumn{3}{c|}{\textbf{Total Time (Mins)}} & \multicolumn{3}{c|}{\textbf{Total Selection Time (Mins)}} & \multicolumn{3}{c}{\textbf{Accuracy}} \\ \midrule
\

\multirow{3}{*}{100} 

& 105 &  & 65.45 &  &  & - &  &  & 88.90 &  \\
 & 89 &  & 55.87 &  &  & - &  &  & 88.64 &  \\
 & 79 &  & 49.74 &  &  & - &  &  & 88.10 &  \\ \midrule

 \multicolumn{1}{c}{} & \multicolumn{1}{c|}{} & \textit{SSI=10} & \textit{SSI=20} & \textit{SSI=50} & \textit{SSI=10} & \textit{SSI=20} & \textit{SSI=50} & \textit{SSI=10} & \textit{SSI=20} & \textit{SSI=50}\\ \midrule

\multirow{3}{*}{90} 
& 105 & 152.17 & 144.45 & 134.64 & 16.73 & 11.36 & 5.05 & 89.59 & 89.46 & 89.64 \\
 & 89 & 129.03 & 122.73 & 113.18 & 13.39 & 9.09 & 2.52 & 89.31 & 89.24 & 89.48 \\
 & 79 & 114.02 & 107.35 & 100.41 & 11.71 & 6.81 & 2.52 & 89.03 & 89.10 & 89.03 \\
\bottomrule
 
 \end{tabular}%
}
\caption{Performance scores for ResNet50V2 with Random Initialization on CIFAR10 and its ....EDPE = Effective data per epoch, method=CRAIG}
\label{tab:r50-c10-craig-rand}
\end{table*}


\begin{table*}[]
\centering
\resizebox{0.9\textwidth}{!}{%
\begin{tabular}{@{}cc | ccc | ccc | ccc@{}}
\toprule
\textbf{EDPE (\%)} & \textbf{Epoch} & \multicolumn{3}{c|}{\textbf{Total Time (Mins)}} & \multicolumn{3}{c|}{\textbf{Total Selection Time (Mins)}} & \multicolumn{3}{c}{\textbf{Accuracy}} \\ \midrule

\multirow{3}{*}{100} 
& 105 &  & 147.46 &  &  & - &  &  & 99.01 &  \\
& 89 &  & 125.88 &  &  & - &  &  & 98.61 &  \\
 & 79 &  & 112.06 &  &  & - &  &  & 98.86 &  \\
 \midrule

 \multicolumn{1}{c}{} & \multicolumn{1}{c|}{} & \textit{SSI=10} & \textit{SSI=20} & \textit{SSI=50} & \textit{SSI=10} & \textit{SSI=20} & \textit{SSI=50} & \textit{SSI=10} & \textit{SSI=20} & \textit{SSI=50}\\ \midrule


\multirow{3}{*}{80} 
& 105 & 209.07 & 193.99 & 183.22 & 30.87 & 15.80 & 5.03 & 99.08 & 99.03 & 99.06 \\
 & 89 & 176.77 & 164.72 & 154.59 & 24.70 & 12.64 & 2.51 & 99.08 & 99.03 & 99.06 \\
 & 79 & 156.05 & 143.92 & 136.95 & 21.61 & 9.48 & 2.51 & 99.04 & 99.03 & 99.06 \\
\bottomrule
 
 \end{tabular}%
}
\caption{Performance scores for ViT-B\_16 with ImageNet 21K weights on CIFAR10 and its ....EDPE = Effective data per epoch, method=GradMatch}
\label{tab:my-table}
\end{table*}


\begin{table*}[]
\centering
\resizebox{0.9\textwidth}{!}{%
\begin{tabular}{@{}cc | ccc | ccc | ccc@{}}
\toprule
\textbf{EDPE (\%)} & \textbf{Epoch} & \multicolumn{3}{c|}{\textbf{Total Time (Mins)}} & \multicolumn{3}{c|}{\textbf{Total Selection Time (Mins)}} & \multicolumn{3}{c}{\textbf{Accuracy}} \\ \midrule

\multirow{3}{*}{100} 
& 105 &  & 147.46 &  &  & - &  &  & 66.84 &  \\
& 89 &  & 125.88 &  &  & - &  &  & 62.21 &  \\
 & 79 &  & 112.06 &  &  & - &  &  & 64.15 &  \\
 \midrule

 \multicolumn{1}{c}{} & \multicolumn{1}{c|}{} & \textit{SSI=10} & \textit{SSI=20} & \textit{SSI=50} & \textit{SSI=10} & \textit{SSI=20} & \textit{SSI=50} & \textit{SSI=10} & \textit{SSI=20} & \textit{SSI=50}\\ \midrule

\multirow{3}{*}{80} 
& 105 & 209.07 & 193.99 & 183.22 & 30.87 & 15.80 & 5.03 & 66.79 & 67.05 & 67.22 \\
 & 89 & 176.77 & 164.72 & 154.59 & 24.70 & 12.64 & 2.51 & 66.33 & 66.70 & 66.76 \\
 & 79 & 156.05 & 143.92 & 136.95 & 21.61 & 9.48 & 2.51 & 65.92 & 65.45 & 66.08 \\ \midrule

 \multirow{3}{*}{50} 
 & 105 & 138.02 & 123.88 & 116.13 & 26.94 & 12.80 & 5.04 & 62.90 & 63.56 & 63.53 \\
 & 89 & 116.32 & 105.01 & 97.29 & 21.55 & 10.24 & 2.52 & 62.66 & 63.00 & 63.08 \\
 & 79 & 102.61 & 91.43 & 86.27 & 18.85 & 7.68 & 2.52 & 61.57 & 62.74 & 62.49 \\ \midrule

 \multirow{3}{*}{10} 
 & 105 & 48.38 & 35.97 & 28.16 & 25.23 & 12.82 & 5.01 & 50.53 & 50.16 & 49.04 \\
 & 89 & 40.02 & 30.09 & 22.34 & 20.19 & 10.26 & 2.50 & 50.45 & 50.00 & 48.95 \\
 & 79 & 35.29 & 25.32 & 20.14 & 17.66 & 7.69 & 2.50 & 50.06 & 49.44 & 48.45 \\ \midrule

 \multirow{3}{*}{1} 
 & 105 & 27.15 & 14.95 & 7.25 & 24.84 & 12.63 & 4.93 & 27.77 & 27.55 & 30.00 \\
 & 89 & 21.85 & 12.09 & 4.45 & 19.87 & 10.10 & 2.46 & 27.51 & 27.55 & 30.00 \\
 & 79 & 19.15 & 9.34 & 4.23 & 17.38 & 8.45 & 2.46 & 22.27 & 27.55 & 30.00 \\ \bottomrule
 
 \end{tabular}%
}
\caption{Performance scores for ViT-B\_16 with random weights weights on CIFAR10 and its ....EDPE = Effective data per epoch, method=GradMatch}
\label{tab:my-table}
\end{table*}


\begin{table*}[]
\centering
\resizebox{0.9\textwidth}{!}{%
\begin{tabular}{@{}cc | ccc | ccc | ccc@{}}
\toprule
\textbf{EDPE (\%)} & \textbf{Epoch} & \multicolumn{3}{c|}{\textbf{Total Time (Mins)}} & \multicolumn{3}{c|}{\textbf{Total Selection Time (Mins)}} & \multicolumn{3}{c}{\textbf{Accuracy}} \\ \midrule
\

\multirow{3}{*}{100} 
& 105 &  & 65.45 &  &  & - &  &  & 97.96 &  \\
 & 89 &  & 55.87 &  &  & - &  &  & 97.10 &  \\
 & 79 &  & 49.74 &  &  & - &  &  & 97.09 &  \\ \midrule

 \multicolumn{1}{c}{} & \multicolumn{1}{c|}{} & \textit{SSI=10} & \textit{SSI=20} & \textit{SSI=50} & \textit{SSI=10} & \textit{SSI=20} & \textit{SSI=50} & \textit{SSI=10} & \textit{SSI=20} & \textit{SSI=50}\\ \midrule

\multirow{3}{*}{80} 
& 105 & 124.89 & 119.48 & 115.79 & 11.43 & 6.02 & 2.33 & 97.93 & 98.00 & 97.88 \\
 & 89 & 105.98 & 101.65 & 97.99 & 9.15 & 4.82 & 1.16 & 97.93 & 97.94 & 97.81 \\
 & 79 & 93.61 & 89.21 & 86.77 & 8.00 & 3.61 & 1.16 & 97.90 & 97.93 & 97.80 \\ \midrule
 
 \multirow{3}{*}{50} 
 & 105 & 82.37 & 77.27 & 73.27 & 11.64 & 6.54 & 2.54 & 97.87 & 97.90 & 97.63 \\
 & 89 & 69.65 & 65.57 & 61.61 & 9.31 & 5.23 & 1.27 & 97.85 & 97.90 & 97.58 \\
 & 79 & 61.47 & 57.25 & 54.59 & 8.15 & 3.92 & 1.27 & 97.80 & 97.72 & 97.58 \\ \midrule

 \multirow{3}{*}{10} 
 & 105 & 30.69 & 22.63 & 18.14 & 15.95 & 7.9 & 3.40 & 97.70 & 97.30 & 96.65 \\
 & 89 & 25.39 & 18.94 & 14.33 & 12.76 & 6.31 & 1.70 & 97.60 & 97.25 & 96.65 \\
 & 79 & 22.39 & 15.96 & 12.92 & 5.15 & 4.73 & 1.70 & 97.48 & 97.07 & 96.56 \\ \midrule

 \multirow{3}{*}{1} 
 & 105 & 11.31 & 6.65 & 3.62 & 9.84 & 5.17 & 2.15 & 93.14 & 92.31 & 92.91 \\
 & 89 & 9.13 & 5.40 & 2.33 & 7.87 & 4.14 & 1.07 & 92.93 & 92.29 & 92.91 \\
 & 79 & 8.01 & 4.23 & 2.19 & 6.89 & 3.10 & 1.07 & 92.93 & 92.29 & 92.91 \\ \bottomrule
 
 \end{tabular}%
}
\caption{Performance scores for ResNet50V2 with ImageNet21k weights Initialization on CIFAR10 and its ....EDPE = Effective data per epoch, method=GradMatch}
\label{tab:my-table}
\end{table*}

\begin{table*}[]
\centering
\resizebox{0.9\textwidth}{!}{%
\begin{tabular}{@{}cc | ccc | ccc | ccc@{}}
\toprule
\textbf{EDPE (\%)} & \textbf{Epoch} & \multicolumn{3}{c|}{\textbf{Total Time (Mins)}} & \multicolumn{3}{c|}{\textbf{Total Selection Time (Mins)}} & \multicolumn{3}{c}{\textbf{Accuracy}} \\ \midrule
\

\multirow{3}{*}{100} 

& 105 &  & 65.45 &  &  & - &  &  & 88.90 &  \\
 & 89 &  & 55.87 &  &  & - &  &  & 88.64 &  \\
 & 79 &  & 49.74 &  &  & - &  &  & 88.10 &  \\ \midrule

 \multicolumn{1}{c}{} & \multicolumn{1}{c|}{} & \textit{SSI=10} & \textit{SSI=20} & \textit{SSI=50} & \textit{SSI=10} & \textit{SSI=20} & \textit{SSI=50} & \textit{SSI=10} & \textit{SSI=20} & \textit{SSI=50}\\ \midrule

\multirow{3}{*}{80} 
& 105 & 124.89 & 119.48 & 115.79 & 11.43 & 6.02 & 2.33 & 89.55 & 89.78 & 89.55 \\
 & 89 & 105.98 & 101.65 & 97.99 & 9.15 & 4.82 & 1.16 & 89.40 & 89.69 & 89.52 \\
 & 79 & 93.61 & 89.21 & 86.77 & 8.00 & 3.61 & 1.16 & 89.01 & 89.31 & 89.40 \\ \midrule

 \multirow{3}{*}{50} 
 & 105 & 82.37 & 77.27 & 73.27 & 11.64 & 6.54 & 2.54 & 86.81 & 86.26 & 85.09 \\
 & 89 & 69.65 & 65.57 & 61.61 & 9.31 & 5.23 & 1.27 & 86.65 & 86.19 & 85.09 \\
 & 79 & 61.47 & 57.25 & 54.59 & 8.15 & 3.92 & 1.27 & 86.08 & 85.74 & 84.86 \\ \midrule

 \multirow{3}{*}{10} 
 & 105 & 30.69 & 23.33 & 18.57 & 15.95 & 8.59 & 3.83 & 63.91 & 64.35 & 60.61 \\
 & 89 & 25.39 & 19.50 & 14.54 & 12.76 & 6.87 & 1.91 & 63.59 & 61.49 & 60.61 \\
 & 79 & 22.39 & 16.38 & 13.14 & 5.15 & 3.77 & 1.91 & 61.99 & 61.49 & 60.61 \\ \midrule

 \multirow{3}{*}{1} 
 & 105 & 11.31 & 6.65 & 3.62 & 9.84 & 5.17 & 2.15 & 23.74 & 24.92 & 27.15 \\
 & 89 & 9.13 & 5.40 & 2.33 & 7.87 & 4.14 & 1.07 & 23.74 & 24.92 & 27.08 \\
 & 79 & 8.01 & 4.23 & 2.19 & 6.89 & 3.10 & 1.07 & 22.83 & 24.92 & 26.76 \\ \bottomrule
 
 \end{tabular}%
}
\caption{Performance scores for ResNet50V2 with Random Initialization on CIFAR10 and its ....EDPE = Effective data per epoch, method=GradMatch}
\label{tab:my-table}
\end{table*}



\subsubsection{CRAIG, GradMatch, GLISTER and Random Results on Tiny ImageNet}

This section shows results with different subsets methods on Tiny Imagenet with  ViT\_B16, ResNet50, and MobileNet.


\begin{table*}[]
\centering
\resizebox{0.9\textwidth}{!}{%
\begin{tabular}{@{}cc | ccc | ccc | ccc@{}}
\toprule
\textbf{EDPE (\%)} & \textbf{Epoch} & \multicolumn{3}{c|}{\textbf{Total Time (Mins)}} & \multicolumn{3}{c|}{\textbf{Total Selection Time (Mins)}} & \multicolumn{3}{c}{\textbf{Accuracy}} \\ \midrule

\multirow{3}{*}{100} 
& 105 &  & 311.64 &  &  & - &  &  & 89.44 &  \\
& 89 &  & 266.52 &  &  & - &  &  & 88.94 &  \\
 & 79 &  & 236.23 &  &  & - &  &  & 88.94 &  \\
 \midrule
 \multicolumn{1}{c}{} & \multicolumn{1}{c|}{} & \textit{SSI=10} & \textit{SSI=20} & \textit{SSI=50} & \textit{SSI=10} & \textit{SSI=20} & \textit{SSI=50} & \textit{SSI=10} & \textit{SSI=20} & \textit{SSI=50}\\ \midrule
 
\multirow{3}{*}{80} 
& 105 & 575.97 & 474.68 & 421.88 & 186.06 & 84.77 & 31.97 & 89.46 & 89.28 & 89.23 \\
 & 89 & 480.69 & 399.65 & 347.82 & 148.85 & 67.81 & 15.98 & 89.46 & 89.28 & 89.23 \\
 & 79 & 425.21 & 345.83 & 310.95 & 130.24 & 50.86 & 15.98 & 89.41 & 89.28 & 89.12 \\
\bottomrule
 
 \end{tabular}%
}
\caption{Performance scores for ViT-B\_16 with ImageNet-21k weights Initialization on Tiny Imagenet and its ....EDPE = Effective data per epoch, method=CRAIG}
\label{tab:vit-img-craig}
\end{table*}


\begin{table*}[]
\centering
\resizebox{0.9\textwidth}{!}{%
\begin{tabular}{@{}cc | ccc | ccc | ccc@{}}
\toprule
\textbf{EDPE (\%)} & \textbf{Epoch} & \multicolumn{3}{c|}{\textbf{Total Time (Mins)}} & \multicolumn{3}{c|}{\textbf{Total Selection Time (Mins)}} & \multicolumn{3}{c}{\textbf{Accuracy}} \\ \midrule

\multirow{3}{*}{100} 
& 105 &  & 311.64 &  &  & - &  &  & 89.44 &  \\
& 89 &  & 266.52 &  &  & - &  &  & 88.94 &  \\
 & 79 &  & 236.23 &  &  & - &  &  & 88.94 &  \\
 \midrule

 \multicolumn{1}{c}{} & \multicolumn{1}{c|}{} & \textit{SSI=10} & \textit{SSI=20} & \textit{SSI=50} & \textit{SSI=10} & \textit{SSI=20} & \textit{SSI=50} & \textit{SSI=10} & \textit{SSI=20} & \textit{SSI=50}\\ \midrule

\multirow{3}{*}{80} 
& 105 & 518.21 & 490.10 & 452.71 & 120.78 & 61.71 & 24.32 & 89.50 & 90.19 & 89.96 \\
 & 89 & 434.47 & 415.78 & 378.57 & 96.62 & 49.37 & 12.16 & 
 89.50 & 90.19 & 89.96 \\
 & 79 & 384.77 & 361.86 & 336.99 & 84.55 & 37.03 & 12.16 & 
 89.50 & 90.19 & 89.96  \\
\bottomrule
 
 \end{tabular}%
}
\caption{Performance scores for ViT-B\_16 with ImageNet-21k weights on Tiny Imagenet and its ....EDPE = Effective data per epoch, method=GradMatch}
\label{tab:vit-img-grad}
\end{table*}

\begin{table*}[]
\centering
\resizebox{0.9\textwidth}{!}{%
\begin{tabular}{@{}cc | ccc | ccc | ccc@{}}
\toprule
\textbf{EDPE (\%)} & \textbf{Epoch} & \multicolumn{3}{c|}{\textbf{Total Time (Mins)}} & \multicolumn{3}{c|}{\textbf{Total Selection Time (Mins)}} & \multicolumn{3}{c}{\textbf{Accuracy}} \\ \midrule

\multirow{3}{*}{100} 
& 105 &  & 311.64 &  &  & - &  &  & 89.44 &  \\
& 89 &  & 266.52 &  &  & - &  &  & 88.94 &  \\
 & 79 &  & 236.23 &  &  & - &  &  & 88.94 &  \\
 \midrule

 \multicolumn{1}{c}{} & \multicolumn{1}{c|}{} & \textit{SSI=10} & \textit{SSI=20} & \textit{SSI=50} & \textit{SSI=10} & \textit{SSI=20} & \textit{SSI=50} & \textit{SSI=10} & \textit{SSI=20} & \textit{SSI=50}\\ \midrule

\multirow{3}{*}{80} 
& 105 & 472.06 & 490.10 & 452.71 & 82.14 & 39.50 & 15.48 & 
89.60 & 90.19 & 89.96 \\
 & 89 & 397.56 & 415.78 & 378.57 & 65.71 & 31.60 & 7.74 & 
 89.56 & 90.19 & 89.96 \\
 & 79 & 352.47 & 361.86 & 336.99 & 57.50 & 23.70 & 7.74 & 
 89.56 & 90.19 & 89.96  \\
\bottomrule
 
 \end{tabular}%
}
\caption{Performance scores for ViT-B\_16 with ImageNet-21k weights on Tiny Imagenet and its ....EDPE = Effective data per epoch, method=GLISTER}
\label{tab:vit-img-glis}
\end{table*}


\begin{table*}[]
\centering
\resizebox{0.9\textwidth}{!}{%
\begin{tabular}{@{}cc | ccc | ccc | ccc@{}}
\toprule
\textbf{EDPE (\%)} & \textbf{Epoch} & \multicolumn{3}{c|}{\textbf{Total Time (Mins)}} & \multicolumn{3}{c|}{\textbf{Total Selection Time (Mins)}} & \multicolumn{3}{c}{\textbf{Accuracy}} \\ \midrule

\multirow{3}{*}{100} 
& 105 &  & 311.64 &  &  & - &  &  & 89.44 &  \\
& 89 &  & 266.52 &  &  & - &  &  & 88.94 &  \\
 & 79 &  & 236.23 &  &  & - &  &  & 88.94 &  \\
 \midrule

 \multicolumn{1}{c}{} & \multicolumn{1}{c|}{} & \textit{SSI=10} & \textit{SSI=20} & \textit{SSI=50} & \textit{SSI=10} & \textit{SSI=20} & \textit{SSI=50} & \textit{SSI=10} & \textit{SSI=20} & \textit{SSI=50}\\ \midrule

\multirow{3}{*}{80} 
& 105 & 389.91 & 389.91 & 389.91 & 0.00 & 0.00 & 0.00 & 
89.33 & 90.04 & 89.93 \\
 & 89 & 331.84 & 331.84 & 331.84 & 0.00 & 0.00 & 0.00 & 
 89.33 & 90.04 & 89.93 \\
 & 79 & 294.97 & 294.97 & 294.97 & 0.00 & 0.00 & 0.00 & 
 89.33 & 90.04 & 89.93  \\
\bottomrule
 
 \end{tabular}%
}
\caption{Performance scores for ViT-B\_16 with ImageNet-21k weights on Tiny Imagenet and its ....EDPE = Effective data per epoch, method=Random}
\label{tab:vit-img-rand}
\end{table*}

\begin{table*}[]
\centering
\resizebox{0.9\textwidth}{!}{%
\begin{tabular}{@{}cc | ccc | ccc | ccc@{}}
\toprule
\textbf{EDPE (\%)} & \textbf{Epoch} & \multicolumn{3}{c|}{\textbf{Total Time (Mins)}} & \multicolumn{3}{c|}{\textbf{Total Selection Time (Mins)}} & \multicolumn{3}{c}{\textbf{Accuracy}} \\ \midrule
\

\multirow{3}{*}{100} 

& 105 &  & 111.61 &  &  & - &  &  & 79.02 &  \\
 & 89 &  & 95.45 &  &  & - &  &  & 77.94 &  \\
 & 79 &  & 84.60 &  &  & - &  &  & 77.54 &  \\ \midrule

 \multicolumn{1}{c}{} & \multicolumn{1}{c|}{} & \textit{SSI=10} & \textit{SSI=20} & \textit{SSI=50} & \textit{SSI=10} & \textit{SSI=20} & \textit{SSI=50} & \textit{SSI=10} & \textit{SSI=20} & \textit{SSI=50}\\ \midrule

\multirow{3}{*}{80} 
& 105 & 389.09 & 326.49 & 279.40 & 136.72 & 74.11 & 27.02 & 79.76 & 79.70 & 79.16 \\
 & 89 & 324.16 & 274.08 & 228.30 & 109.37 & 59.29 & 13.51 & 79.52 & 79.70 & 79.16 \\
 & 79 & 286.62 & 235.39 & 204.43 & 95.70 & 44.47 & 13.51 & 79.24 & 79.29 & 79.16 \\
\bottomrule
 
 \end{tabular}%
}
\caption{Performance scores for BiT-M-R50x1 with ImageNet-21k Initialization on Tiny-ImageNet and its ....EDPE = Effective data per epoch, method=CRAIG}
\label{tab:res-img-craig}
\end{table*}



\begin{table*}[]
\centering
\resizebox{0.9\textwidth}{!}{%
\begin{tabular}{@{}cc | ccc | ccc | ccc@{}}
\toprule
\textbf{EDPE (\%)} & \textbf{Epoch} & \multicolumn{3}{c|}{\textbf{Total Time (Mins)}} & \multicolumn{3}{c|}{\textbf{Total Selection Time (Mins)}} & \multicolumn{3}{c}{\textbf{Accuracy}} \\ \midrule

\multirow{3}{*}{100} 
& 105 &  & 111.61 &  &  & - &  &  & 79.02 &  \\
& 89 &  & 95.45 &  &  & - &  &  & 77.94 &  \\
 & 79 &  & 84.60 &  &  & - &  &  & 77.54 &  \\
 \midrule

 \multicolumn{1}{c}{} & \multicolumn{1}{c|}{} & \textit{SSI=10} & \textit{SSI=20} & \textit{SSI=50} & \textit{SSI=10} & \textit{SSI=20} & \textit{SSI=50} & \textit{SSI=10} & \textit{SSI=20} & \textit{SSI=50}\\ \midrule

\multirow{3}{*}{80} 
& 105 & 321.22 & 283.43 & 258.94 & 77.97 & 40.18 & 15.70 & 79.40 & 79.42 & 79.16 \\
 & 89 & 269.39 & 239.16 & 214.86 & 62.38 & 32.15 & 7.85 & 79.37 & 79.38 & 79.16 \\
 & 79 & 238.59 & 208.12 & 191.86 & 0.00 & 0.00 & 0.00 & 79.37 & 79.16 & 79.16  \\
\bottomrule
 
 \end{tabular}%
}
\caption{Performance scores for BiT-M-R50x1 with ImageNet-21k weights on Tiny Imagenet and its ....EDPE = Effective data per epoch, method=GradMatch}
\label{tab:rn50-img-grad}
\end{table*}

\begin{table*}[]
\centering
\resizebox{0.9\textwidth}{!}{%
\begin{tabular}{@{}cc | ccc | ccc | ccc@{}}
\toprule
\textbf{EDPE (\%)} & \textbf{Epoch} & \multicolumn{3}{c|}{\textbf{Total Time (Mins)}} & \multicolumn{3}{c|}{\textbf{Total Selection Time (Mins)}} & \multicolumn{3}{c}{\textbf{Accuracy}} \\ \midrule

\multirow{3}{*}{100} 
& 105 &  & 111.61 &  &  & - &  &  & 79.02 &  \\
& 89 &  & 95.45 &  &  & - &  &  & 77.94 &  \\
 & 79 &  & 84.60 &  &  & - &  &  & 77.54 &  \\
 \midrule

 \multicolumn{1}{c}{} & \multicolumn{1}{c|}{} & \textit{SSI=10} & \textit{SSI=20} & \textit{SSI=50} & \textit{SSI=10} & \textit{SSI=20} & \textit{SSI=50} & \textit{SSI=10} & \textit{SSI=20} & \textit{SSI=50}\\ \midrule

\multirow{3}{*}{80} 
& 105 & 309.69 & 278.64 & 259.50 & 62.48 & 31.44 & 12.29 & 
79.09 & 79.09 & 79.09 \\
 & 89 & 261.88 & 237.04 & 218.03 & 49.99 & 25.15 & 6.14 & 
 79.09 & 79.09 & 79.09 \\
 & 79 & 232.08 & 207.21 & 194.49 & 43.74 & 18.86 & 6.14 & 
 79.09 & 79.09 & 79.09  \\
\bottomrule
 
 \end{tabular}%
}
\caption{Performance scores for BiT-M-R50x1 with ImageNet-21k weights on Tiny Imagenet and its ....EDPE = Effective data per epoch, method=GLISTER}
\label{tab:vit-img-glis}
\end{table*}


\begin{table*}[]
\centering
\resizebox{0.9\textwidth}{!}{%
\begin{tabular}{@{}cc | ccc | ccc | ccc@{}}
\toprule
\textbf{EDPE (\%)} & \textbf{Epoch} & \multicolumn{3}{c|}{\textbf{Total Time (Mins)}} & \multicolumn{3}{c|}{\textbf{Total Selection Time (Mins)}} & \multicolumn{3}{c}{\textbf{Accuracy}} \\ \midrule

\multirow{3}{*}{100} 
& 105 &  & 111.61 &  &  & - &  &  & 79.02 &  \\
& 89 &  & 95.45 &  &  & - &  &  & 77.94 &  \\
 & 79 &  & 84.60 &  &  & - &  &  & 77.54 &  \\
 \midrule

 \multicolumn{1}{c}{} & \multicolumn{1}{c|}{} & \textit{SSI=10} & \textit{SSI=20} & \textit{SSI=50} & \textit{SSI=10} & \textit{SSI=20} & \textit{SSI=50} & \textit{SSI=10} & \textit{SSI=20} & \textit{SSI=50}\\ \midrule

\multirow{3}{*}{80} 
& 105 & 250.73 & 250.73 & 250.73 & 0.00 & 0.00 & 0.00 & 78.81 & 78.81 & 78.81 \\
 & 89 & 211.88 & 211.88 & 211.88 & 0.00 & 0.00 & 0.00 & 78.81 & 78.81 & 78.81 \\
 & 79 & 188.34 & 188.34 & 188.34 & 0.00 & 0.00 & 0.00 & 78.81 & 78.81 & 78.81  \\
\bottomrule
 
 \end{tabular}%
}
\caption{Performance scores for BiT-M-R50x1 with ImageNet-21k weights on Tiny Imagenet and its ....EDPE = Effective data per epoch, method=Random}
\label{tab:rn50-img-rand}
\end{table*}


\begin{table*}[]
\centering
\resizebox{0.9\textwidth}{!}{%
\begin{tabular}{@{}cc | ccc | ccc | ccc@{}}
\toprule
\textbf{EDPE (\%)} & \textbf{Epoch} & \multicolumn{3}{c|}{\textbf{Total Time (Mins)}} & \multicolumn{3}{c|}{\textbf{Total Selection Time (Mins)}} & \multicolumn{3}{c}{\textbf{Accuracy}} \\ \midrule

\multirow{3}{*}{100} 
& 105 &  & 113.96 &  &  & - &  &  & 74.83 &  \\
& 89 &  & 97.68 &  &  & - &  &  & 74.79 &  \\
 & 79 &  & 86.82 &  &  & - &  &  & 74.56 &  \\
 \midrule
 \multicolumn{1}{c}{} & \multicolumn{1}{c|}{} & \textit{SSI=10} & \textit{SSI=20} & \textit{SSI=50} & \textit{SSI=10} & \textit{SSI=20} & \textit{SSI=50} & \textit{SSI=10} & \textit{SSI=20} & \textit{SSI=50}\\ \midrule

\multirow{3}{*}{80} 
& 105 & 398.84 & 307.91 & 265.86 & 161.91 & 70.98 & 28.93 & 75.76 & 75.24 & 74.88 \\
 & 89 & 331.17 & 258.42 & 216.10 & 129.53 & 56.78 & 14.46 & 75.51 & 75.17 & 74.88 \\
 & 79 & 292.57 & 221.82 & 193.70 & 113.33 & 42.59 & 14.46 & 75.24 & 75.05 & 74.47 \\
\bottomrule
 
 \end{tabular}%
}
\caption{Performance scores for MobileNetV3 with ImageNet-1k weights Initialization on Tiny Imagenet and its ....EDPE = Effective data per epoch, method=CRAIG}
\label{tab:mobv3-img-craig}
\end{table*}

\begin{table*}[]
\centering
\resizebox{0.9\textwidth}{!}{%
\begin{tabular}{@{}cc | ccc | ccc | ccc@{}}
\toprule
\textbf{EDPE (\%)} & \textbf{Epoch} & \multicolumn{3}{c|}{\textbf{Total Time (Mins)}} & \multicolumn{3}{c|}{\textbf{Total Selection Time (Mins)}} & \multicolumn{3}{c}{\textbf{Accuracy}} \\ \midrule

\multirow{3}{*}{100} 
& 105 &  & 113.96 &  &  & - &  &  & 74.83 &  \\
& 89 &  & 97.68 &  &  & - &  &  & 74.79 &  \\
 & 79 &  & 86.82 &  &  & - &  &  & 74.56 &  \\
 \midrule
 \multicolumn{1}{c}{} & \multicolumn{1}{c|}{} & \textit{SSI=10} & \textit{SSI=20} & \textit{SSI=50} & \textit{SSI=10} & \textit{SSI=20} & \textit{SSI=50} & \textit{SSI=10} & \textit{SSI=20} & \textit{SSI=50}\\ \midrule

\multirow{3}{*}{80} 
& 105 & 314.47 & 275.92 & 252.20 & 77.55 & 38.99 & 15.27 & 
75.39 & 75.58 & 74.92 \\
 & 89 & 263.68 & 232.83 & 209.27 & 62.04 & 31.19 & 7.63 & 
 75.31 & 75.29 & 74.91 \\
 & 79 & 233.52 & 202.63 & 186.87 & 54.28 & 23.39 & 7.63 & 
 75.25 & 74.87 & 74.82 \\
\bottomrule
 
 \end{tabular}%
}
\caption{Performance scores for MobileNetV3 with ImageNet-1k weights Initialization on Tiny Imagenet and its ....EDPE = Effective data per epoch, method=GradMatch}
\label{tab:mobv3-img-grad}
\end{table*}

\begin{table*}[]
\centering
\resizebox{0.9\textwidth}{!}{%
\begin{tabular}{@{}cc | ccc | ccc | ccc@{}}
\toprule
\textbf{EDPE (\%)} & \textbf{Epoch} & \multicolumn{3}{c|}{\textbf{Total Time (Mins)}} & \multicolumn{3}{c|}{\textbf{Total Selection Time (Mins)}} & \multicolumn{3}{c}{\textbf{Accuracy}} \\ \midrule

\multirow{3}{*}{100} 
& 105 &  & 113.96 &  &  & - &  &  & 74.83 &  \\
& 89 &  & 97.68 &  &  & - &  &  & 74.79 &  \\
 & 79 &  & 86.82 &  &  & - &  &  & 74.56 &  \\
 \midrule
 \multicolumn{1}{c}{} & \multicolumn{1}{c|}{} & \textit{SSI=10} & \textit{SSI=20} & \textit{SSI=50} & \textit{SSI=10} & \textit{SSI=20} & \textit{SSI=50} & \textit{SSI=10} & \textit{SSI=20} & \textit{SSI=50}\\ \midrule

\multirow{3}{*}{80} 
& 105 & 275.87 & 256.01 & 244.63 & 38.94 & 19.08 & 7.70 & 
75.87 & 75.63 & 75.22 \\
 & 89 & 232.79 & 216.90 & 205.49 & 31.15 & 15.26 & 3.85 & 
 75.71 & 75.48 & 75.04 \\
 & 79 & 206.49 & 190.68 & 183.08 & 27.25 & 11.44 & 3.85 & 
 75.45 & 75.24 & 74.76 \\
\bottomrule
 
 \end{tabular}%
}
\caption{Performance scores for MobileNetV3 with ImageNet-1k weights Initialization on Tiny Imagenet and its ....EDPE = Effective data per epoch, method=GLISTER}
\label{tab:mobv3-img-glis}
\end{table*}

\begin{table*}[]
\centering
\resizebox{0.9\textwidth}{!}{%
\begin{tabular}{@{}cc | ccc | ccc | ccc@{}}
\toprule
\textbf{EDPE (\%)} & \textbf{Epoch} & \multicolumn{3}{c|}{\textbf{Total Time (Mins)}} & \multicolumn{3}{c|}{\textbf{Total Selection Time (Mins)}} & \multicolumn{3}{c}{\textbf{Accuracy}} \\ \midrule

\multirow{3}{*}{100} 
& 105 &  & 113.96 &  &  & - &  &  & 74.83 &  \\
& 89 &  & 97.68 &  &  & - &  &  & 74.79 &  \\
 & 79 &  & 86.82 &  &  & - &  &  & 74.56 &  \\
 \midrule
 \multicolumn{1}{c}{} & \multicolumn{1}{c|}{} & \textit{SSI=10} & \textit{SSI=20} & \textit{SSI=50} & \textit{SSI=10} & \textit{SSI=20} & \textit{SSI=50} & \textit{SSI=10} & \textit{SSI=20} & \textit{SSI=50}\\ \midrule

\multirow{3}{*}{80} 
& 105 & 236.92 & 236.92 & 236.92 & 0.0 & 0.0 & 0.0 & 
74.44 & 74.12 & 74.92 \\
 & 89 & 201.64 & 201.64 & 201.64 & 0.0 & 0.0 & 0.0 & 
 74.40 & 74.10 & 74.91 \\
 & 79 & 179.23 & 179.23 & 179.23 & 0.0 & 0.0 & 0.0 & 
 74.12 & 74.00 & 74.82 \\
\bottomrule
 
 \end{tabular}%
}
\caption{Performance scores for MobileNetV3 with ImageNet-1k weights Initialization on Tiny Imagenet and its ....EDPE = Effective data per epoch, method=Random}
\label{tab:mobv3-img-rand}
\end{table*}


\begin{table*}[]
\centering
\resizebox{0.9\textwidth}{!}{%
\begin{tabular}{@{}cc | ccc | ccc | ccc@{}}
\toprule
\textbf{EDPE (\%)} & \textbf{Epoch} & \multicolumn{3}{c|}{\textbf{Total Time (Mins)}} & \multicolumn{3}{c|}{\textbf{Total Selection Time (Mins)}} & \multicolumn{3}{c}{\textbf{Accuracy}} \\ \midrule

\multirow{3}{*}{100} 
& 105 &  & 311.64 &  &  & - &  &  & - &  \\
& 89 &  & 266.52 &  &  & - &  &  & - &  \\
 & 79 &  & 236.23 &  &  & - &  &  & - &  \\
 \midrule
 \multicolumn{1}{c}{} & \multicolumn{1}{c|}{} & \textit{SSI=10} & \textit{SSI=20} & \textit{SSI=50} & \textit{SSI=10} & \textit{SSI=20} & \textit{SSI=50} & \textit{SSI=10} & \textit{SSI=20} & \textit{SSI=50}\\ \midrule

\multirow{3}{*}{80} 
& 105 & 1150.82 & 307.91 & 265.86 & 226.14 & 103.75 & 40.27 & 90.89 & - & - \\
 & 89 & 967.87 & 258.42 & 216.10 & 180.91 & 83.00 & 20.13 & 90.78 & - & - \\
 & 79 & 857.81 & 221.82 & 193.70 & 158.30 & 62.25 & 20.13 & 90.60 & - & - \\
\bottomrule
 
 \end{tabular}%
}
\caption{Performance scores for Swin TransformerV2 with ImageNet-21K and ImageNet1k fine-tuned weights Initialization on Tiny Imagenet and its ....EDPE = Effective data per epoch, method=CRAIG}
\label{tab:swin-img-craig}
\end{table*}

\begin{table*}[]
\centering
\resizebox{0.9\textwidth}{!}{%
\begin{tabular}{@{}cc | ccc | ccc | ccc@{}}
\toprule
\textbf{EDPE (\%)} & \textbf{Epoch} & \multicolumn{3}{c|}{\textbf{Total Time (Mins)}} & \multicolumn{3}{c|}{\textbf{Total Selection Time (Mins)}} & \multicolumn{3}{c}{\textbf{Accuracy}} \\ \midrule

\multirow{3}{*}{100} 
& 105 &  & 311.64 &  &  & - &  &  & 89.44 &  \\
& 89 &  & 266.52 &  &  & - &  &  & 88.94 &  \\
 & 79 &  & 236.23 &  &  & - &  &  & 88.94 &  \\
 \midrule
 \multicolumn{1}{c}{} & \multicolumn{1}{c|}{} & \textit{SSI=10} & \textit{SSI=20} & \textit{SSI=50} & \textit{SSI=10} & \textit{SSI=20} & \textit{SSI=50} & \textit{SSI=10} & \textit{SSI=20} & \textit{SSI=50}\\ \midrule

\multirow{3}{*}{80} 
& 105 & 398.84 & 307.91 & 265.86 & 125.24 & 62.45 & 25.20 & 
- & - & - \\
 & 89 & 331.17 & 258.42 & 216.10 & 100.19 & 49.96 & 12.60 & 
 - & - & - \\
 & 79 & 292.57 & 221.82 & 193.70 & 87.67 & 37.47 & 20.13 & 
 - & - & - \\
\bottomrule
 
 \end{tabular}%
}
\caption{Performance scores for Swin TransformerV2 with ImageNet-21K and ImageNet1k fine-tuned weights Initialization on Tiny Imagenet and its ....EDPE = Effective data per epoch, method=GradMatch}
\label{tab:swin-img-grad}
\end{table*}

\begin{table*}[]
\centering
\resizebox{0.9\textwidth}{!}{%
\begin{tabular}{@{}cc | ccc | ccc | ccc@{}}
\toprule
\textbf{EDPE (\%)} & \textbf{Epoch} & \multicolumn{3}{c|}{\textbf{Total Time (Mins)}} & \multicolumn{3}{c|}{\textbf{Total Selection Time (Mins)}} & \multicolumn{3}{c}{\textbf{Accuracy}} \\ \midrule

\multirow{3}{*}{100} 
& 105 &  & 311.64 &  &  & - &  &  & - &  \\
& 89 &  & 266.52 &  &  & - &  &  & - &  \\
 & 79 &  & 236.23 &  &  & - &  &  & - &  \\
 \midrule
 \multicolumn{1}{c}{} & \multicolumn{1}{c|}{} & \textit{SSI=10} & \textit{SSI=20} & \textit{SSI=50} & \textit{SSI=10} & \textit{SSI=20} & \textit{SSI=50} & \textit{SSI=10} & \textit{SSI=20} & \textit{SSI=50}\\ \midrule

\multirow{3}{*}{80} 
& 105 & 1024.94 & 307.91 & 265.86 & 226.14 & 103.75 & 40.27 & 90.97 & - & - \\
 & 89 & 867.17 & 258.42 & 216.10 & 180.91 & 83.00 & 20.13 & 90.95 & - & - \\
 & 79 & 769.70 & 221.82 & 193.70 & 158.30 & 62.25 & 20.13 & 90.95 & - & - \\
\bottomrule
 
 \end{tabular}%
}
\caption{Performance scores for Swin TransformerV2 with ImageNet-21K and ImageNet1k fine-tuned weights Initialization on Tiny Imagenet and its ....EDPE = Effective data per epoch, method=GLISTER}
\label{tab:swin-img-glis}
\end{table*}

\begin{table*}[]
\centering
\resizebox{0.9\textwidth}{!}{%
\begin{tabular}{@{}cc | ccc | ccc | ccc@{}}
\toprule
\textbf{EDPE (\%)} & \textbf{Epoch} & \multicolumn{3}{c|}{\textbf{Total Time (Mins)}} & \multicolumn{3}{c|}{\textbf{Total Selection Time (Mins)}} & \multicolumn{3}{c}{\textbf{Accuracy}} \\ \midrule

\multirow{3}{*}{100} 
& 105 &  & 311.64 &  &  & - &  &  & 89.44 &  \\
& 89 &  & 266.52 &  &  & - &  &  & 88.94 &  \\
 & 79 &  & 236.23 &  &  & - &  &  & 88.94 &  \\
 \midrule
 \multicolumn{1}{c}{} & \multicolumn{1}{c|}{} & \textit{SSI=10} & \textit{SSI=20} & \textit{SSI=50} & \textit{SSI=10} & \textit{SSI=20} & \textit{SSI=50} & \textit{SSI=10} & \textit{SSI=20} & \textit{SSI=50}\\ \midrule

\multirow{3}{*}{80} 
& 105 & 924.67 & 924.67 & 924.67 & 0.0 & 0.0 & 0.0 & 
- & - & - \\
 & 89 & 786.95 & 786.95 & 786.95 & 0.0 & 0.0 & 0.0 & 
 - & - & - \\
 & 79 & 699.51 & 699.51 & 699.51 & 0.0 & 0.0 & 0.0 & 
 - & - & - \\
\bottomrule
 
 \end{tabular}%
}
\caption{Performance scores for Swin TransformerV2 with ImageNet-21K and ImageNet1k fine-tuned weights Initialization on Tiny Imagenet and its ....EDPE = Effective data per epoch, method=Random}
\label{tab:swin-img-rand}
\end{table*}

\subsubsection{CRAIG, GradMatch, GLISTER and Random Results on UltraMNIST}


To examine how well transformers can learn a robust representation in a dataset, we used UltraMNIST to compare ViT\_B16 and ResNet50. Tables [\ref{tab:vit-um-craig}, \ref{tab:vit-um-grad}, \ref{tab:vit-um-glis}, \ref{tab:vit-um-rand}, \ref{tab:rn50-um-craig}, \ref{tab:rn50-um-grad}, \ref{tab:rn50-um-glis}, \ref{tab:rn50-um-rand}] delivers a comparative study for these models, and we used Fig. \ref{fig:um-coresets} to exhibit that CNN outperforms across all coreset sizes and that due to nonavailability of similar pretraining on such non-natural image dataset, transformers tend to suffer in learning desired semantic context as mentioned in the section 'Learning semantic coherency in non-natural images' in the main paper. For UltraMNIST, GradMatch seems to perform best when compared to other methods, and random selection performs worst. These comments are uniform with varying SSI values, as shown in Fig. \ref{fig:um-ssi} in the main document. 


\begin{table}[h]
\centering
\resizebox{0.9\textwidth}{!}{%
\begin{tabular}{@{}cc | ccc | ccc | ccc@{}}
\toprule
\textbf{EDPE (\%)} & \textbf{Epoch} & \multicolumn{3}{c|}{\textbf{Total Time (Mins)}} & \multicolumn{3}{c|}{\textbf{Total Selection Time (Mins)}} & \multicolumn{3}{c}{\textbf{Accuracy}} \\ \midrule

\multirow{3}{*}{100} 
& 105 &  & 239.53 &  &  & - &  &  & 40.35 &  \\
& 89 &  & 205.31 &  &  & - &  &  & 40.35 &  \\
 & 79 &  & 181.98 &  &  & - &  &  & 40.35 &  \\
 \midrule

 \multicolumn{1}{c}{} & \multicolumn{1}{c|}{} & \textit{SSI=10} & \textit{SSI=20} & \textit{SSI=50} & \textit{SSI=10} & \textit{SSI=20} & \textit{SSI=50} & \textit{SSI=10} & \textit{SSI=20} & \textit{SSI=50}\\ \midrule


 \multirow{3}{*}{50} 
 & 105 & 282.06 & 200.85 & 153.73 & 162.29 & 81.09 & 33.97 & 24.53 & 22.96 & 19.35 \\
 & 89 & 232.38 & 167.42 & 119.53 & 129.83 & 64.87 & 16.98 & 
 24.53 & 22.96 & 19.35 \\
 & 79 & 204.23 & 139.28 & 107.61 & 113.60 & 48.65 & 16.98 & 
 24.10 & 22.25 & 19.35 \\ \midrule

 \multirow{3}{*}{30} 
 & 105 & 233.06 & 151.86 & 104.74 & 162.29 & 81.09 & 33.97 & 
 16.60 & 15.21 & 13.32 \\
 & 89 & 190.08 & 125.12 & 77.230 & 129.83 & 64.87 & 16.98 & 
 16.28 & 15.14 & 13.32 \\
 & 79 & 167.32 & 102.36 & 70.69 & 113.60 & 48.65 & 16.98 &
 15.96 & 14.60 & 13.32 \\ \midrule

 \multirow{3}{*}{10} 
 & 105 & 184.07 & 102.86 & 55.74 & 162.29 & 81.09 & 33.97 & 9.71 & 8.75 & 7.39 \\
 & 89 & 148.50 & 83.54 & 35.64 & 129.83 & 64.87 & 16.98 & 9.53 & 8.14 & 7.39 \\
 & 79 & 130.19 & 65.24 & 33.57 & 113.60 & 48.65 & 16.98 & 8.75 & 7.82 & 7.39 \\ \midrule

 \multirow{3}{*}{1} 
 & 105 & 173.18 & 91.98 & 44.85 & 162.29 & 81.09 & 33.97 & 
 4.64 & 4.42 & 4.53 \\
 & 89 & 139.16 & 74.20 & 26.31 & 129.83 & 64.87 & 16.98 & 
 4.64 & 4.42 & 4.53  \\
 & 79 & 121.69 & 56.74 & 25.07 & 113.60 & 48.65 & 16.98 & 
 4.64 & 4.42 & 4.53 \\ \bottomrule
 
 \end{tabular}%
}
\caption{Performance scores for ViT-B\_16 with ImageNet-21k weights on Ultramnist and its ....EDPE = Effective data per epoch, method=CRAIG}
\label{tab:vit-um-craig}
\end{table}



\begin{table}[h]
\centering
\resizebox{0.9\textwidth}{!}{%
\begin{tabular}{@{}cc | ccc | ccc | ccc@{}}
\toprule
\textbf{EDPE (\%)} & \textbf{Epoch} & \multicolumn{3}{c|}{\textbf{Total Time (Mins)}} & \multicolumn{3}{c|}{\textbf{Total Selection Time (Mins)}} & \multicolumn{3}{c}{\textbf{Accuracy}} \\ \midrule

\multirow{3}{*}{100} 
& 105 &  & 239.53 &  &  & - &  &  & 40.35 &  \\
& 89 &  & 205.31 &  &  & - &  &  & 40.35 &  \\
 & 79 &  & 181.98 &  &  & - &  &  & 40.35 &  \\
 \midrule

 \multicolumn{1}{c}{} & \multicolumn{1}{c|}{} & \textit{SSI=10} & \textit{SSI=20} & \textit{SSI=50} & \textit{SSI=10} & \textit{SSI=20} & \textit{SSI=50} & \textit{SSI=10} & \textit{SSI=20} & \textit{SSI=50}\\ \midrule

\multirow{3}{*}{80} 
& 105 & 276.06 & 233.25 & 207.70 & 85.52 & 42.71 & 17.17 & 
36.75 & 33.21 & 32.75 \\
 & 89 & 231.73 & 197.49 & 171.90 & 68.42 & 34.17 & 8.58 & 
 36.57 & 32.71 & 32.75 \\
 & 79 & 205.04 & 170.80 & 153.75 & 59.87 & 25.63 & 8.58 & 
 36.10 & 32.60 & 32.75  \\ \midrule

 \multirow{3}{*}{50} 
 & 105 & 205.29 & 162.48 & 136.93 & 85.52 & 42.71 & 17.17 & 25.71 & 22.82 & 19.46 \\
 & 89 & 169.52 & 135.27 & 109.68 & 68.42 & 34.17 & 8.58 & 
 24.96 & 22.57 & 19.28 \\
 & 79 & 150.60 & 116.36 & 99.31 & 59.87 & 25.63 & 8.58 & 
 24.96 & 21.96 & 19.10 \\ \midrule

 \multirow{3}{*}{30} 
 & 105 & 156.29 & 113.49 & 87.94 & 85.52 & 42.71 & 17.17 & 
 18.82 & 14.78 & 13.28 \\
 & 89 & 128.04 & 93.79 & 68.20 & 68.42 & 34.17 & 8.58 & 
 18.21 & 14.78 & 13.28 \\
 & 79 & 111.71 & 77.47 & 60.43 & 59.87 & 25.63 & 8.58 &
 18.21 & 14.46 & 13.07 \\ \midrule

 \multirow{3}{*}{10} 
 & 105 & 107.30 & 64.49 & 38.94 & 85.52 & 42.71 & 17.17 & 9.25 & 7.53 & 7.42 \\
 & 89 & 86.56 & 52.32 & 26.73 & 68.42 & 34.17 & 8.58 & 9.25 & 7.53 & 6.75 \\
 & 79 & 75.42 & 41.18 & 24.13 & 59.87 & 25.63 & 8.58 & 7.42 & 7.53 & 6.75 \\ \midrule

 \multirow{3}{*}{1} 
 & 105 & 96.41 & 53.60 & 28.05 & 85.52 & 42.71 & 17.17 & 3.96 & 4.14 & 4.39 \\
 & 89 & 77.75 & 43.50 & 17.91 & 68.42 & 34.17 & 8.58 & 3.96 & 4.14 & 4.32  \\
 & 79 & 67.95 & 33.71 & 16.67 & 59.87 & 25.63 & 8.58 & 3.96 & 4.14 & 4.28 \\ \bottomrule
 
 \end{tabular}%
}
\caption{Performance scores for ViT-B\_16 with ImageNet-21k weights on Ultramnist and its ....EDPE = Effective data per epoch, method=GradMatch}
\label{tab:vit-um-grad}
\end{table}



\begin{table}[h]
\centering
\resizebox{0.9\textwidth}{!}{%
\begin{tabular}{@{}cc | ccc | ccc | ccc@{}}
\toprule
\textbf{EDPE (\%)} & \textbf{Epoch} & \multicolumn{3}{c|}{\textbf{Total Time (Mins)}} & \multicolumn{3}{c|}{\textbf{Total Selection Time (Mins)}} & \multicolumn{3}{c}{\textbf{Accuracy}} \\ \midrule

\multirow{3}{*}{100} 
& 105 &  & 239.53 &  &  & - &  &  & 40.35 &  \\
& 89 &  & 205.31 &  &  & - &  &  & 40.35 &  \\
 & 79 &  & 181.98 &  &  & - &  &  & 40.35 &  \\
 \midrule

 \multicolumn{1}{c}{} & \multicolumn{1}{c|}{} & \textit{SSI=10} & \textit{SSI=20} & \textit{SSI=50} & \textit{SSI=10} & \textit{SSI=20} & \textit{SSI=50} & \textit{SSI=10} & \textit{SSI=20} & \textit{SSI=50}\\ \midrule

\multirow{3}{*}{80} 
& 105 & 337.64 & 268.18 & 220.34 & 147.10 & 77.64 & 29.80 & 
37.35 & 35.42 & 32.78 \\
 & 89 & 280.06 & 224.50 & 177.28 & 117.68 & 62.11 & 14.90 & 
 37.35 & 35.28 & 32.78 \\
 & 79 & 246.69 & 190.30 & 158.62 & 102.97 & 46.58 & 14.90 & 
 37.35 & 35.28 & 32.78  \\ \midrule

 \multirow{3}{*}{50} 
 & 105 & 266.86 & 197.41 & 149.57 & 147.10 & 77.64 & 29.80 & 24.35 & 24.28 & 20.14 \\
 & 89 & 220.23 & 164.67 & 117.45 & 117.68 & 62.11 & 14.90 & 
 24.35 & 24.28 & 19.82 \\
 & 79 & 193.60 & 137.21 & 105.53 & 102.97 & 46.58 & 14.90 & 
 24.35 & 24.00 & 19.82 \\ \midrule

 \multirow{3}{*}{30} 
 & 105 & 217.87 & 148.41 & 100.57 & 147.10 & 77.64 & 29.80 & 
 19.07 & 15.64 & 13.53 \\
 & 89 & 177.92 & 122.36 & 75.14 & 117.68 & 62.11 & 14.90 & 
 17.89 & 15.64 & 13.53 \\
 & 79 & 156.68 & 100.30 & 68.61 & 102.97 & 46.58 & 14.90 &
 17.39 & 14.57 & 13.53 \\ \midrule

 \multirow{3}{*}{10} 
 & 105 & 168.87 & 99.42 & 51.58 & 147.10 & 77.64 & 29.80 & 5.00 & 7.46 & 6.92 \\
 & 89 & 136.34 & 80.78 & 33.56 & 117.68 & 62.11 & 14.90 & 5.00 & 7.46 & 6.92 \\
 & 79 & 119.56 & 63.17 & 31.49 & 102.97 & 46.58 & 14.90 & 5.00 & 7.46 & 6.75 \\ \midrule

 \multirow{3}{*}{1} 
 & 105 & 157.99 & 88.53 & 40.69 & 147.10 & 77.64 & 29.80 & 
 4.14 & 3.96 & 4.21 \\
 & 89 & 127.01 & 71.45 & 24.23 & 117.68 & 62.11 & 14.90 & 
 4.14 & 3.96 & 4.21  \\
 & 79 & 111.06 & 54.67 & 22.99 & 102.97 & 46.58 & 14.90 & 
 4.14 & 3.96 & 4.21 \\ \bottomrule
 
 \end{tabular}%
}
\caption{Performance scores for ViT-B\_16 with ImageNet-21k weights on Ultramnist and its ....EDPE = Effective data per epoch, method=GLISTER}
\label{tab:vit-um-glis}
\end{table}

\begin{table}[h]
\centering
\resizebox{0.9\textwidth}{!}{%
\begin{tabular}{@{}cc | ccc | ccc | ccc@{}}
\toprule
\textbf{EDPE (\%)} & \textbf{Epoch} & \multicolumn{3}{c|}{\textbf{Total Time (Mins)}} & \multicolumn{3}{c|}{\textbf{Total Selection Time (Mins)}} & \multicolumn{3}{c}{\textbf{Accuracy}} \\ \midrule

\multirow{3}{*}{100} 
& 105 &  & 239.53 &  &  & - &  &  & 40.35 &  \\
& 89 &  & 205.31 &  &  & - &  &  & 40.35 &  \\
 & 79 &  & 181.98 &  &  & - &  &  & 40.35 &  \\
 \midrule

 \multicolumn{1}{c}{} & \multicolumn{1}{c|}{} & \textit{SSI=10} & \textit{SSI=20} & \textit{SSI=50} & \textit{SSI=10} & \textit{SSI=20} & \textit{SSI=50} & \textit{SSI=10} & \textit{SSI=20} & \textit{SSI=50}\\ \midrule

\multirow{3}{*}{80} 
& 105 & 190.53 & 190.53 & 190.53 & 0.00 & 0.00 & 0.00 & 
- & 33.89 & - \\
 & 89 & 162.38 & 162.38 & 162.38 & 0.00 & 0.00 & 0.00 & 
 - & 33.89 & - \\
 & 79 & 143.71 & 143.71 & 143.71 & 0.00 & 0.00 & 0.00 & 
 - & 33.89 & -  \\ \midrule

 \multirow{3}{*}{50} 
 & 105 & 119.76 & 119.76 & 119.76 & 0.00 & 0.00 & 0.00 & 
 18.39 & 18.92 & 19.28 \\
 & 89 & 102.55 & 102.55 & 102.55 & 0.00 & 0.00 & 0.00 & 
 18.39 & 18.92 & 19.28 \\
 & 79 & 90.62 & 90.62 & 90.62 & 0.00 & 0.00 & 0.00 & 
 18.39 & 18.92 & 19.25 \\ \midrule

 \multirow{3}{*}{30} 
 & 105 & 70.77 & 70.77 & 70.77 & 0.00 & 0.00 & 0.00 & 
 12.17 & 13.03 & 12.75 \\
 & 89 & 60.24 & 60.24 & 60.24 & 0.00 & 0.00 & 0.00 & 
 12.17 & 13.03 & 12.75 \\
 & 79 & 53.71 & 53.71 & 53.71 & 0.00 & 0.00 & 0.00 & 
 12.17 & 13.03 & 12.75 \\ \midrule

 \multirow{3}{*}{10} 
 & 105 & 21.77 & 21.77 & 21.77 & 0.00 & 0.00 & 0.00 & 
 7.03 & 7.50 & 7.75 \\
 & 89 & 18.66 & 18.66 & 18.66 & 0.00 & 0.00 & 0.00 & 
 7.03 & 7.50 & 7.75 \\
 & 79 & 16.59 & 16.59 & 16.59 & 0.00 & 0.00 & 0.00 & 
 7.03 & 7.50 & 7.75 \\ \midrule

 \multirow{3}{*}{1} 
 & 105 & 10.88 & 10.88 & 10.88 & 0.00 & 0.00 & 0.00 & 
 4.00 & 3.96 & 4.21 \\
 & 89 & 9.33 & 9.33 & 9.33 & 0.00 & 0.00 & 0.00 & 
 4.00 & 3.96 & 4.21  \\
 & 79 & 8.08 & 8.08 & 8.08 & 0.00 & 0.00 & 0.00 & 
 4.00 & 3.96 & 4.21 \\ \bottomrule
 
 \end{tabular}%
}
\caption{Performance scores for ViT-B\_16 with ImageNet-21k weights on UltraMNIST and its ....EDPE = Effective data per epoch, method=Random}
\label{tab:vit-um-rand}
\end{table}



\begin{table}[h]
\centering
\resizebox{0.9\textwidth}{!}{%
\begin{tabular}{@{}cc | ccc | ccc | ccc@{}}
\toprule
\textbf{EDPE (\%)} & \textbf{Epoch} & \multicolumn{3}{c|}{\textbf{Total Time (Mins)}} & \multicolumn{3}{c|}{\textbf{Total Selection Time (Mins)}} & \multicolumn{3}{c}{\textbf{Accuracy}} \\ \midrule

\multirow{3}{*}{100} 
& 105 &  & 114.99 &  &  & - &  &  & 61.28 &  \\
& 89 &  & 98.56 &  &  & - &  &  & 61.28 &  \\
 & 79 &  & 87.36 &  &  & - &  &  & 61.21 &  \\
 \midrule

 \multicolumn{1}{c}{} & \multicolumn{1}{c|}{} & \textit{SSI=10} & \textit{SSI=20} & \textit{SSI=50} & \textit{SSI=10} & \textit{SSI=20} & \textit{SSI=50} & \textit{SSI=10} & \textit{SSI=20} & \textit{SSI=50}\\ \midrule

\multirow{3}{*}{80} 
& 105 & 138.70 & 115.05 & 100.78 & 47.23 & 23.58 & 9.31 &
54.28 & 51.78 & 48.21 \\
 & 89 & 115.74 & 96.82 & 82.61 & 37.78 & 18.86 & 4.65 &
 53.60 & 51.75 & 47.39 \\
 & 79 & 102.05 & 83.14 & 73.65 & 33.06 & 14.15 & 4.65 &
 52.07 & 50.85 & 46.92  \\ \midrule

 \multirow{3}{*}{50} 
 & 105 & 104.72 & 81.07 & 66.80 & 47.23 & 23.58 & 9.31 &
 40.28 & 34.25 & 28.82 \\
 & 89 & 87.01 & 68.09 & 53.88 & 37.78 & 18.86 & 4.65 & 
 40.28 & 34.25 & 28.67 \\
 & 79 & 76.56 & 57.65 & 48.16 & 33.06 & 14.15 & 4.65 & 
 39.89 & 32.71 & 28.32 \\ \midrule

 \multirow{3}{*}{30} 
 & 105 & 81.20 & 57.55 & 43.28 & 47.23 & 23.58 & 9.31 & 
 24.57 & 21.39 & 17.39 \\
 & 89 & 66.70 & 47.78 & 33.57 & 37.78 & 18.86 & 4.65 & 
 24.46 & 21.39 & 17.28 \\
 & 79 & 58.84 & 39.93 & 30.44 & 33.06 & 14.15 & 4.65 & 
 24.46 & 20.67 & 17.28 \\ \midrule

 \multirow{3}{*}{10} 
 & 105 & 57.68 & 34.03 & 19.76 & 47.23 & 23.58 & 9.31 & 
 15.78 & 13.00 & 11.46 \\
 & 89 & 46.74 & 27.82 & 13.61 & 37.78 & 18.86 & 4.65 & 
 15.78 & 13.00 & 11.46 \\
 & 79 & 41.02 & 22.11 & 12.62 & 33.06 & 14.15 & 4.65 & 
 15.78 & 13.00 & 11.46 \\ \midrule

 \multirow{3}{*}{1} 
 & 105 & 52.45 & 28.81 & 14.54 & 47.23 & 23.58 & 9.31 & 
 6.92 & 5.50 & 5.85 \\
 & 89 & 42.26 & 23.34 & 9.13 & 37.78 & 18.86 & 4.65 & 
 6.78 & 5.50 & 5.60  \\
 & 79 & 36.94 & 18.03 & 8.53 & 33.06 & 14.15 & 4.65 & 
 6.25 & 5.46 & 5.39 \\ \bottomrule
 
 \end{tabular}%
}
\caption{Performance scores for ResNet50 with ImageNet-21k weights on UltraMNIST and its ....EDPE = Effective data per epoch, method=CRAIG}
\label{tab:rn50-um-craig}
\end{table}



\begin{table}[h]
\centering
\resizebox{0.9\textwidth}{!}{%
\begin{tabular}{@{}cc | ccc | ccc | ccc@{}}
\toprule
\textbf{EDPE (\%)} & \textbf{Epoch} & \multicolumn{3}{c|}{\textbf{Total Time (Mins)}} & \multicolumn{3}{c|}{\textbf{Total Selection Time (Mins)}} & \multicolumn{3}{c}{\textbf{Accuracy}} \\ \midrule

\multirow{3}{*}{100} 
& 105 &  & 114.99 &  &  & - &  &  & 61.28 &  \\
& 89 &  & 98.56 &  &  & - &  &  & 61.28 &  \\
 & 79 &  & 87.36 &  &  & - &  &  & 61.21 &  \\
 \midrule

 \multicolumn{1}{c}{} & \multicolumn{1}{c|}{} & \textit{SSI=10} & \textit{SSI=20} & \textit{SSI=50} & \textit{SSI=10} & \textit{SSI=20} & \textit{SSI=50} & \textit{SSI=10} & \textit{SSI=20} & \textit{SSI=50}\\ \midrule

\multirow{3}{*}{80} 
& 105 & 126.53 & 109.21 & 98.26 & 35.06 & 17.74 & 6.79 &
52.71 & 51.46 & 49.03 \\
 & 89 & 105.20 & 91.35 & 80.55 & 28.05 & 14.19 & 3.39 &
 52.71 & 51.39 & 48.96 \\
 & 79 & 94.23 & 80.33 & 73.08 & 24.54 & 10.64 & 3.39 &
 52.57 & 50.82 & 48.82  \\ \midrule

 \multirow{3}{*}{50} 
 & 105 & 92.55 & 75.23 & 64.29 & 35.06 & 17.74 & 6.79 &
 42.17 & 38.07 & 32.32 \\
 & 89 & 75.34 & 61.48 & 50.68 & 28.05 & 14.19 & 3.39 & 
 41.64 & 37.64 & 31.28 \\
 & 79 & 66.85 & 52.95 & 45.71 & 24.54 & 10.64 & 3.39 & 
 41.46 & 36.67 & 31.28 \\ \midrule

 \multirow{3}{*}{30} 
 & 105 & 69.03 & 51.71 & 40.76 & 35.06 & 17.74 & 6.79 & 
 28.92 & 22.46 & 18.89 \\
 & 89 & 55.42 & 41.57 & 30.77 & 28.05 & 14.19 & 3.39 & 
 28.10 & 21.60 & 18.89 \\
 & 79 & 49.43 & 35.53 & 28.28 & 24.54 & 10.64 & 3.39 & 
 26.17 & 21.10 & 18.89 \\ \midrule

 \multirow{3}{*}{10} 
 & 105 & 45.51 & 28.19 & 17.24 & 35.06 & 17.74 & 6.79 & 
 15.10 & 13.28 & 10.10 \\
 & 89 & 38.00 & 24.14 & 13.35 & 28.05 & 14.19 & 3.39 & 
 15.10 & 13.28 & 10.10 \\
 & 79 & 32.01 & 18.11 & 10.86 & 24.54 & 10.64 & 3.39 & 
 14.82 & 13.03 & 10.10 \\ \midrule

 \multirow{3}{*}{1} 
 & 105 & 40.29 & 22.96 & 12.02 & 35.06 & 17.74 & 6.79 & 
 6.53 & 6.67 & 5.10 \\
 & 89 & 32.53 & 18.67 & 7.87 & 28.05 & 14.19 & 3.39 & 
 6.53 & 6.67 & 5.10  \\
 & 79 & 28.42 & 14.52 & 7.28 & 24.54 & 10.64 & 3.39 & 
 6.53 & 5.96 & 5.00 \\ \bottomrule
 
 \end{tabular}%
}
\caption{Performance scores for ResNet50 with ImageNet-21k weights on UltraMNIST and its ....EDPE = Effective data per epoch, method=GradMatch}
\label{tab:rn50-um-grad}
\end{table}



\begin{table}[h]
\centering
\resizebox{0.9\textwidth}{!}{%
\begin{tabular}{@{}cc | ccc | ccc | ccc@{}}
\toprule
\textbf{EDPE (\%)} & \textbf{Epoch} & \multicolumn{3}{c|}{\textbf{Total Time (Mins)}} & \multicolumn{3}{c|}{\textbf{Total Selection Time (Mins)}} & \multicolumn{3}{c}{\textbf{Accuracy}} \\ \midrule

\multirow{3}{*}{100} 
& 105 &  & 114.99 &  &  & - &  &  & 61.28 &  \\
& 89 &  & 98.56 &  &  & - &  &  & 61.28 &  \\
 & 79 &  & 87.36 &  &  & - &  &  & 61.21 &  \\
 \midrule

 \multicolumn{1}{c}{} & \multicolumn{1}{c|}{} & \textit{SSI=10} & \textit{SSI=20} & \textit{SSI=50} & \textit{SSI=10} & \textit{SSI=20} & \textit{SSI=50} & \textit{SSI=10} & \textit{SSI=20} & \textit{SSI=50}\\ \midrule

\multirow{3}{*}{80} 
& 105 & 142.94 & 111.20 & 100.78 & 51.47 & 19.73 & 6.78 &
53.92 & 53.00 & - \\
 & 89 & 119.13 & 93.74 & 82.61 & 41.18 & 15.78 & 3.39 &
 53.92 & 52.67 & - \\
 & 79 & 105.02 & 80.83 & 73.65 & 36.03 & 11.84 & 3.39 &
 53.92 & 50.82 & -  \\ \midrule

 \multirow{3}{*}{50} 
 & 105 & 108.97 & 77.23 & 64.28 & 51.47 & 19.73 & 6.78 &
 40.53 & 34.85 & 31.00 \\
 & 89 & 90.41 & 65.02 & 52.62 & 41.18 & 15.78 & 3.39 & 
 40.53 & 34.82 & 31.00 \\
 & 79 & 79.54 & 55.35 & 46.90 & 36.03 & 11.84 & 3.39 & 
 39.53 & 34.35 & 31.00 \\ \midrule

 \multirow{3}{*}{30} 
 & 105 & 85.45 & 53.71 & 40.76 & 51.47 & 19.73 & 6.78 & 
 28.25 & 22.78 & 20.71 \\
 & 89 & 70.10 & 44.71 & 32.31 & 41.18 & 15.78 & 3.39 & 
 28.25 & 22.78 & 20.42 \\
 & 79 & 61.81 & 37.62 & 29.18 & 36.03 & 11.84 & 3.39 & 
 27.25 & 11.84 & 20.42 \\ \midrule

 \multirow{3}{*}{10} 
 & 105 & 61.92 & 30.19 & 17.24 & 51.47 & 19.73 & 6.78 & 
 17.50 & 13.64 & 11.46 \\
 & 89 & 50.14 & 24.75 & 12.35 & 41.18 & 15.78 & 3.39 & 
 16.85 & 13.64 & 11.46 \\
 & 79 & 43.99 & 19.80 & 11.35 & 36.03 & 11.84 & 3.39 & 
 16.85 & 13.64 & 11.46 \\ \midrule

 \multirow{3}{*}{1} 
 & 105 & 56.70 & 24.96 & 12.01 & 51.47 & 19.73 & 6.78 & 
 9.50 & 7.82 & 6.03 \\
 & 89 & 45.66 & 20.27 & 7.87 & 41.18 & 15.78 & 3.39 & 
 9.50 & 7.67 & 6.03  \\
 & 79 & 39.91 & 15.72 & 7.27 & 36.03 & 11.84 & 3.39 & 
 8.42 & 7.25 & 6.03 \\ \bottomrule
 
 \end{tabular}%
}
\caption{Performance scores for ResNet50 with ImageNet-21k weights on UltraMNIST and its ....EDPE = Effective data per epoch, method=GLISTER}
\label{tab:rn50-um-glis}
\end{table}


\begin{table}[h]
\centering
\resizebox{0.9\textwidth}{!}{%
\begin{tabular}{@{}cc | ccc | ccc | ccc@{}}
\toprule
\textbf{EDPE (\%)} & \textbf{Epoch} & \multicolumn{3}{c|}{\textbf{Total Time (Mins)}} & \multicolumn{3}{c|}{\textbf{Total Selection Time (Mins)}} & \multicolumn{3}{c}{\textbf{Accuracy}} \\ \midrule

\multirow{3}{*}{100} 
& 105 &  & 239.53 &  &  & - &  &  & 40.35 &  \\
& 89 &  & 205.31 &  &  & - &  &  & 40.35 &  \\
 & 79 &  & 181.98 &  &  & - &  &  & 40.35 &  \\
 \midrule

 \multicolumn{1}{c}{} & \multicolumn{1}{c|}{} & \textit{SSI=10} & \textit{SSI=20} & \textit{SSI=50} & \textit{SSI=10} & \textit{SSI=20} & \textit{SSI=50} & \textit{SSI=10} & \textit{SSI=20} & \textit{SSI=50}\\ \midrule

\multirow{3}{*}{80} 
& 105 & 91.47 & 91.47 & 91.47 & 0.00 & 0.00 & 0.00 & 
48.75 & 49.35 & 49.75 \\
 & 89 & 77.95 & 77.95 & 77.95 & 0.00 & 0.00 & 0.00 & 
 48.71 & 49.35 & 49.71 \\
 & 79 & 68.99 & 68.99 & 68.99 & 0.00 & 0.00 & 0.00 & 
 48.42 & 49.32 & 49.60  \\ \midrule

 \multirow{3}{*}{50} 
 & 105 & 57.49 & 57.49 & 57.49 & 0.00 & 0.00 & 0.00 & 
 30.25 & 28.75 & 29.75 \\
 & 89 & 49.23 & 49.23 & 49.23 & 0.00 & 0.00 & 0.00 & 
 30.25 & 28.75 & 29.71 \\
 & 79 & 43.50 & 43.50 & 43.50 & 0.00 & 0.00 & 0.00 & 
 30.25 & 28.75 & 29.67 \\ \midrule

 \multirow{3}{*}{30} 
 & 105 & 33.97 & 33.97 & 33.97 & 0.00 & 0.00 & 0.00 & 
 20.07 & 19.17 & 19.14 \\
 & 89 & 28.92 & 28.92 & 28.92 & 0.00 & 0.00 & 0.00 & 
 20.07 & 19.17 & 19.14 \\
 & 79 & 25.78 & 25.78 & 25.78 & 0.00 & 0.00 & 0.00 & 
 20.07 & 19.17 & 19.14 \\ \midrule

 \multirow{3}{*}{10} 
 & 105 & 10.45 & 10.45 & 10.45 & 0.00 & 0.00 & 0.00 & 
 12.46 & 12.89 & 11.89 \\
 & 89 & 8.96 & 8.96 & 8.96 & 0.00 & 0.00 & 0.00 & 
 12.46 & 12.89 & 11.89 \\
 & 79 & 7.96 & 7.96 & 7.96 & 0.00 & 0.00 & 0.00 & 
 12.46 & 12.89 & 11.89 \\ \midrule

 \multirow{3}{*}{1} 
 & 105 & 5.22 & 5.22 & 5.22 & 0.00 & 0.00 & 0.00 & 
 5.75 & 6.89 & 5.67 \\
 & 89 & 4.48 & 4.48 & 4.48 & 0.00 & 0.00 & 0.00 & 
 5.75 & 6.89 & 5.67  \\
 & 79 & 3.88 & 3.88 & 3.88 & 0.00 & 0.00 & 0.00 & 
 5.64 & 6.82 & 5.53 \\ \bottomrule
 
 \end{tabular}%
}
\caption{Performance scores for ResNet50 with ImageNet-21k weights on UltraMNIST and its ....EDPE = Effective data per epoch, method=Random}
\label{tab:rn50-um-rand}
\end{table}

\subsubsection{CRAIG, GradMatch, GLISTER and Random Results on Medical Dataset}

We further expanded our experiments vision to a medical dataset, to study the effect of pretraining when using ViT\_B16 and ResNet50 with different coreset sizes. From the tables [\ref{tab:vit-ap-craig}, \ref{tab:vit-ap-grad}, \ref{tab:vit-ap-glis}, \ref{tab:vit-ap-rand}, \ref{tab:rn50-ap-craig}, \ref{tab:rn50-ap-grad}, \ref{tab:rn50-ap-glis}, \ref{tab:rn50-ap-rand}] and the generated Fig. \ref{fig:ap-coresets} from these tables assisted us to show that without pretraining CNNs outperform on Transformers for different coreset values using random selection. It was also intriguing to see that random selection is better or almost at par with GradMatch, which is itself superior to other coreset methods. 
\begin{table}[h]
\centering
\resizebox{0.9\textwidth}{!}{%
\begin{tabular}{@{}cc | ccc | ccc | ccc@{}}
\toprule
\textbf{EDPE (\%)} & \textbf{Epoch} & \multicolumn{3}{c|}{\textbf{Total Time (Mins)}} & \multicolumn{3}{c|}{\textbf{Total Selection Time (Mins)}} & \multicolumn{3}{c}{\textbf{Quadratic $\kappa$}} \\ \midrule

\multirow{3}{*}{100} 
& 105 &  & 16.59 &  &  & - &  &  & 0.88 &  \\
& 89 &  & 14.22 &  &  & - &  &  & 0.88 &  \\
 & 79 &  & 12.64 &  &  & - &  &  &  0.88 &  \\
 \midrule

 \multicolumn{1}{c}{} & \multicolumn{1}{c|}{} & \textit{SSI=10} & \textit{SSI=20} & \textit{SSI=50} & \textit{SSI=10} & \textit{SSI=20} & \textit{SSI=50} & \textit{SSI=10} & \textit{SSI=20} & \textit{SSI=50}\\ \midrule

\multirow{3}{*}{80} 
& 105 & 15.91 & 14.88 & 14.25 & 2.08 & 1.05 & 0.42 &
0.87 & 0.89 & 0.90 \\
 & 89 & 13.52 & 12.69 & 12.06 & 1.67 & 0.84 & 0.21 &
 0.87 & 0.89 & 0.90 \\
 & 79 & 11.99 & 11.16 & 10.74 & 1.46 & 0.63 & 0.21 &
 0.87 & 0.89 & 0.90  \\ \midrule

 \multirow{3}{*}{50} 
 & 105 & 10.38 & 9.35 & 8.72 & 2.08 & 1.05 & 0.42 &
 0.87 & 0.86 & 0.88 \\
 & 89 & 8.78 & 7.95 & 7.32 & 1.67 & 0.84 & 0.21 & 
 0.87 & 0.86 & 0.88 \\
 & 79 & 7.78 & 6.95 & 6.53 & 1.46 & 0.63 & 0.21 & 
 0.87 & 0.86 & 0.88 \\ \midrule

 \multirow{3}{*}{30} 
 & 105 & 7.61 & 6.58 & 5.95 & 2.08 & 1.05 & 0.42 & 
 0.86 & 0.86 & 0.87 \\
 & 89 & 6.41 & 5.58 & 4.95 & 1.67 & 0.84 & 0.21 & 
 0.86 & 0.86 & 0.87 \\
 & 79 & 5.67 & 4.84 & 4.42 & 1.46 & 0.63 & 0.21 & 
 0.86 & 0.86 & 0.87 \\ \midrule

 \multirow{3}{*}{10} 
 & 105 & 4.16 & 3.12 & 2.50 & 2.08 & 1.05 & 0.42 & 
 0.86 & 0.84 & 0.85 \\
 & 89 & 3.44 & 2.62 & 1.99 & 1.67 & 0.84 & 0.21 & 
 0.85 & 0.83 & 0.85 \\
 & 79 & 3.04 & 2.21 & 1.79 & 1.46 & 0.63 & 0.21 & 
 0.85 & 0.83 & 0.85 \\ \midrule

 \multirow{3}{*}{1} 
 & 105 & 2.26 & 1.22 & 0.60 & 2.08 & 1.05 & 0.42 & 
 0.73 & 0.72 & 0.66 \\
 & 89 & 1.81 & 0.99 & 0.36 & 1.67 & 0.84 & 0.21 & 
 0.73 & 0.72 & 0.66  \\
 & 79 & 1.59 & 0.76 & 0.34 & 1.46 & 0.63 & 0.21 & 
 0.71 & 0.72 & 0.66 \\ \bottomrule
 
 \end{tabular}%
}
\caption{Performance scores for ViT\_B16 with ImageNet-21k weights on Medical Dataset APTOS-2019 and its ....EDPE = Effective data per epoch, method=CRAIG}
\label{tab:vit-ap-craig}
\end{table}

\begin{table}[h]
\centering
\resizebox{0.9\textwidth}{!}{%
\begin{tabular}{@{}cc | ccc | ccc | ccc@{}}
\toprule
\textbf{EDPE (\%)} & \textbf{Epoch} & \multicolumn{3}{c|}{\textbf{Total Time (Mins)}} & \multicolumn{3}{c|}{\textbf{Total Selection Time (Mins)}} & \multicolumn{3}{c}{\textbf{Quadratic $\kappa$}} \\ \midrule

\multirow{3}{*}{100} 
& 105 &  & 16.59 &  &  & - &  &  & 0.88 &  \\
& 89 &  & 14.22 &  &  & - &  &  & 0.88 &  \\
 & 79 &  & 12.64 &  &  & - &  &  &  0.88 &  \\
 \midrule

 \multicolumn{1}{c}{} & \multicolumn{1}{c|}{} & \textit{SSI=10} & \textit{SSI=20} & \textit{SSI=50} & \textit{SSI=10} & \textit{SSI=20} & \textit{SSI=50} & \textit{SSI=10} & \textit{SSI=20} & \textit{SSI=50}\\ \midrule

\multirow{3}{*}{80} 
& 105 & 15.61 & 14.83 & 14.20 & 1.79 & 1.00 & 0.37 &
0.88 & 0.88 & 0.88 \\
 & 89 & 13.28 & 12.65 & 12.04 & 1.43 & 0.80 & 0.18 &
 0.88 & 0.88 & 0.88 \\
 & 79 & 11.78 & 11.13 & 10.72 & 1.25 & 0.60 & 0.18 &
 0.88 & 0.88 & 0.88  \\ \midrule

 \multirow{3}{*}{50} 
 & 105 & 10.08 & 9.30 & 8.67 & 1.79 & 1.00 & 0.37 &
 0.89 & 0.88 & 0.88 \\
 & 89 & 8.54 & 7.91 & 7.30 & 1.43 & 0.80 & 0.18 & 
 0.89 & 0.88 & 0.88 \\
 & 79 & 7.57 & 6.92 & 6.51 & 1.25 & 0.60 & 0.18 & 
 0.89 & 0.88 & 0.88 \\ \midrule

 \multirow{3}{*}{30} 
 & 105 & 7.32 & 6.53 & 5.90 & 1.79 & 1.00 & 0.37 & 
 0.88 & 0.87 & 0.88 \\
 & 89 & 6.17 & 5.54 & 4.93 & 1.43 & 0.80 & 0.18 & 
 0.88 & 0.87 & 0.88 \\
 & 79 & 5.46 & 4.81 & 4.40 & 1.25 & 0.60 & 0.18 & 
 0.88 & 0.87 & 0.88 \\ \midrule

 \multirow{3}{*}{10} 
 & 105 & 3.86 & 3.08 & 2.45 & 1.79 & 1.00 & 0.37 & 
 0.85 & 0.84 & 0.85 \\
 & 89 & 3.21 & 2.58 & 1.96 & 1.43 & 0.80 & 0.18 & 
 0.85 & 0.84 & 0.85 \\
 & 79 & 2.83 & 2.18 & 1.76 & 1.25 & 0.60 & 0.18 & 
 0.85 & 0.84 & 0.85 \\ \midrule

 \multirow{3}{*}{1} 
 & 105 & 1.96 & 1.17 & 0.55 & 1.79 & 1.00 & 0.37 & 
 0.66 & 0.74 & 0.70 \\
 & 89 & 1.58 & 0.95 & 0.33 & 1.43 & 0.80 & 0.18 & 
 0.66 & 0.74 & 0.70  \\
 & 79 & 1.38 & 0.73 & 0.32 & 1.25 & 0.60 & 0.18 & 
 0.66 & 0.74 & 0.70 \\ \bottomrule
 
 \end{tabular}%
}
\caption{Performance scores for ViT\_B16 with ImageNet-21k weights on Medical Dataset APTOS-2019 and its ....EDPE = Effective data per epoch, method=GradMatch}
\label{tab:vit-ap-grad}
\end{table}

\begin{table}[h]
\centering
\resizebox{0.9\textwidth}{!}{%
\begin{tabular}{@{}cc | ccc | ccc | ccc@{}}
\toprule
\textbf{EDPE (\%)} & \textbf{Epoch} & \multicolumn{3}{c|}{\textbf{Total Time (Mins)}} & \multicolumn{3}{c|}{\textbf{Total Selection Time (Mins)}} & \multicolumn{3}{c}{\textbf{Quadratic $\kappa$}} \\ \midrule

\multirow{3}{*}{100} 
& 105 &  & 16.59 &  &  & - &  &  & 0.88 &  \\
& 89 &  & 14.22 &  &  & - &  &  & 0.88 &  \\
 & 79 &  & 12.64 &  &  & - &  &  &  0.88 &  \\
 \midrule

 \multicolumn{1}{c}{} & \multicolumn{1}{c|}{} & \textit{SSI=10} & \textit{SSI=20} & \textit{SSI=50} & \textit{SSI=10} & \textit{SSI=20} & \textit{SSI=50} & \textit{SSI=10} & \textit{SSI=20} & \textit{SSI=50}\\ \midrule

\multirow{3}{*}{80} 
& 105 & 16.46 & 15.77 & 14.51 & 3.41 & 1.94 & 0.68 &
0.88 & 0.88 & 0.88 \\
 & 89 & 13.95 & 13.40 & 12.19 & 2.73 & 1.55 & 0.34 &
 0.88 & 0.88 & 0.88 \\
 & 79 & 12.37 & 11.70 & 10.87 & 2.38 & 1.16 & 0.34 &
 0.88 & 0.88 & 0.88  \\ \midrule

 \multirow{3}{*}{50} 
 & 105 & 11.71 & 10.24 & 8.98 & 3.41 & 1.94 & 0.68 &
 0.87 & 0.88 & 0.87 \\
 & 89 & 9.84 & 8.66 & 7.45 & 2.73 & 1.55 & 0.34 & 
 0.87 & 0.88 & 0.87 \\
 & 79 & 8.71 & 7.48 & 6.66 & 2.38 & 1.16 & 0.34 & 
 0.87 & 0.88 & 0.87 \\ \midrule

 \multirow{3}{*}{30} 
 & 105 & 8.94 & 7.47 & 6.22 & 3.41 & 1.94 & 0.68 & 
 0.86 & 0.87 & 0.87 \\
 & 89 & 7.47 & 6.29 & 5.08 & 2.73 & 1.55 & 0.34 & 
 0.86 & 0.87 & 0.87 \\
 & 79 & 6.60 & 5.38 & 4.55 & 2.38 & 1.16 & 0.34 & 
 0.86 & 0.87 & 0.87 \\ \midrule

 \multirow{3}{*}{10} 
 & 105 & 5.48 & 4.01 & 2.76 & 3.41 & 1.94 & 0.68 & 
 0.75 & 0.76 & 0.81 \\
 & 89 & 4.50 & 3.33 & 2.12 & 2.73 & 1.55 & 0.34 & 
 0.75 & 0.76 & 0.81 \\
 & 79 & 3.96 & 2.74 & 1.92 & 2.38 & 1.16 & 0.34 & 
 0.75 & 0.76 & 0.81 \\ \midrule

 \multirow{3}{*}{1} 
 & 105 & 3.58 & 2.11 & 0.86 & 3.41 & 1.94 & 0.68 & 
 0.70 & 0.67 & 0.54 \\
 & 89 & 2.87 & 1.70 & 0.49 & 2.73 & 1.55 & 0.34 & 
 0.58 & 0.67 & 0.54  \\
 & 79 & 2.52 & 1.29 & 0.47 & 2.38 & 1.16 & 0.34 & 
 0.58 & 0.67 & 0.54 \\ \bottomrule
 
 \end{tabular}%
}
\caption{Performance scores for ViT\_B16 with ImageNet-21k weights on Medical Dataset APTOS-2019 and its ....EDPE = Effective data per epoch, method=GLISTER}
\label{tab:vit-ap-glis}
\end{table}

\begin{table}[h]
\centering
\resizebox{0.9\textwidth}{!}{%
\begin{tabular}{@{}cc | ccc | ccc | ccc@{}}
\toprule
\textbf{EDPE (\%)} & \textbf{Epoch} & \multicolumn{3}{c|}{\textbf{Total Time (Mins)}} & \multicolumn{3}{c|}{\textbf{Total Selection Time (Mins)}} & \multicolumn{3}{c}{\textbf{Quadratic $\kappa$}} \\ \midrule

\multirow{3}{*}{100} 
& 105 &  & 16.59 &  &  & - &  &  & 0.88 &  \\
& 89 &  & 14.22 &  &  & - &  &  & 0.88 &  \\
 & 79 &  & 12.64 &  &  & - &  &  &  0.88 &  \\
 \midrule

 \multicolumn{1}{c}{} & \multicolumn{1}{c|}{} & \textit{SSI=10} & \textit{SSI=20} & \textit{SSI=50} & \textit{SSI=10} & \textit{SSI=20} & \textit{SSI=50} & \textit{SSI=10} & \textit{SSI=20} & \textit{SSI=50}\\ \midrule

\multirow{3}{*}{80} 
& 105 & 13.82 & 13.82 & 13.82 & 0.0 & 0.0 & 0.0 &
0.88 & 0.88 & 0.89 \\
 & 89 & 11.85 & 11.85 & 11.85 & 0.0 & 0.0 & 0.0 &
 0.88 & 0.88 & 0.89 \\
 & 79 & 10.53 & 10.53 & 10.53 & 0.0 & 0.0 & 0.0 &
 0.88 & 0.88 & 0.89  \\ \midrule

 \multirow{3}{*}{50} 
 & 105 & 8.29 & 8.29 & 8.29 & 0.0 & 0.0 & 0.0 &
 0.88 & 0.88 & 0.87 \\
 & 89 & 7.11 & 7.11 & 7.11 & 0.0 & 0.0 & 0.0 & 
 0.88 & 0.88 & 0.87 \\
 & 79 & 6.32 & 6.32 & 6.32 & 0.0 & 0.0 & 0.0 & 
 0.88 & 0.88 & 0.87 \\ \midrule

 \multirow{3}{*}{30} 
 & 105 & 5.53 & 5.53 & 5.53 & 0.0 & 0.0 & 0.0 & 
 0.87 & 0.87 & 0.87 \\
 & 89 & 4.74 & 4.74 & 4.74 & 0.0 & 0.0 & 0.0 & 
 0.87 & 0.87 & 0.87 \\
 & 79 & 4.21 & 4.21 & 4.21 & 0.0 & 0.0 & 0.0 & 
 0.87 & 0.87 & 0.87 \\ \midrule

 \multirow{3}{*}{10} 
 & 105 & 2.07 & 2.07 & 2.07 & 0.0 & 0.0 & 0.0 & 
 0.83 & 0.83 & 0.86 \\
 & 89 & 1.77 & 1.77 & 1.77 & 0.0 & 0.0 & 0.0 & 
 0.83 & 0.83 & 0.86 \\
 & 79 & 1.58 & 1.58 & 1.58 & 0.0 & 0.0 & 0.0 & 
 0.83 & 0.83 & 0.85 \\ \midrule

 \multirow{3}{*}{1} 
 & 105 & 0.17 & 0.17 & 0.17 & 0.0 & 0.0 & 0.0 & 
 0.69 & 0.66 & 0.75 \\
 & 89 & 0.14 & 0.14 & 0.14 & 0.0 & 0.0 & 0.0 & 
 0.69 & 0.66 & 0.75  \\
 & 79 & 0.13 & 0.13 & 0.13 & 0.0 & 0.0 & 0.0 & 
 0.69 & 0.66 & 0.75 \\ \bottomrule
 
 \end{tabular}%
}
\caption{Performance scores for ViT\_B16 with ImageNet-21k weights on Medical Dataset APTOS-2019 and its ....EDPE = Effective data per epoch, method=Random}
\label{tab:vit-ap-rand}
\end{table}


\begin{table}[h]
\centering
\resizebox{0.9\textwidth}{!}{%
\begin{tabular}{@{}cc | ccc | ccc | ccc@{}}
\toprule
\textbf{EDPE (\%)} & \textbf{Epoch} & \multicolumn{3}{c|}{\textbf{Total Time (Mins)}} & \multicolumn{3}{c|}{\textbf{Total Selection Time (Mins)}} & \multicolumn{3}{c}{\textbf{Quadratic $\kappa$}} \\ \midrule

\multirow{3}{*}{100} 
& 105 &  & 10.73 &  &  & - &  &  & 0.88 &  \\
& 89 &  & 9.20 &  &  & - &  &  & 0.88 &  \\
 & 79 &  & 8.18 &  &  & - &  &  & 0.88 &  \\
 \midrule

 \multicolumn{1}{c}{} & \multicolumn{1}{c|}{} & \textit{SSI=10} & \textit{SSI=20} & \textit{SSI=50} & \textit{SSI=10} & \textit{SSI=20} & \textit{SSI=50} & \textit{SSI=10} & \textit{SSI=20} & \textit{SSI=50}\\ \midrule

\multirow{3}{*}{80} 
& 105 & 10.76 & 9.86 & 9.32 & 1.81 & 0.91 & 0.37 &
0.90 & 0.90 & 0.90 \\
 & 89 & 9.12 & 8.40 & 7.85 & 1.45 & 0.73 & 0.18 &
 0.90 & 0.89 & 0.90 \\
 & 79 & 8.08 & 7.37 & 7.00 & 1.26 & 0.55 & 0.18 &
 0.89 & 0.89 & 0.90  \\ \midrule

 \multirow{3}{*}{50} 
 & 105 & 7.18 & 6.28 & 5.74 & 1.81 & 0.91 & 0.37 &
 0.88 & 0.88 & 0.88 \\
 & 89 & 6.05 & 5.33 & 4.79 & 1.45 & 0.73 & 0.18 & 
 0.88 & 0.88 & 0.88 \\
 & 79 & 5.36 & 4.64 & 4.27 & 1.26 & 0.55 & 0.18 & 
 0.87 & 0.88 & 0.88 \\ \midrule

 \multirow{3}{*}{30} 
 & 105 & 5.39 & 4.49 & 3.95 & 1.81 & 0.91 & 0.37 & 
 0.87 & 0.87 & 0.88 \\
 & 89 & 4.51 & 3.80 & 3.25 & 1.45 & 0.73 & 0.18 & 
 0.83 & 0.85 & 0.88 \\
 & 79 & 3.99 & 3.27 & 2.91 & 1.26 & 0.55 & 0.18 & 
 0.83 & 0.84 & 0.87 \\ \midrule

 \multirow{3}{*}{10} 
 & 105 & 3.15 & 2.26 & 1.71 & 1.81 & 0.91 & 0.37 & 
 0.86 & 0.82 & 0.84 \\
 & 89 & 2.60 & 1.88 & 1.33 & 1.45 & 0.73 & 0.18 & 
 0.85 & 0.82 & 0.84 \\
 & 79 & 2.29 & 1.57 & 1.21 & 1.26 & 0.55 & 0.18 & 
 0.85 & 0.82 & 0.84 \\ \midrule

 \multirow{3}{*}{1} 
 & 105 & 1.92 & 1.03 & 0.48 & 1.81 & 0.91 & 0.37 & 
 0.81 & 0.74 & 0.84 \\
 & 89 & 1.54 & 0.83 & 0.28 & 1.45 & 0.73 & 0.18 & 
 0.81 & 0.71 & 0.83  \\
 & 79 & 1.35 & 0.63 & 0.27 & 1.26 & 0.55 & 0.18 & 
 0.79 & 0.69 & 0.83 \\ \bottomrule
 
 \end{tabular}%
}
\caption{Performance scores for ResNet50 with ImageNet-21k weights on Medical Dataset APTOS-2019 and its ....EDPE = Effective data per epoch, method=CRAIG}
\label{tab:rn50-ap-craig}
\end{table}


\begin{table}[h]
\centering
\resizebox{0.9\textwidth}{!}{%
\begin{tabular}{@{}cc | ccc | ccc | ccc@{}}
\toprule
\textbf{EDPE (\%)} & \textbf{Epoch} & \multicolumn{3}{c|}{\textbf{Total Time (Mins)}} & \multicolumn{3}{c|}{\textbf{Total Selection Time (Mins)}} & \multicolumn{3}{c}{\textbf{Quadratic $\kappa$}} \\ \midrule

\multirow{3}{*}{100} 
& 105 &  & 10.73 &  &  & - &  &  & 0.88 &  \\
& 89 &  & 9.20 &  &  & - &  &  & 0.88 &  \\
 & 79 &  & 8.18 &  &  & - &  &  & 0.88 &  \\
 \midrule

 \multicolumn{1}{c}{} & \multicolumn{1}{c|}{} & \textit{SSI=10} & \textit{SSI=20} & \textit{SSI=50} & \textit{SSI=10} & \textit{SSI=20} & \textit{SSI=50} & \textit{SSI=10} & \textit{SSI=20} & \textit{SSI=50}\\ \midrule

\multirow{3}{*}{80} 
& 105 & 10.46 & 9.71 & 9.26 & 1.51 & 0.76 & 0.31 &
0.90 & 0.89 & 0.89 \\
 & 89 & 8.88 & 8.28 & 7.82 & 1.21 & 0.61 & 0.15 &
 0.90 & 0.89 & 0.89 \\
 & 79 & 7.88 & 7.27 & 6.97 & 1.06 & 0.45 & 0.15 &
 0.90 & 0.89 & 0.89  \\ \midrule

 \multirow{3}{*}{50} 
 & 105 & 6.88 & 6.13 & 5.68 & 1.51 & 0.76 & 0.31 &
 0.88 & 0.89 & 0.89 \\
 & 89 & 5.81 & 5.21 & 4.75 & 1.21 & 0.61 & 0.15 & 
 0.88 & 0.89 & 0.89 \\
 & 79 & 5.15 & 4.54 & 4.24 & 1.06 & 0.45 & 0.15 & 
 0.88 & 0.89 & 0.89 \\ \midrule

 \multirow{3}{*}{30} 
 & 105 & 5.09 & 4.34 & 3.89 & 1.51 & 0.76 & 0.31 & 
 0.88 & 0.88 & 0.88 \\
 & 89 & 4.28 & 3.67 & 3.22 & 1.21 & 0.61 & 0.15 & 
 0.88 & 0.88 & 0.88 \\
 & 79 & 3.79 & 3.18 & 2.88 & 1.06 & 0.45 & 0.15 & 
 0.87 & 0.88 & 0.88 \\ \midrule

 \multirow{3}{*}{10} 
 & 105 & 2.86 & 2.10 & 1.65 & 1.51 & 0.76 & 0.31 & 
 0.86 & 0.87 & 0.85 \\
 & 89 & 2.36 & 1.76 & 1.30 & 1.21 & 0.61 & 0.15 & 
 0.86 & 0.87 & 0.85 \\
 & 79 & 2.08 & 1.48 & 1.17 & 1.06 & 0.45 & 0.15 & 
 0.86 & 0.86 & 0.85 \\ \midrule

 \multirow{3}{*}{1} 
 & 105 & 1.62 & 0.87 & 0.42 & 1.51 & 0.76 & 0.31 & 
 0.79 & 0.80 & 0.73 \\
 & 89 & 1.31 & 0.70 & 0.25 & 1.21 & 0.61 & 0.15 & 
 0.79 & 0.80 & 0.73  \\
 & 79 & 1.14 & 0.54 & 0.24 & 1.06 & 0.45 & 0.15 & 
 0.79 & 0.80 & 0.73 \\ \bottomrule
 
 \end{tabular}%
}
\caption{Performance scores for ResNet50 with ImageNet-21k weights on Medical Dataset APTOS-2019 and its ....EDPE = Effective data per epoch, method=GradMatch}
\label{tab:rn50-ap-grad}
\end{table}


\begin{table}[h]
\centering
\resizebox{0.9\textwidth}{!}{%
\begin{tabular}{@{}cc | ccc | ccc | ccc@{}}
\toprule
\textbf{EDPE (\%)} & \textbf{Epoch} & \multicolumn{3}{c|}{\textbf{Total Time (Mins)}} & \multicolumn{3}{c|}{\textbf{Total Selection Time (Mins)}} & \multicolumn{3}{c}{\textbf{Quadratic $\kappa$}} \\ \midrule

\multirow{3}{*}{100} 
& 105 &  & 10.73 &  &  & - &  &  & 0.88 &  \\
& 89 &  & 9.20 &  &  & - &  &  & 0.88 &  \\
 & 79 &  & 8.18 &  &  & - &  &  & 0.88 &  \\
 \midrule

 \multicolumn{1}{c}{} & \multicolumn{1}{c|}{} & \textit{SSI=10} & \textit{SSI=20} & \textit{SSI=50} & \textit{SSI=10} & \textit{SSI=20} & \textit{SSI=50} & \textit{SSI=10} & \textit{SSI=20} & \textit{SSI=50}\\ \midrule

\multirow{3}{*}{80} 
& 105 & 12.46 & 10.70 & 9.65 & 3.51 & 1.76 & 0.70 &
0.90 & 0.90 & 0.89 \\
 & 89 & 10.48 & 9.07 & 8.02 & 2.81 & 1.40 & 0.35 &
 0.90 & 0.90 & 0.89 \\
 & 79 & 9.28 & 7.87 & 7.17 & 2.46 & 1.05 & 0.35 &
 0.89 & 0.89 & 0.89  \\ \midrule

 \multirow{3}{*}{50} 
 & 105 & 8.88 & 7.13 & 6.07 & 3.51 & 1.76 & 0.70 &
 0.88 & 0.88 & 0.87 \\
 & 89 & 7.41 & 6.01 & 4.95 & 2.81 & 1.40 & 0.35 & 
 0.86 & 0.88 & 0.87 \\
 & 79 & 6.55 & 5.14 & 4.44 & 2.46 & 1.05 & 0.35 & 
 0.86 & 0.85 & 0.87 \\ \midrule

 \multirow{3}{*}{30} 
 & 105 & 7.09 & 5.34 & 4.28 & 3.51 & 1.76 & 0.70 & 
 0.73 & 0.85 & 0.88 \\
 & 89 & 5.88 & 4.47 & 3.42 & 2.81 & 1.40 & 0.35 & 
 0.73 & 0.85 & 0.88 \\
 & 79 & 5.18 & 3.78 & 3.08 & 2.46 & 1.05 & 0.35 & 
 0.73 & 0.85 & 0.88 \\ \midrule

 \multirow{3}{*}{10} 
 & 105 & 4.85 & 3.10 & 2.05 & 3.51 & 1.76 & 0.70 & 
 0.78 & 0.81 & 0.85 \\
 & 89 & 3.96 & 2.55 & 1.50 & 2.81 & 1.40 & 0.35 & 
 0.78 & 0.81 & 0.83 \\
 & 79 & 3.48 & 2.07 & 1.37 & 2.46 & 1.05 & 0.35 & 
 0.78 & 0.80 & 0.83 \\ \midrule

 \multirow{3}{*}{1} 
 & 105 & 3.62 & 1.87 & 0.82 & 3.51 & 1.76 & 0.70 & 
 0.64 & 0.68 & 0.74 \\
 & 89 & 2.90 & 1.50 & 0.45 & 2.81 & 1.40 & 0.35 & 
 0.64 & 0.62 & 0.74  \\
 & 79 & 2.54 & 1.14 & 0.43 & 2.46 & 1.05 & 0.35 & 
 0.64 & 0.62 & 0.74 \\ \bottomrule
 
 \end{tabular}%
}
\caption{Performance scores for ResNet50 with ImageNet-21k weights on Medical Dataset APTOS-2019 and its ....EDPE = Effective data per epoch, method=GLISTER}
\label{tab:rn50-ap-glis}
\end{table}


\begin{table}[h]
\centering
\resizebox{0.9\textwidth}{!}{%
\begin{tabular}{@{}cc | ccc | ccc | ccc@{}}
\toprule
\textbf{EDPE (\%)} & \textbf{Epoch} & \multicolumn{3}{c|}{\textbf{Total Time (Mins)}} & \multicolumn{3}{c|}{\textbf{Total Selection Time (Mins)}} & \multicolumn{3}{c}{\textbf{Quadratic $\kappa$}} \\ \midrule

\multirow{3}{*}{100} 
& 105 &  & 10.73 &  &  & - &  &  & 0.88 &  \\
& 89 &  & 9.20 &  &  & - &  &  & 0.88 &  \\
 & 79 &  & 8.18 &  &  & - &  &  & 0.88 &  \\
 \midrule

 \multicolumn{1}{c}{} & \multicolumn{1}{c|}{} & \textit{SSI=10} & \textit{SSI=20} & \textit{SSI=50} & \textit{SSI=10} & \textit{SSI=20} & \textit{SSI=50} & \textit{SSI=10} & \textit{SSI=20} & \textit{SSI=50}\\ \midrule

\multirow{3}{*}{80} 
& 105 & 8.94 & 8.94 & 8.94 & 0.0 & 0.0 & 0.0 &
0.89 & 0.89 & 0.89 \\
 & 89 & 7.67 & 7.67 & 7.67 & 0.0 & 0.0 & 0.0 &
 0.89 & 0.89 & 0.89 \\
 & 79 & 6.81 & 6.81 & 6.81 & 0.0 & 0.0 & 0.0 &
 0.89 & 0.89 & 0.89  \\ \midrule

 \multirow{3}{*}{50} 
 & 105 & 5.36 & 5.36 & 5.36 & 0.0 & 0.0 & 0.0 &
 0.88 & 0.89 & 0.88 \\
 & 89 & 4.60 & 4.60 & 4.60 & 0.0 & 0.0 & 0.0 & 
 0.88 & 0.89 & 0.88 \\
 & 79 & 4.09 & 4.09 & 4.09 & 0.0 & 0.0 & 0.0 & 
 0.88 & 0.89 & 0.88 \\ \midrule

 \multirow{3}{*}{30} 
 & 105 & 3.57 & 3.57 & 3.57 & 0.0 & 0.0 & 0.0 & 
 0.87 & 0.87 & 0.87 \\
 & 89 & 3.06 & 3.06 & 3.06 & 0.0 & 0.0 & 0.0 & 
 0.87 & 0.87 & 0.87 \\
 & 79 & 2.72 & 2.72 & 2.72 & 0.0 & 0.0 & 0.0 & 
 0.87 & 0.87 & 0.87 \\ \midrule

 \multirow{3}{*}{10} 
 & 105 & 1.34 & 1.34 & 1.34 & 0.0 & 0.0 & 0.0 & 
 0.84 & 0.86 & 0.86 \\
 & 89 & 1.15 & 1.15 & 1.15 & 0.0 & 0.0 & 0.0 & 
 0.84 & 0.85 & 0.86 \\
 & 79 & 1.02 & 1.02 & 1.02 & 0.0 & 0.0 & 0.0 & 
 0.84 & 0.85 & 0.86 \\ \midrule

 \multirow{3}{*}{1} 
 & 105 & 0.11 & 0.11 & 0.11 & 0.0 & 0.0 & 0.0 & 
 0.82 & 0.79 & 0.78 \\
 & 89 & 0.09 & 0.09 & 0.09 & 0.0 & 0.0 & 0.0 & 
 0.82 & 0.79 & 0.78  \\
 & 79 & 0.08 & 0.08 & 0.08 & 0.0 & 0.0 & 0.0 & 
 0.82 & 0.79 & 0.78 \\ \bottomrule
 
 \end{tabular}%
}
\caption{Performance scores for ResNet50 with ImageNet-21k weights on Medical Dataset APTOS-2019 and its ....EDPE = Effective data per epoch, method=Random}
\label{tab:rn50-ap-rand}
\end{table}

\clearpage
{
\small
\bibliographystyle{tmlr}
\bibliography{egbib}
}

\end{document}